\documentclass{article}

\usepackage[main,final]{neurips_2026}
\usepackage{amsmath}
\usepackage{enumitem}
\usepackage{hyperref}
\usepackage{cleveref}
\usepackage{algorithm}
\usepackage{algpseudocode}

\usepackage[utf8]{inputenc} % allow utf-8 input
\usepackage[T1]{fontenc}    % use 8-bit T1 fonts
\usepackage{hyperref}       % hyperlinks
\usepackage{url}            % simple URL typesetting
\usepackage{booktabs}       % professional-quality tables
\usepackage{amsfonts}       % blackboard math symbols
\usepackage{nicefrac}       % compact symbols for 1/2, etc.
\usepackage{microtype}      % microtypography
\usepackage{xcolor}         % colors
\usepackage{microtype}
\usepackage{graphicx}
\usepackage{subcaption}
\usepackage{booktabs} % for professional tables
\usepackage{hyperref}
\usepackage{amsthm}
\usepackage{amsmath,amsfonts,bm}
\usepackage{booktabs}

\usepackage{amsmath}
\usepackage{graphicx}
\usepackage{xcolor}
\usepackage{tikz}
\usetikzlibrary{arrows.meta}
\usepackage{subcaption}
\newcommand{\varbrac}[1]{ \left\{ #1 \right\} }

\newcommand{\brac}[1]{ \left( #1 \right) }
\newcommand{\Fbrac}[1]{ \left [ #1 \right] }

\def\eqref#1{equation~\ref{#1}}
\def\1{\bm{1}}

\DeclareMathAlphabet{\mathsfit}{\encodingdefault}{\sfdefault}{m}{sl}
\SetMathAlphabet{\mathsfit}{bold}{\encodingdefault}{\sfdefault}{bx}{n}

\newcommand{\E}{\mathbb{E}}

\theoremstyle{plain}
\newtheorem{theorem}{Theorem}[section]

\newtheorem{lemma}[theorem]{Lemma}
\newtheorem{corollary}[theorem]{Corollary}
\theoremstyle{definition}

\theoremstyle{remark}

\title{Improved Sample Complexity of Robust Reinforcement Learning with KL Uncertainty Sets}

\author{%
 Yudan Wang\\
 University of Central Florida\\
  \texttt{yudan.wang@ucf.edu} \\
  \And
  Zilong Deng \\
  Arizona State University \\
  \texttt{zdeng43@asu.edu} \\
  \AND
  Nathaniel D. Bastian \\
  Syracuse University \\
  \texttt{ndbastia@syr.edu} \\
  \And
  Shaofeng Zou\\
  Arizona State University \\
  \texttt{zou@asu.edu} \\
}

\begin{document}

\maketitle

\begin{abstract}
In this paper, we study distributionally robust reinforcement learning with Kullback--Leibler (KL) divergence-defined uncertainty sets. The goal is to find a policy that maximizes the robust value function, defined as the worst-case value over all transition kernels in the uncertainty set. Assuming access to a generative model, we aim to understand the sample complexity of finding an $\epsilon$-optimal robust policy. The best-known sample complexity results in the literature show a non-trivial gap of $\mathcal{O}\{\max\{p_\wedge^{-1}(1-\gamma)^{-1},(1-\gamma)^{-2}\}\}$ between the upper and lower bounds, where $p_\wedge$ denotes the minimum positive transition probability of the nominal transition kernel, and $\gamma$ is the discount factor. More importantly, existing results on the lower bound only cover a limited range of the uncertainty level. In this paper, we develop tighter and complete upper and lower bounds for robust RL with KL-defined uncertainty sets. Our upper bound is the tightest among all existing studies, and it improves upon the best-known bound by at least the order of $\mathcal{O}(\min\{p_\wedge^{-1},(1-\gamma)^{-1}\})$ for uncertainty level  $\delta\in(0,1)$. Furthermore, our lower bound holds for any uncertainty level. Our upper and lower bounds (nearly) match in various regimes, providing improved results for this problem.

\end{abstract}

\section{Introduction}
Reinforcement learning (RL)~\citep{sutton2018reinforcement} provides a powerful framework for agents to learn optimal policies through interaction with an environment, and it has achieved remarkable successes in recent years such as beating human champions in the game of Go \cite{silver2016mastering,silver2017mastering}, achieving grandmaster level in video games \cite{vinyals2019grandmaster,mnih2013playing,hessel2018rainbow}, autonomous driving~\citep{kiran2021deep}, robotics~\citep{kober2013reinforcement}, and large language models~\citep{du2023guiding}. 
However, existing RL approaches usually assume that a learned policy will be deployed in the same environment as the one in which it was trained. Such an assumption is often violated in practice due to, e.g., adversarial perturbations \citep{pinto2017robust}, modeling errors between the simulator and real-world applications \citep{zhao2020sim}, non-stationary environments \cite{padakandla2020reinforcement}, and a limited amount of training data \cite{wang2024achieving}. The discrepancy between the training and test environments leads to a model mismatch, resulting in a notable decline in performance and limiting the suitability of RL in crucial domains.

Noteworthy efforts have been made to develop distributionally robust RL approaches to address the above challenge. The basic idea is to take a conservative approach in the face of model mismatch/uncertainty. An uncertainty set of models is constructed, e.g., one consisting of models close to a simulator, and the robust policy is obtained by optimizing the worst-case performance over the uncertainty set.

The robust planning setting where the uncertainty set is fully known has been well studied in the literature \cite{iyengar2005robust,nilim2004robustness,wang2023robust}. However, the statistical sample complexity of learning a robust policy in the learning setting, where only samples from the nominal transition kernel are available, has not been well understood until recently. We note that robust RL problems with different uncertainty sets are fundamentally different, and in this paper we focus on the uncertainty set defined by Kullback--Leibler (KL) divergence:
\begin{align}
    \mathcal P(\delta)=\{P: \text{KL}(P\|P^0)\leq \delta\},
\end{align}
where $\text{KL}$ denotes the KL divergence, $\delta$ denotes the uncertainty level and $P^0$ is the nominal transition kernel. In the learning setting, $P^0$ is not known, and samples from $P^0$ are available. The goal is to optimize the worst-case value function over the uncertainty set $\mathcal P(\delta)$. Other divergence measures to construct the uncertainty set have also been studied in the literature, e.g., total variation, $\chi^2$, $R$-contamination, Wasserstein distance, and we will discuss these results in related work.

\begin{figure}[htbp]
\centering
\begin{subfigure}[t]{0.32\textwidth}
 \centering
 \begingroup
\definecolor{figureUpper}{HTML}{56B4E9}
\definecolor{figureLower}{HTML}{D55E00}
\definecolor{figurePrevious}{HTML}{39c928}
\color{black}\sffamily
\resizebox{\linewidth}{!}{%
\begin{tikzpicture}[x=0.90cm,y=0.72cm,
  every node/.style={font=\small},
  upper/.style={figureUpper,line width=1.3pt},
  lower/.style={figureLower,line width=1.0pt,densely dashed},
  previous/.style={figurePrevious,line width=1.1pt},
  guide/.style={black!25,thin,densely dashed}]
% The axes are schematic. Each curve label gives its full rate.
% The displayed ordering assumes 1 < log(1/p_wedge).
% The first three x-axis intervals are half their previous widths.
\path[use as bounding box] (-2.55,-1.8) rectangle (10.05,8.45);

% Legend.
\draw[upper] (1.05,8.0)--(1.75,8.0);
\node[anchor=west] at (1.88,8.0) {Upper bound $\widetilde{\mathcal O}$};
\draw[lower] (5.25,8.0)--(5.95,8.0);
\node[anchor=west] at (6.08,8.0) {Lower bound $\Omega$};

% Regime boundaries.
\draw[guide] (0.7,0)--(0.7,2.6);
\draw[guide] (1.6,0)--(1.6,4.226);
\draw[guide] (2.25,0)--(2.25,5.4);
\draw[guide] (8.25,0)--(8.25,7.1);

% Axes.
\draw[-{Latex},thick] (0,0)--(9.75,0);
\draw[-{Latex},thick] (0,0)--(0,7.55);
\node[anchor=south west] at (-2.50,7.65) {Sample complexity};

% Exact rate levels.
\draw (-0.10,1.0)--(0.10,1.0);
\node[anchor=east] at (-0.18,1.0)
 {$\dfrac{|\mathcal S||\mathcal A|}{p_\wedge}$};
\draw (-0.10,2.6)--(0.10,2.6);
\node[anchor=east] at (-0.18,2.6)
 {$\dfrac{|\mathcal S||\mathcal A|}{(1-\gamma)^3\epsilon^2}$};
\draw (-0.10,5.4)--(0.10,5.4);
\node[anchor=east] at (-0.18,5.4)
 {$\dfrac{|\mathcal S||\mathcal A|}{(1-\gamma)^4\epsilon^2}$};
\draw (-0.10,7.1)--(0.10,7.1);
\node[anchor=east] at (-0.18,7.1)
 {$\dfrac{|\mathcal S||\mathcal A|}{p_\wedge(1-\gamma)^4\epsilon^2}$};

% Previous-work upper bound, as in the supplied original figure.
\draw[previous] (0.05,7.1)--(9.55,7.1);
\node[text=figurePrevious,anchor=north] at (3.2,6.92)
 {Upper bound [Li et al., 2025]};

% Upper bound: constant, a straight segment labeled by its square-root rate, then exponential growth.
\draw[upper] (0.05,2.6)--(0.7,2.6)--(2.25,5.4);
\draw[upper] (2.25,5.4)
 .. controls (4.778,5.62) and (7.178,6.45) .. (8.25,7.1);

% Lower bound: constant up to 1-gamma, then linear and exponential growth.
\draw[lower] (0.05,2.6)--(1.6,2.6)--(2.25,5.4);
\draw[lower] (2.25,5.4)
 .. controls (4.778,5.62) and (7.178,6.45) .. (8.25,7.1);
\draw[figureUpper,fill=white,line width=1.1pt] (8.25,7.1) circle[radius=0.055];

% Rate annotations for the two different intermediate bounds.
\node[text=figureUpper] at (1.05,5.50)
 {$\dfrac{|\mathcal S||\mathcal A|\sqrt{\delta}}
              {(1-\gamma)^4\epsilon^2}$};
\draw[-{Latex},figureUpper,thin] (1.05,5.02)--(1.35,3.774);
\node[text=figureLower] at (3.2,2.80)
 {$\dfrac{|\mathcal S||\mathcal A|\delta}
              {(1-\gamma)^4\epsilon^2}$};
\draw[-{Latex},figureLower,thin] (2.5,3.2)--(1.95,4.108);
\node at (6.4,4.15)
 {$\dfrac{|\mathcal S||\mathcal A|e^\delta}
              {(1-\gamma)^4\epsilon^2}$};
\draw[-{Latex},black!65,thin] (6.4,4.67)--(6.4,6.32);

% Worst-path regime: a separate branch at and beyond the threshold.
\draw[upper] (8.25,1.0)--(9.55,1.0);
\draw[lower] (8.25,1.0)--(9.55,1.0);
\fill[figureLower] (8.25,1.0) circle[radius=0.055];
\node[align=center,font=\scriptsize,fill=white,inner sep=2pt] at (8.90,2.00)
 {Worst-path\\regime};

% Exact uncertainty thresholds, with no extra breakpoint at p_wedge.
\node[anchor=north east] at (-0.05,-0.06) {$0$};
\foreach \x in {0.7,1.6,2.25,8.25}
  \draw[thick] (\x,-0.08)--(\x,0.08);
\node[anchor=north,font=\scriptsize] at (0.7,-0.15) {$(1-\gamma)^2$};
\draw[black!40,thin] (1.6,-0.08)--(1.6,-0.61);
\node[anchor=north,font=\scriptsize] at (1.6,-0.65) {$1-\gamma$};
\node[anchor=north,font=\scriptsize] at (2.25,-0.15) {$1$};
\node[anchor=north] at (8.25,-0.15) {$\log(1/p_\wedge)$};

\node[anchor=north] at (4.875,-1.05) {Uncertainty level $\delta$};
\end{tikzpicture}%
}
\endgroup
 \caption{$p_\wedge\geq1-\gamma$.}
\end{subfigure}
\begin{subfigure}[t]{0.32\textwidth}
 \centering
 \begingroup
\definecolor{figureUpper}{HTML}{56B4E9}
\definecolor{figureLower}{HTML}{D55E00}
\definecolor{figurePrevious}{HTML}{39c928}
\color{black}\sffamily
\resizebox{\linewidth}{!}{%
\begin{tikzpicture}[x=0.90cm,y=0.72cm,
  every node/.style={font=\small},
  upper/.style={figureUpper,line width=1.3pt},
  lower/.style={figureLower,line width=1.0pt,densely dashed},
  previous/.style={figurePrevious,line width=1.1pt},
  guide/.style={black!25,thin,densely dashed}]
% Case 2: (1-gamma)^2 <= p_wedge < 1-gamma.
% Schematic coordinates; formulas reproduce the supplied table cells.
% Discontinuous table branches are drawn with separate endpoints.
\path[use as bounding box] (-2.55,-2.0) rectangle (10.05,8.45);

\draw[upper] (1.05,8.0)--(1.75,8.0);
\node[anchor=west] at (1.88,8.0) {Upper bound $\widetilde{\mathcal O}$};
\draw[lower] (5.25,8.0)--(5.95,8.0);
\node[anchor=west] at (6.08,8.0) {Lower bound $\Omega$};

\draw[guide] (0.45,0)--(0.45,2.0);
\draw[guide] (0.95,0)--(0.95,3.65);
\draw[guide] (1.5,0)--(1.5,4.348);
\draw[guide] (2.25,0)--(2.25,5.3);
\draw[guide] (5.5,0)--(5.5,5.3);
\draw[guide] (8.25,0)--(8.25,7.1);

\draw[-{Latex},thick] (0,0)--(9.75,0);
\draw[-{Latex},thick] (0,0)--(0,7.55);
\node[anchor=south west] at (-2.50,7.65) {Sample complexity};

\draw (-0.10,0.7)--(0.10,0.7);
\node[anchor=east] at (-0.18,0.7)
 {$\dfrac{|\mathcal S||\mathcal A|}{p_\wedge}$};
\draw (-0.10,2.0)--(0.10,2.0);
\node[anchor=east] at (-0.18,2.0)
 {$\dfrac{|\mathcal S||\mathcal A|}{(1-\gamma)^3\epsilon^2}$};
\draw (-0.10,3.9)--(0.10,3.9);
\node[anchor=east] at (-0.18,3.9)
 {$\dfrac{|\mathcal S||\mathcal A|}{(1-\gamma)^4\epsilon^2}$};
\draw (-0.10,5.3)--(0.10,5.3);
\node[anchor=east] at (-0.18,5.3)
 {$\dfrac{|\mathcal S||\mathcal A|}{(1-\gamma)^3p_\wedge\epsilon^2}$};
\draw (-0.10,7.1)--(0.10,7.1);
\node[anchor=east] at (-0.18,7.1)
 {$\dfrac{|\mathcal S||\mathcal A|}{(1-\gamma)^4p_\wedge\epsilon^2}$};

\draw[previous] (0.05,7.1)--(9.55,7.1);
\node[text=figurePrevious,anchor=north,font=\scriptsize] at (5.2,7.02)
 {Upper bound [Li et al., 2025]};

% Upper bound changes to a different displayed rate just above p_wedge.
\draw[upper] (0.05,2.0)--(0.45,2.0)--(0.95,2.8);
\fill[figureUpper] (0.95,2.8) circle[radius=0.045];
\draw[upper] (0.95,2.8)--(2.25,5.3)--(5.5,5.3);
\draw[figureUpper,fill=white,line width=1.1pt]
 (0.95,2.8) circle[radius=0.05];
\draw[upper] (5.5,5.3)
 .. controls (6.65,5.52) and (7.60,6.45) .. (8.25,7.1);

\draw[lower] (0.05,2.0)--(1.5,2.0)--(2.25,3.9);
\draw[lower] (2.25,3.9)
 .. controls (3.4,4.03) and (4.6,4.7) .. (5.5,5.3);
\draw[lower] (5.5,5.3)
 .. controls (6.65,5.52) and (7.60,6.45) .. (8.25,7.1);
\draw[figureUpper,fill=white,line width=1.1pt]
 (8.25,7.1) circle[radius=0.055];

\node[text=figureUpper,font=\scriptsize] at (0.95,5.95)
 {$\dfrac{|\mathcal S||\mathcal A|\sqrt{\delta}}
 {(1-\gamma)^4\epsilon^2}$};
\draw[-{Latex},figureUpper,thin]
 (0.25,5.48)--(0.25,2.90)--(0.70,2.40);
\node[text=figureUpper,font=\scriptsize] at (3.20,5.95)
 {$\dfrac{|\mathcal S||\mathcal A|\sqrt{\delta}}
 {(1-\gamma)^3p_\wedge\epsilon^2}$};
\draw[-{Latex},figureUpper,thin] (2.60,5.58)--(1.70,4.602);
\node[text=figureLower] at (3.50,1.3)
 {$\dfrac{|\mathcal S||\mathcal A|\delta}
 {(1-\gamma)^4\epsilon^2}$};
\draw[-{Latex},figureLower,thin] (2.70,1.75)--(1.95,3.14);
\node at (6.50,3.40)
 {$\dfrac{|\mathcal S||\mathcal A|e^\delta}
 {(1-\gamma)^4\epsilon^2}$};
\draw[-{Latex},black!65,thin] (6.50,3.92)--(6.50,5.789);

\draw[upper] (8.25,0.7)--(9.55,0.7);
\draw[lower] (8.25,0.7)--(9.55,0.7);
\fill[figureLower] (8.25,0.7) circle[radius=0.055];
\node[align=center,font=\scriptsize,fill=white,inner sep=2pt] at (8.90,1.75)
 {Worst-path\\regime};

\node[anchor=east] at (-0.08,0) {$0$};
\foreach \x in {0.45,0.95,1.5,2.25,5.5,8.25}
 \draw[thick] (\x,-0.08)--(\x,0.08);
\node[anchor=north,font=\scriptsize] at (0.30,-0.15) {$(1-\gamma)^2$};
\node[anchor=south,font=\scriptsize] at (0.95,0.16) {$p_\wedge$};
\node[anchor=north,font=\scriptsize] at (1.60,-0.15) {$1-\gamma$};
\node[anchor=north,font=\scriptsize] at (2.25,-0.15) {$1$};
\node[anchor=north,font=\scriptsize] at (5.5,-0.15)
 {$1+\log\!\left(\dfrac{1-\gamma}{p_\wedge}\right)$};
\node[anchor=north,font=\scriptsize] at (8.25,-0.15)
 {$\log(1/p_\wedge)$};
\node[anchor=north] at (4.875,-1.25) {Uncertainty level $\delta$};
\end{tikzpicture}%
}
\endgroup
 \caption{$(1-\gamma)^2\leq p_\wedge<1-\gamma$.}
 \label{fig:sample-complexity-case2}
\end{subfigure}\hfill
\begin{subfigure}[t]{0.32\textwidth}
 \centering
 \begingroup
\definecolor{figureUpper}{HTML}{56B4E9}
\definecolor{figureLower}{HTML}{D55E00}
\definecolor{figurePrevious}{HTML}{39c928}
\color{black}\sffamily
\resizebox{\linewidth}{!}{%
\begin{tikzpicture}[x=0.90cm,y=0.72cm,
  every node/.style={font=\small},
  upper/.style={figureUpper,line width=1.3pt},
  lower/.style={figureLower,line width=1.0pt,densely dashed},
  previous/.style={figurePrevious,line width=1.1pt},
  guide/.style={black!25,thin,densely dashed}]
% Case 3: p_wedge < (1-gamma)^2. The axes are schematic.
% All rate labels reproduce the supplied table; discontinuous branches
% are kept separate, with open and closed endpoints as appropriate.
% The displayed threshold order assumes 1-gamma < exp(-1).
\path[use as bounding box] (-2.55,-2.45) rectangle (10.05,8.45);

% Legend.
\draw[upper] (1.05,8.0)--(1.75,8.0);
\node[anchor=west] at (1.88,8.0) {Upper bound $\widetilde{\mathcal O}$};
\draw[lower] (5.25,8.0)--(5.95,8.0);
\node[anchor=west] at (6.08,8.0) {Lower bound $\Omega$};

% Regime boundaries.
\draw[guide] (0.7,0)--(0.7,3.0);
\draw[guide] (1.6,0)--(1.6,4.51);
\draw[guide] (2.25,0)--(2.25,5.6);
\draw[guide] (4.65,0)--(4.65,5.6);
\draw[guide] (6.65,0)--(6.65,5.6);
\draw[guide] (8.5,0)--(8.5,7.1);

% Axes.
\draw[-{Latex},thick] (0,0)--(9.75,0);
\draw[-{Latex},thick] (0,0)--(0,7.55);
\node[anchor=south west] at (-2.50,7.65) {Sample complexity};

% Exact constant rate levels.
\draw (-0.10,0.7)--(0.10,0.7);
\node[anchor=east] at (-0.18,0.7)
 {$\dfrac{|\mathcal S||\mathcal A|}{p_\wedge}$};
\draw (-0.10,2.0)--(0.10,2.0);
\node[anchor=east] at (-0.18,2.0)
 {$\dfrac{|\mathcal S||\mathcal A|}{(1-\gamma)^3\epsilon^2}$};
\draw (-0.10,4.1)--(0.10,4.1);
\node[anchor=east] at (-0.18,4.1)
 {$\dfrac{|\mathcal S||\mathcal A|}{(1-\gamma)^2p_\wedge\epsilon^2}$};
\draw (-0.10,5.6)--(0.10,5.6);
\node[anchor=east] at (-0.18,5.6)
 {$\dfrac{|\mathcal S||\mathcal A|}{(1-\gamma)^3p_\wedge\epsilon^2}$};
\draw (-0.10,7.1)--(0.10,7.1);
\node[anchor=east] at (-0.18,7.1)
 {$\dfrac{|\mathcal S||\mathcal A|}{p_\wedge(1-\gamma)^4\epsilon^2}$};

% Previous-work upper bound from the supplied original figure.
\draw[previous] (0.05,7.1)--(9.55,7.1);
\node[text=figurePrevious,anchor=north,font=\scriptsize] at (5.2,7.02)
 {Upper bound [Li et al., 2025]};

% Upper bound: baseline, square-root rate, plateau, exponential rate.
\draw[upper] (0.05,2.0)--(0.7,2.0);
\draw[upper] (0.7,3.0)--(2.25,5.6)--(6.65,5.6);
\draw[upper] (6.65,5.6)
 .. controls (7.20,5.85) and (7.95,6.5) .. (8.5,7.1);

% Lower bound: baseline, linear rate, plateau, exponential rate.
\draw[lower] (0.05,2.0)--(1.6,2.0);
\draw[lower] (1.6,3.1)--(2.25,4.1)--(4.65,4.1);
\draw[lower] (4.65,4.1)
 .. controls (5.35,4.30) and (6.05,5.0) .. (6.65,5.6)
 .. controls (7.20,5.85) and (7.95,6.5) .. (8.5,7.1);

% Endpoints of the separate rate branches.
\fill[figureLower] (0.7,2.0) circle[radius=0.055];
\draw[figureUpper,fill=white,line width=1.1pt] (0.7,3.0) circle[radius=0.055];
\draw[figureLower,fill=white,line width=1.0pt] (1.6,2.0) circle[radius=0.055];
\fill[figureLower] (1.6,3.1) circle[radius=0.055];
\draw[figureUpper,fill=white,line width=1.1pt] (8.5,7.1) circle[radius=0.055];

% Exact rates of the varying branches.
\node[text=figureUpper,font=\scriptsize] at (1.10,5.95)
 {$\dfrac{|\mathcal S||\mathcal A|\sqrt{\delta}}
              {(1-\gamma)^3p_\wedge\epsilon^2}$};
\draw[-{Latex},figureUpper,thin] (1.10,5.48)--(1.35,4.09);
\node[text=figureLower] at (3.15,2.80)
 {$\dfrac{|\mathcal S||\mathcal A|\delta^2}
              {(1-\gamma)^2p_\wedge\epsilon^2}$};
\draw[-{Latex},figureLower,thin] (2.45,3.20)--(1.95,3.64);
\node at (7.50,3.05)
 {$\dfrac{|\mathcal S||\mathcal A|e^\delta}
              {(1-\gamma)^4\epsilon^2}$};
\draw[-{Latex},black!65,thin] (7.50,3.57)--(7.50,6.2);

% Worst-path regime.
\draw[upper] (8.5,0.7)--(9.55,0.7);
\draw[lower] (8.5,0.7)--(9.55,0.7);
\fill[figureLower] (8.5,0.7) circle[radius=0.055];
\node[align=center,font=\scriptsize,fill=white,inner sep=2pt] at (9.02,1.65)
 {Worst-path\\regime};

% Compact first three intervals; staggered longer threshold labels.
\node[anchor=north east] at (-0.05,-0.06) {$0$};
\foreach \x in {0.7,1.6,2.25,4.65,6.65,8.5}
  \draw[thick] (\x,-0.08)--(\x,0.08);
\node[anchor=north,font=\scriptsize] at (0.7,-0.15) {$p_\wedge$};
\draw[black!40,thin] (1.6,-0.08)--(1.6,-0.61);
\node[anchor=north,font=\scriptsize] at (1.6,-0.65) {$\sqrt{1-\gamma}$};
\node[anchor=north,font=\scriptsize] at (2.25,-0.15) {$1$};
\node[anchor=north,font=\scriptsize] at (4.65,-0.15)
 {$1+\log\!\left(\dfrac{(1-\gamma)^2}{p_\wedge}\right)$};
\draw[black!40,thin] (6.65,-0.08)--(6.65,-0.91);
\node[anchor=north,font=\scriptsize] at (6.65,-0.95)
 {$1+\log\!\left(\dfrac{1-\gamma}{p_\wedge}\right)$};
\node[anchor=north,font=\scriptsize] at (8.5,-0.15) {$\log(1/p_\wedge)$};
\node[anchor=north] at (4.875,-1.90) {Uncertainty level $\delta$};
\end{tikzpicture}%
}
\endgroup
 \caption{$0<p_\wedge<(1-\gamma)^2$.}
 \label{fig:sample-complexity-case3}
\end{subfigure}
\caption{ Illustrations of the obtained sample complexity upper and lower bounds under KL-defined uncertainty sets}
\label{fig:sota}
\end{figure}
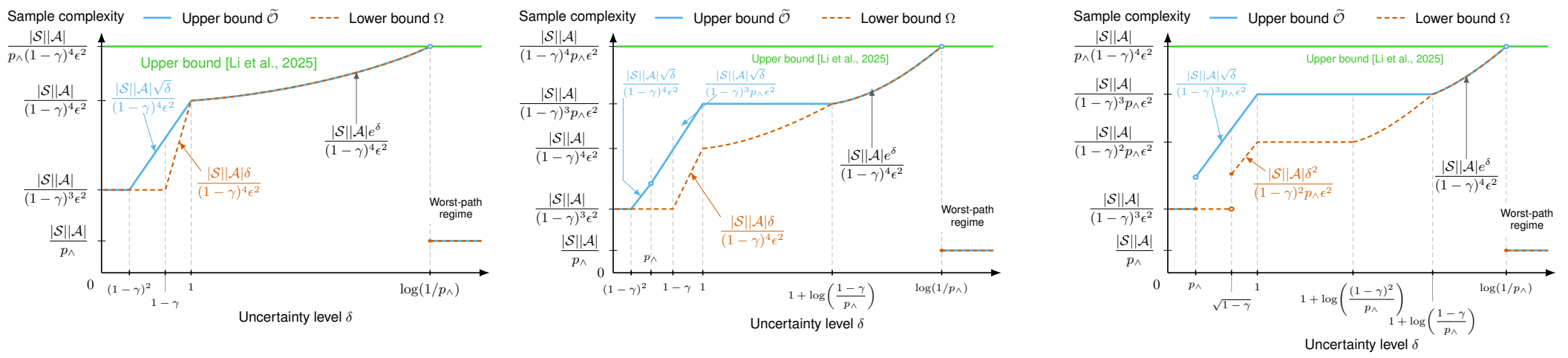

There have been several studies on the sample complexity of robust RL with KL-defined uncertainty set. So far, the best-known upper bound in the literature is $\widetilde{\mathcal{O}}\left(p_\wedge^{-1}(1-\gamma)^{-4}\epsilon^{-2}\lvert\mathcal{S}\rvert\lvert\mathcal{A}\rvert \right) $  \citep{li2025near}, where $p_\wedge$ denotes the minimum positive transition probability of the nominal transition kernel $P^0$ (see definition in Section \ref{sec:formulation}), $\gamma$ denotes the discount factor, $\mathcal{S}$ and $\mathcal{A}$ denote the state and action spaces, and $\epsilon$ denotes the optimality gap. However, when the uncertainty level $\delta\rightarrow0$, the problem shall reduce to a standard RL problem, whose minimax optimal sample complexity shall be $\widetilde{\Theta}\left((1-\gamma)^{-3}\epsilon^{-2}\lvert\mathcal{S}\rvert\lvert\mathcal{A}\rvert \right) $ \cite{wainwright2019variance,li2024q,agarwal2020model}. Therefore, the best-known bound may not be tight when $\delta$ is small (also see Figure~\ref{fig:sota}). 
Moreover, as we summarized in Table~\ref{table:1}, existing results on the lower bound only cover a limited range of the uncertainty level $\delta$, and there exists a non-trivial gap of at least $\mathcal{O}(\max\{p_{\wedge}^{-1}(1-\gamma)^{-1}, (1-\gamma)^{-2}\})$ between the existing upper and lower bounds. 
% which does not capture the special case where the uncertainty level $\delta$ is small. In addition, the existing lower bounds for the KL uncertainty set are only partially characterized (see Figure~\ref{fg1}). Besides, a nontrivial gap remains between the known upper and lower bounds for KL-defined DR-RL.

In this paper, we develop tighter and complete upper and lower bounds for robust RL with KL-defined uncertainty sets. 
\vspace{-0.1in}

\begin{itemize}[leftmargin=*]
\itemsep0em
    \item We first establish an upper bound of
$\widetilde{\mathcal{O}}\big((1-\gamma)^{-3}\epsilon^{-2}\lvert\mathcal{S}\rvert\lvert\mathcal{A}\rvert \big(1+\min\{\exp(\delta/p_\wedge),\sqrt{\delta}/p_\wedge, 1/p_\wedge\}+(1-\gamma)^{-1}\min\{\sqrt{\delta},1\}\big) \big)$ for $\delta\in(0,1)$, which is the state-of-the-art (SOTA) in the literature. This result improves upon the best-known bound by at least an order of $\mathcal{O}(\min\{p_\wedge^{-1},(1-\gamma)^{-1}\})$. Moreover, when the uncertainty level is small, i.e., $\delta\in (0, \min\{(1-\gamma)^2, p_\wedge\})$, our complexity order reduces to the minimax optimal sample complexity for standard MDP \cite{agarwal2020model}. 
    \item We further provide a lower bound of $\widetilde{\Omega}((1-\gamma)^{-3}\epsilon^{-2}\lvert\mathcal{S}\rvert\lvert\mathcal{A}\rvert\max\varbrac{1,\delta e^\delta/(1-\gamma), \delta^2(1-\gamma)/p_\wedge   }) $ for any $\delta \in [1-\gamma, -\log(p_\wedge))$. Moreover, for $\delta \in [-\log(p_\wedge),\infty)$, we prove that this KL-defined robust RL problem reduces to a worst-path RL problem \citep{du2022provably,deng2025near}, and the lower bound is then 
    $\Omega\left(\frac{|\mathcal{S}||\mathcal{A}|}{p_\wedge}\right)$. Together with the lower bound results in \citep{shi2022distributionally} for $\delta<1-\gamma$, these results provide the first complete lower bound characterization that applies to any uncertainty level $\delta$.   For any uncertainty level, the gap between our upper and lower bounds is at most $\mathcal{O}(\max\{(1-\gamma)^{-1}, (1-\gamma)/p_\wedge\})$.  
    \item For a large uncertainty level $\delta \in [-\log(p_\wedge),\infty)$, we further characterize the minimax optimal sample complexity $\widetilde{\Theta}\left(\frac{|\mathcal{S}||\mathcal{A}|}{p_\wedge}\right)$ by proving that the problem is equivalent to a worst-path RL problem \cite{du2022provably,deng2025near}. 
    \item We comprehensively discuss the tightness of our upper and lower bounds, and show that in various regimes the upper and lower bounds match (up to logarithmic terms), and further quantify the gap for regimes where the bounds do not match. See \Cref{fig:sota} for a complete schematic illustration. Overall, our results provide a near-minimax-optimal characterization of the sample complexity for this problem. 
  
    \item We develop a novel analysis of the upper bounds. Specifically, we 
bound the variance induced by the distributionally robust optimization problem using the variance of the value function under worst-case transitions. This approach can be extended to other distributionally robust problems defined with similar uncertainty sets. 
\end{itemize}
\vspace{-0.1in}

\begin{table*}[!ht]
\vskip -0.05in
\begin{center}
\begin{small}
\begin{sc}
\begin{tabular}{p{16mm}|c|c|p{12mm}}
\toprule
Reference  & Upper Bound &Lower Bound &Comments  \\
\midrule
\citep{panaganti2022sample}     & $\widetilde{\mathcal{O }}\brac{\frac{|\mathcal{S}|^2|\mathcal{A}|e^{\frac{1}{1-\gamma}}}{ (1-\gamma)^4\epsilon^2} }$ &-& \\
\hline
\citep{xu2023improved}     & $\widetilde{\mathcal{O }}\brac{\frac{|\mathcal{S}||\mathcal{A}|e^{\frac{1}{1-\gamma}}}{ (1-\gamma)^4\delta^2\epsilon^2} }$ &-& Episodic MDPs\\
\hline
\citep{yang2022toward}     &$\widetilde{\mathcal{O }}\brac{\frac{|\mathcal{S}|^2|\mathcal{A}|}{(1-\gamma)^4p_\wedge^2  \epsilon^2} }$ &- & \\
\hline
\citep{shi2022distributionally}&  $\widetilde{\mathcal{O }}\brac{\frac{|\mathcal{S}||\mathcal{A}| }{p_\wedge \delta (1-\gamma)^4 \epsilon^2} }$&$\widetilde{\Omega}\brac{\frac{|\mathcal{S}||\mathcal{A}|}{(1-\gamma)^3  \epsilon^2} }$, $\delta\in (0,1-\gamma)$ &  \\
&&$\widetilde{\Omega}\brac{\frac{|\mathcal{S}||\mathcal{A}|}{ p_\wedge(1-\gamma)^2  \epsilon^2} }$,$\delta\sim\log(\frac{1}{p_\wedge}) $&\\
\hline
\citep{wang2023finite}  & $\widetilde{\mathcal{O }}\brac{\frac{|\mathcal{S}||\mathcal{A}| }{p_\wedge^6 (1-\gamma)^5 \epsilon^2} }$&-&Model- \\
\cline{1-3}
\citep{wang2023sample}  & $\widetilde{\mathcal{O }}\brac{\frac{|\mathcal{S}||\mathcal{A}| }{p_\wedge^3 (1-\gamma)^5 \epsilon^2} }$&-&Free \\
\cline{1-3}
\citep{wang2024model}  & $\widetilde{\mathcal{O }}\brac{\frac{|\mathcal{S}||\mathcal{A}| }{p_\wedge (1-\gamma)^5 \epsilon^2} }$&-&  \\
\cline{1-3}
\citep{wang2023sample} (VR)  & $\widetilde{\mathcal{O }}\brac{\frac{|\mathcal{S}||\mathcal{A}| }{p_\wedge^3 (1-\gamma)^4 \epsilon^2} }$&-&  \\
\hline
\citep{li2025near}&  $\widetilde{\mathcal{O }}\brac{\frac{|\mathcal{S}||\mathcal{A}| }{p_\wedge (1-\gamma)^4 \epsilon^2} }$ &-&$s$-rec \\
\hline
Our work & $\widetilde{\mathcal{O }}\big(\frac{|\mathcal{S}||\mathcal{A}|}{ (1-\gamma)^3 \epsilon^2}\frac{\min\{p_\wedge e^{\delta/p_\wedge},1\} }{p_\wedge}$  &
$ \widetilde{ {\Omega }}\brac{\frac{\delta^2|\mathcal{S}||\mathcal{A}|}{ p_\wedge (1-\gamma)^2 \epsilon^2
} },\delta \in(1-\gamma,\log(\frac{1}{p_\wedge}))$
&Near-Minimax  \\
&$+\frac{|\mathcal{S}||\mathcal{A}|}{ (1-\gamma)^3 \epsilon^2}\frac{\min\{\sqrt{\delta},1\}e^\delta}{1-\gamma} \big)$&$ \widetilde{ {\Omega }}\brac{\frac{\delta e^\delta|\mathcal{S}||\mathcal{A}|}{  (1-\gamma)^4 \epsilon^2
} }, \delta \in(1-\gamma,\log(\frac{1}{p_\wedge}))$&
\\
& $\widetilde{\mathcal O}\left(\frac{|\mathcal{ S}||\mathcal A|}{p_{\wedge}}\right), \delta\geq \log(\frac{1}{p_\wedge})^{[1]}$&${\Omega}\left(\frac{|\mathcal{ S}||\mathcal A|}{p_{\wedge}}\right), \delta\geq \log(\frac{1}{p_\wedge})^{[1]}$&\\
\bottomrule
\end{tabular}
\end{sc}
\end{small}
\end{center}
\vskip -0.01in
\caption{Upper and Lower Bounds under KL Uncertainty Set. VR: Variance reduction. $s$-rec: $s$-rectangular. ${[1]}$: 1) When $ \delta\geq \log(1/p_\wedge)$, we prove that the KL-defined robust RL reduces to a worst-path RL problem and admits minimax-optimal sample complexity of $\widetilde{\Theta}(\frac{|\mathcal S||\mathcal A|}{p_{\wedge}})$ as shown in \cite{deng2025near}. 2) When $\delta\in (0,1-\gamma)$, we apply the lower bound result of $\widetilde{\Omega}\brac{\frac{|\mathcal{S}||\mathcal{A}|}{(1-\gamma)^3  \epsilon^2} }$ established in \cite{shi2022distributionally}.  }
% VR denotes the result obtained with variance reduce technique.}
\label{table:3}
\end{table*}

 \subsection{Related Work}
% \vspace{-0.1in}
In this section, we review related work on the sample complexity of robust RL. 
\paragraph{Standard RL.} For standard RL, the minimax optimal sample complexity has been derived, $\widetilde{\mathcal{O}}\left((1-\gamma)^{-3}\epsilon^{-2}\lvert\mathcal{S}\rvert\lvert\mathcal{A}\rvert \right) $, for model-based approaches in \cite{agarwal2020model} and for Q-learning-type algorithms \cite{li2024q,wainwright2019variance}. 
\vspace{-0.1in}

\paragraph{Robust RL with KL-defined Uncertainty Set.} 

Sample complexity upper bounds for model-based algorithms have been developed in \citep{panaganti2022sample,yang2022toward,li2025near,shi2022distributionally,zhou2021finite,xu2023improved}. The detailed comparisons are summarized in Table~\ref{table:3}. Early results have an $\exp((1-\gamma)^{-1})$ factor in the upper bound. With a deeper understanding of the KL-defined robust RL problem, the bounds have been progressively improved. The current best-known upper bound is  \cite{li2025near}
$\widetilde{{\mathcal{O}}}(p_\wedge^{-1}(1-\gamma)^{-4}\epsilon^{-2}|\mathcal{S}||\mathcal{A}|).$   However, when compared with the minimax lower bound \citep{shi2022distributionally}, there remains at least a gap of $\mathcal{O}(\max\{p_{\wedge}^{-1}(1-\gamma)^{-1}, (1-\gamma)^{-2}\})$.  Compared to these existing results, our upper and lower bounds are much tighter, and they (nearly) match each other in various regimes. 

% In our work, we provide much tighter upper and lower bounds than existing studies. 
% $\widetilde{\mathcal{O}}\big((1-\gamma)^{-3}\epsilon^{-2}\lvert\mathcal{S}\rvert\lvert\mathcal{A}\rvert \big(\min\{\exp(\delta/p_\wedge), \sqrt{\delta}/p_\wedge\}+(1-\gamma)^{-1}\min\{\sqrt{\delta},1\}\big) \big)$  
% sample complexity, which near matches the lower bound within the uncertainty regime $\delta $. To the best of our knowledge, this represents the state-of-the-art (SOTA) theoretical result for KL-defined DR-RL.  

There are also several studies focusing on model-free algorithms~\citep{wang2023finite,wang2023sample,wang2024model}. These model-free methods generally require larger sample sizes than their model-based counterparts (see Table \ref{table:3}). Although our result is presented in the model-based setting, our analysis can be adapted to model-free algorithms, potentially leading to improved sample complexity.

\paragraph{Other Uncertainty Sets.}
(1) For robust RL with uncertainty sets defined by total variation (TV), several works have developed sample complexity bounds, e.g., \citep{panaganti2022sample,yang2022toward,clavier2023towards,shi2023curious,wang2024model}. In particular, \citep{shi2023curious} established the minimax optimal sample complexity of $\widetilde{\mathcal{O}}\brac{\frac{|\mathcal{S}||\mathcal{A}|}{ (1-\gamma)^2\epsilon^2} \min\varbrac{\frac{1}{\delta},\frac{1}{1-\gamma}} }$.  
(2) For robust RL with $\chi^2$-defined uncertainty sets, there are a few studies, e.g., \citep{panaganti2022sample,yang2022toward,shi2023curious,wang2024model}. Notably, \cite{shi2023curious} provided a near-optimal result, where the upper bound of $\widetilde{\mathcal{O}}\brac{\frac{|\mathcal{S}||\mathcal{A}|}{ (1-\gamma)^3\epsilon^2} \brac{1+\frac{\max\{\delta,\sqrt{\delta}\}}{1-\gamma}} }$ and lower bound of  $\widetilde{\mathcal{O}}\brac{\frac{|\mathcal{S}||\mathcal{A}|}{ (1-\gamma)^3\epsilon^2} \brac{1+\frac{\delta}{1-\gamma}} }$ match within a certain range of the uncertainty level.  
(3) For robust RL with uncertainty sets defined by $L_p$ distance, \cite{clavier2024near} presented a minimax optimal complexity of $\widetilde{\mathcal{O}}\brac{\frac{|\mathcal{S}||\mathcal{A}|}{ (1-\gamma)^2\epsilon^2} \min\varbrac{\frac{1}{\delta},\frac{1}{1-\gamma}} }$. 
    (4) For robust RL with R-contamination uncertainty sets, \cite{wang2021online} presented an upper bound of $\widetilde{\mathcal{O }}(|\mathcal S||\mathcal A|(1-\gamma)^{-5}\epsilon^{-2})$ on the sample complexity. (5) There are also studies on robust RL with Wasserstein distance-defined uncertainty sets, e.g., \citep{mandal2025finite,xu2023improved}.  
We emphasize that robust RL problems with different uncertainty sets are fundamentally different, and therefore, these results for other  uncertainty sets may not be used to solve robust RL with KL-defined uncertainty sets studied in this paper.

\section{Problem Formulation and Preliminaries}\label{sec:formulation}
Distributionally robust RL is typically modeled mathematically as a distributionally robust Markov decision process (DR-MDP). A DR-MDP can be represented by a tuple $\mathcal{M}=(\mathcal{S}, \mathcal{A}, r, \mathcal{P}, \gamma)$. Here, $\mathcal{S},\mathcal{A}$ are the state and action spaces; $r:\mathcal{S}\times\mathcal{A} \to [0,1]$ is the reward function; and $\gamma$  is the discount factor. Different from standard MDPs, $\mathcal{P}$ denotes the uncertainty set of transition kernels. Specifically, in this paper, we consider the following $(s, a)$-rectangular uncertainty set defined by KL divergence:
$\mathcal{P}(\delta)=\prod_{s,a}\mathcal{P}_{s,a}(\delta)= \prod_{s,a} \{P_{s,a}:\text{KL}(P_{s,a}\|P^0_{s,a})\leq \delta \},$
% \begin{align}
%     \mathcal{P}(\delta)=\times_{s,a}\mathcal{P}_{s,a}(\delta)= \times_{s,a} \{P_{s,a}:\text{KL}(P_{s,a}\|P^0_{s,a})\leq \delta \},
% \end{align}
where $\text{KL}$ is the KL divergence: $\text{KL}(\mu\|\mu_0)= \E_{\mu}[\log(\mu/\mu_0)]$ for any two probability distributions $\mu$ and $\mu_0$.  Here, the uncertainty set is centered at a nominal transition kernel $P^0 $ and $P^0_{s,a}$ denotes the nominal transition probability distribution conditioned on $s,a$.   Define the minimal non-zero probability of the nominal transition kernel as 
\begin{align}
    p_\wedge= \min_{s',s,a: P^0(s'|s,a)>0}P^0(s'|s,a). 
\end{align}
\vspace{-0.12in}

A policy $\pi: \mathcal{S}\to \Delta(\mathcal{A})$ denotes the probability of taking actions under different states and represents the strategy of the agent. Here,
$\Delta(\mathcal{A})$ denotes the probability simplex over $\mathcal A$.
Given the transition kernel $P$, the value function of a policy $\pi$ is defined as the expected cumulative reward the agent receives by following the policy starting from $s$: $V^\pi_{{P}}(s)= \E \Fbrac{\sum_{t=0}^\infty  \gamma^t r_t|s_0=s,{P},\pi}$.  
The $Q$-function is defined as the expected cumulative reward starting from $s$ and action $a$: $ Q_{ {P}}^\pi(s,a)=\mathbb E\Fbrac{\sum_{t=0}^\infty  \gamma^t r_t|s_0=s,a_0=a,{ {P}},\pi}. $
% \begin{align}
%      \nonumber
% \end{align}

The goal in a DR-MDP is to optimize the worst-case performance over the uncertainty set, i.e., to optimize the robust value function:
\begin{align}\label{eq:optimal}
    \pi^{*} &=\arg\max_\pi V^{\pi}=\arg\max_\pi \inf_{P\in \mathcal{P}(\delta) } V^\pi_{P},
\end{align}
where $V^{\pi}=\inf_{P\in \mathcal{P}(\delta) } V^\pi_{P}$ denotes the worst-case value function of policy $\pi$.
Let $V^*=V^{\pi^* }$ be the optimal robust value function. 
Define the robust state-action value function: $ Q^{\pi} (s,a)=\inf_{P\in \mathcal{P}(\delta) } Q^\pi_{P}(s,a).$ 
The optimal robust state-action value function is hence denoted by
\begin{align}
    Q^{*} (s,a)=\sup_{\pi}Q^{\pi} (s,a),
\end{align}
and the optimal robust policy can be directly obtained: $\pi^{*} (s)=\arg\max_a  Q^{*} (s,a)$. 
It was shown in \citep{iyengar2005robust} that the optimal robust $Q$-function satisfies the following robust Bellman equation:
\vspace{-0.2in}

\begin{align}\label{eq:bell}
    Q^{*} (s,a)
    = r(s,a)+\gamma\inf_{P\in \mathcal{P}(\delta) }\mathbb E_{P}\Fbrac{\max_{a'\in \mathcal{A}}Q^{*} (s',a') } .
\end{align}

Besides, the robust Bellman optimality operator is defined as
\begin{align}
    \mathcal{T}  (Q(s,a))= r(s,a)+ \gamma \inf_{P\in \mathcal{P}(\delta) }\E_{s'\sim P_{s,a}}[V(s')],
\end{align}
where $V(s')=\max_{a'}Q(s',a')$.

% Hence DR-RL aims to find the optimal robust policy, or equivalently to solve the robust Bellman \eqref{eq:bell}. 

% \section{Related Theoretical Results}

\subsection{Strong Duality}
% In general, the $\rho$-constrained uncertainty set includes infinite available distribution and can not be solved directly. Previous works \citep{iyengar2005robust,hu2013kullback} provide the duality form of minimax problems, which can be solved within finite steps and is widely used to solve DR-RL problems.
For a general uncertainty set $\mathcal{P}$, directly computing $\inf_{p\in\mathcal{P}} p^\top V$ for any vector $V$ is usually computationally expensive. However, this optimization problem can be equivalently solved using its dual, which is a convex optimization problem \citep{iyengar2005robust,hu2013kullback}. These results play a crucial role in our algorithm design; therefore, we introduce the dual form corresponding to the KL-defined uncertainty set as follows.
\begin{lemma}[\cite{iyengar2005robust,hu2013kullback}]\label{lemma:strongduality}
    The optimization problem 
     \begin{align}
        & \min \quad \mathbb E_{s\sim q}[V(s)]
    \qquad\text{subject \quad to }\quad q\in \varbrac{\text{KL} 
        \brac{q\|p}\leq \delta, q\in \Delta(\mathcal{S}) },
        \end{align} is equivalent to
     \begin{align}
        \max_{\lambda\geq0} \varbrac{-\lambda \log\brac{\mathbb E_{s\sim p} \Fbrac{\exp\brac{-\frac{V(s)}{\lambda}} } }-\lambda\delta }. 
    \end{align}
\end{lemma}\label{lm:kl}
% Based on this duality function, the value iteration update rule can be built.

\subsection{Generative Model and Empirical Transition Kernel}
In this work, we study the sample complexity of robust RL with a KL-defined uncertainty set in the generative model framework introduced in \cite{kearns89}. This is a widely used sampling mechanism for understanding the fundamental sample complexity of RL  problems~\citep{panaganti2022sample,yang2022toward,shi2022distributionally,agarwal2020model}. 
In the generative model setting, we assume sampling access to a generative model of the nominal transition kernel $P^0$, and we can query the next state $s'$ sampled from $P^0_{s,a}$,
given any state-action pair $(s,a)$. Specifically, we sample $N$ i.i.d. samples from $P^0_{s,a}$ for any $(s,a)\in \mathcal S\times\mathcal A$:
$$s_{n,s,a} \overset{i.i.d.}{\sim} P^0(\cdot|s,a), n=1,2,...,N, \forall (s,a)\in \mathcal{S}\times\mathcal{A}.$$
Then, the total number of samples is $N |\mathcal{S}||\mathcal{A}|$. 
Using samples from the generative model, we compute the empirical estimate $ \widehat{P}^0= \prod_{s,a}\widehat P^0(\cdot|s,a)$ of the nominal transition kernel using a maximum likelihood estimate
 as follows:
\vspace{-0.3in}
 
\begin{align}\label{eq:empiricalkernel}
    \widehat P^0(s'|s,a):= \frac{1}{N}\sum_{n=1}^N\mathbb{I}\varbrac{s_{n,s,a}=s'}. 
\end{align}
Here, $\mathbb{I}$ denotes the indicator function. In the next section, we will use the empirical estimate of the nominal transition kernel in the robust value iteration algorithm \citep{iyengar2005robust,nilim2004robustness} and understand its sample complexity.

\paragraph{Notation.} Throughout this paper, $\mathcal{O}(\cdot)$ denotes an upper bound up to universal constant factors. Specifically, given functions $f$ and $g$ defined on a domain $D$, we write $f(x)=O(g(x))$ if there exists a constant $M>0$ such that $|f(x)| \le M g(x)$ for all $x \in D$. Besides, $\widetilde{\mathcal{O}}(\cdot)$ denotes an upper bound up to universal constant and logarithmic factors.    
Similarly, $\Omega(\cdot)$ denotes a lower bound up to universal constant factors, and $\widetilde{\Omega}(\cdot)$ denotes a lower bound up to universal constant and logarithmic factors. Moreover, $\Theta(\cdot)$ denotes a tight (minimax-optimal) bound up to universal constant factors, that is, $f(x)=\Theta(g(x))$ if $f(x)=\mathcal{O}(g(x))$ and $f(x)=\Omega(g(x))$. Similarly, $\widetilde{\Theta}(\cdot)$ denotes a tight bound up to universal constant and logarithmic factors.

\section{Main Results}

We use the standard robust value iteration algorithm with the empirically estimated uncertainty set. 
Firstly, from the generative model, we draw $N$ samples from every state-action pair, and we can get the empirical nominal transition kernel $\widehat{P}^0(\cdot|s,a) $ for $\forall (s,a)\in \mathcal{S}\times\mathcal{A}$. Based on the empirical nominal transition kernel, we further define the empirical uncertainty set, where we replace the unknown true nominal transition kernel $P^0$ with its empirical estimate:
\begin{align}
    \widehat{\mathcal{P}}_{s,a}&=\{q: \text{KL}(q\mid \widehat{P}^0_{s,a})\leq\delta\}, \qquad
    \text{and } \widehat{\mathcal{P}}=\prod_{s,a} \widehat{\mathcal{P}}_{s,a}.
\end{align}
Then, we further define the empirical robust Bellman optimality operator as follows: 
\begin{align}
   \widehat{ \mathcal{T}}  (Q(s,a))= r(s,a)+ \gamma \inf_{q\in \widehat{\mathcal{P}}_{s,a}(\delta)}\E_q[V(s')],
\end{align}
where $V(s')=\max_{a'} Q(s',a')$. By Lemma \ref{lm:kl}, the robust Bellman optimality operator can be computed as follows:
\begin{align}\label{eq:rbo}
       \widehat{ \mathcal{T}}   (Q(s,a))=r(s,a)
    + \gamma\max_{\lambda\geq 0}\left\{ 
    -\lambda \log\left( \E_{\widehat{P}^0(\cdot|s,a)}\left[ \exp\left(-\frac{V}{\lambda}  \right)  \right]  \right)-\lambda\delta
    \right\}. 
\end{align}

% Here, we provide a brief algorithm here. 
\begin{algorithm}[tb]
  \caption{Robust Value Iteration Algorithm}\label{alg:ravi}
  \begin{algorithmic}
\State{\bfseries Initialization:}  step size $\beta$, iteration number $T$,sample count $N$,  and $Q = 0$ 
\State Sample $N$ i.i.d. samples from $P^0(\cdot|s,a),\forall (s,a)$
% \ENDFOR
\State Compute the empirical transition kernel $\widehat{P}^0(\cdot|s,a)$ according to Eq. \ref{eq:empiricalkernel}
\For{$t=0,...,T-1$}
\State Update the robust $Q$-function: 
 $ Q(s,a)\leftarrow \hat{\mathcal{T}} (Q(s,a))$ according to Eq. \ref{eq:rbo}, $\forall(s,a)\in\mathcal{S}\times\mathcal{A}$

\EndFor
\State{\bfseries Output:} $\pi_T(\cdot|s)=\arg\max_a Q(s,a), \forall s\in\mathcal S$
\end{algorithmic}
\end{algorithm}
In Algorithm \ref{alg:ravi}, we present the robust value iteration algorithm with an empirically estimated uncertainty set. We note that the convergence of the robust value iteration algorithm has been shown in prior works \citep{iyengar2005robust,nilim2004robustness} for any given uncertainty set. 
In the following, our focus is on understanding how many samples are needed to guarantee that the solution to Algorithm \ref{alg:ravi} using the empirical uncertainty set is close to the optimal robust policy defined w.r.t. the original uncertainty set in \cref{eq:optimal}.
\subsection{Sample Complexity Upper Bound}\label{sec:upper}
We first present the upper bound on the sample complexity of Algorithm \ref{alg:ravi}.
\begin{theorem}\label{thm:up}
   Let $\pi_T $ be the output of Algorithm \ref{alg:ravi} with uncertainty level $0<\delta\le 1$, $T\geq\left\lceil
 \frac{\log(24/((1-\gamma)^2\epsilon))}{-\log\gamma}
 \right\rceil$ and $V_T= V^{\pi_T}$.  Then, with probability at least $1-\eta\in(0,1)$, the gap between the optimal robust policy and output policy can be bounded by
   \begin{align}
      \|V^*-V_T\|_\infty\leq  \widetilde{\mathcal{O}}\left((1-\gamma)^{-3/2}\frac 1{\sqrt{N}}\sqrt{\min\varbrac{\exp\brac{\frac{2\delta}{p_\wedge}},1 +\frac{2\sqrt{\delta}}{p_\wedge}, \frac{2}{p_\wedge}}+\frac{\sqrt{\delta}}{1-\gamma}}\right)  .
       \nonumber
   \end{align}
   To obtain an $\epsilon$-optimal policy, i.e. $\|V^*-V_T\|_\infty\leq \epsilon$, the total sample complexity is 
   \begin{align} 
       \mathcal{O}\left((1-\gamma)^{-3}\epsilon^{-2}\lvert\mathcal{S}\rvert\lvert\mathcal{A}\rvert\log\brac{\frac{|\mathcal{S}||\mathcal{A}|N}{(1-\gamma)\eta}}  \brac{\min\varbrac{\exp\brac{\frac{2\delta}{p_\wedge}},1+\frac{2\sqrt{\delta}}{p_\wedge}, \frac{2}{p_\wedge}}+\frac{\sqrt{\delta}}{1-\gamma} }\right).\nonumber
   \end{align}
\end{theorem}     
% \zou{check all big O and Omega notation in the paper}
As we discussed in the introduction, existing upper bound results have a gap of $\mathcal O(p_\wedge^{-1}(1-\gamma)^{-1})$ compared to the minimax lower bound of standard RL when the uncertainty level is small. The intuition is that the problem shall reduce to the standard RL setting when the uncertainty level is small (see Figs. \ref{fig:sota}). In contrast, our sample complexity reduces to the minimax-optimal sample complexity of standard RL:  $\mathcal{O}((1-\gamma)^{-3}\epsilon^{-2} \lvert\mathcal{S}\rvert\lvert\mathcal{A}\rvert)$~\citep{agarwal2020model,wainwright2019variance,li2024q}, when the uncertainty level is small.
\begin{corollary}
    When the uncertainty level $\delta$ is small: $\delta\in (0, \min\{(1-\gamma)^2, p_\wedge\})$, the sample complexity of Algorithm \ref{alg:ravi} reduces to the minimax-optimal sample complexity of standard RL:  
    \begin{align}
        \mathcal{O}((1-\gamma)^{-3}\epsilon^{-2} \lvert\mathcal{S}\rvert\lvert\mathcal{A}\rvert).
    \end{align}
\end{corollary}

% From the results in Theorem \ref{thm:up}, we conclude that the sample complexity order matches $\mathcal{O}((1-\gamma)^{-3}\epsilon^{-2} \lvert\mathcal{S}\rvert\lvert\mathcal{A}\rvert)$ whenever the uncertainty level satisfies $\delta\in (0, \min\{p_\wedge, (1-\gamma)^2\}). $ 

% Here, we impose the condition $e  < p_\wedge^{-1}$, since otherwise the worst-case transition would concentrate all probability mass on the worst-case state. 

% This discussion could help us better compare with lower bound for different conditions.  
% , we could see that the results in our work include a factor $\mathcal{O}(1/p_\wedge)$. Whether the term $\mathcal{O}(p_\wedge )$ is necessary in the complexity results is still opening. The related lower bound design shows that this term is necessary under some uncertainty set range \citep{shi2022distributionally}. 
% Compared with previous robust RL work within KL uncertainty set, we could see that our result is a near-optimal result with a certain uncertainty level, i.e. when $\delta \in (\log (1/p_\wedge)-6, \log (1/p_\wedge)-5 ) $ \cite{shi2022distributionally}. 

% \begin{remark}
Compared with the previous KL-defined robust RL works, our results establish a state-of-the-art (SOTA) upper bound which improves upon the best known bound by at least an order of $\mathcal{O}(\min\{(1-\gamma)^{-1}, p^{-1}_\wedge\})$ ~\citep{panaganti2022sample,yang2022toward,li2025near} when uncertainty level $\delta\le 1$. Moreover, our work is the first to achieve the same sample complexity order as standard (non-robust) RL when the uncertainty level is sufficiently small.  
% We note that when the uncertainty level $\delta$ is large, the problem reduces to a worst-case RL problem, which we discuss in \Cref{sec:worstpath}.
% \zou{add discussion corresponding to the two cases of the two figs.}
% Moreover, we show that our bound attains the minimax-optimal sample complexity when the uncertainty level satisfies $\delta \geq \log p_\wedge - 6$ \citep{shi2022distributionally}. Along the way, we establish several new lemmas and structural properties of KL-defined DR-RL, which may be of independent interest and further illuminate the foundations of robust RL. Finally, the analysis framework we introduce is general and can be extended to a broader class of distributionally robust RL problems. 
% \end{remark}

% \subsection{Model-Free Method with Variance Reduction} 

% \begin{algorithm}
% \caption{Robust Model-Free T-MLMC Algorithm with Variance Reduction}\label{alg:rmf}
% \begin{algorithmic}
% \State{\bfseries Initial:}  stepsizes $\beta$, iteration number $T$ and $N$, $Q$ 
% \FOR{$t=0,...,T-1$}
% \FOR{$(s,a)\in\mathcal{S}\times\mathcal{A}$}
% \FOR{$n=1,...,N$}
% \State Sample $s_{s,a,n}\sim P^0(\cdot|s,a)$
% \ENDFOR
% \State Obtain the empirical transition $\widehat{P}^0(\cdot|s,a) $
% \State Robust Bellman operator from \eqref{eq:rbo} 
% % $\mathcal{T} (Q(s,a)) $
% \State Update the robust $Q$-table: 
% \State $\qquad Q(s,a)\to \mathcal{T} (Q(s,a))$
% \ENDFOR
% \ENDFOR
% \State{\bfseries Output:} $\pi_T$, where $\pi_T(\cdot|s)=\arg\max_a Q$
% \end{algorithmic}
% \end{algorithm}

\begin{theorem}[Large uncertainty level $\delta>1$]\label{thm:up2}
   Let $\pi_T $ be the output of Algorithm \ref{alg:ravi} with uncertainty level $1 \le\delta\le \log(\frac{1}{p_\wedge})$, $T\geq\left\lceil
 \frac{\log(24/((1-\gamma)^2\epsilon))}{-\log\gamma}
 \right\rceil$ and $V_T= V^{\pi_T}$.  Then, with probability at least $1-\eta\in(0,1)$, the gap between the optimal robust policy and output policy can be bounded by
   \begin{align}
      \|V^*-V_T\|_\infty\leq  \widetilde{\mathcal{O}}\left((1-\gamma)^{-3/2}\frac1{\sqrt{N}}
       \sqrt{{ \frac{1}{p_\wedge}}+\frac{e^\delta}{1-\gamma}}\right)  .
       \nonumber
   \end{align}
   To obtain an $\epsilon$-optimal policy, i.e. $\|V^*-V_T\|_\infty\leq \epsilon$, the total sample complexity is 
   \begin{align} 
       \mathcal{O}\left((1-\gamma)^{-3}\epsilon^{-2}\lvert\mathcal{S}\rvert\lvert\mathcal{A}\rvert\log\brac{\frac{|\mathcal{S}||\mathcal{A}|N}{(1-\gamma)\eta}}  \brac{ \frac{1}{p_\wedge}+\frac{e^\delta}{1-\gamma} }\right).\nonumber
   \end{align}
\end{theorem}     

Here, we consider the regime $e^\delta<p_\wedge^{-1}$. When the uncertainty level satisfies $\delta\geq \log\brac{p_\wedge^{-1}}$, the robust RL problem with KL uncertainty sets reduces to a worst-path problem; see the discussion in \Cref{sec:worstpath}.

\subsection{Sample Complexity Lower Bounds}\label{sec:lower}
To fully understand the improvement and optimality of the upper bound in Theorem~\ref{thm:up}, we derive  minimax lower bounds.  In this section, we focus on the case when the uncertainty level $\delta < \log(p_\wedge^{-1})$. The case when $\delta \geq \log(p_\wedge^{-1})$ will be studied in \cref{sec:worstpath}.

% $p_\wedge> 1-\gamma$
\begin{theorem}\label{thm:lb1}
Consider any $(\mathcal{S},\mathcal{A},\gamma, \delta, \epsilon) $ such that $\gamma\in (3/4,1),\delta\in[1-\gamma, \log(1/p_\wedge)) $, and $\epsilon<\frac {1}{1000(1-\gamma)(\delta+\log(2/(1-\gamma)))} $. Then there exist two DR-MDPs $\mathcal{M}_0, \mathcal{M}_1$ defined by the KL-defined uncertainty sets, and a dataset with $N$ independent samples per $(s, a)$ pair from the nominal transition kernel (for $\mathcal{M}_0$ and $\mathcal{M}_1$ respectively), such that 
\begin{align}\label{eq:inf}
    \inf_{\widehat\pi} \max &\big\{\mathbb P_0\brac{||V_0^*-V_0^{\widehat{\pi}}||_\infty >\epsilon }, \mathbb P_1\brac{||V_1^*-V_1^{\widehat{\pi}}||_\infty >\epsilon }\big\}>\frac{1}{8},
\end{align}
provided that the total number of samples satisfies
\begin{align*}
     N |\mathcal{S}||\mathcal{A}|\leq \widetilde{\Omega} \brac{\frac{|\mathcal{S}||\mathcal{A}|\delta e^\delta}{(1-\gamma)^4\epsilon^2}}.
\end{align*}
\end{theorem}    

Here, $V_0^*$ (resp. $V_1^*$) denotes the optimal robust value function under $\mathcal M_0$ (resp. $\mathcal M_1$), $V_0^{\hat \pi}$ (resp. $V_1^{\hat \pi}$) denotes the robust value function of the policy returned by any learning algorithm $\hat \pi$ using samples from  $\mathcal M_0$ (resp. $\mathcal M_1$). 
Moreover, $\mathbb P_0\brac{||V_0^*-V_0^{\widehat{\pi}}||_\infty >\epsilon }$ (resp. $\mathbb P_1$)  denotes the probability that the output policy $\widehat{\pi}$ fails to converge to the optimal robust policy using samples from the DR-MDP $\mathcal{M}_0$. Specifically, this event occurs when the gap between the optimal value function $V_0^*$ and the value function induced by the algorithm’s output policy $V_0^{\widehat{\pi}}$ exceeds $\epsilon$ in the $\ell_\infty$ norm. When \cref{eq:inf} holds, there exists at least one MDP in $\mathcal{M}_0, \mathcal{M}_1$ under which no algorithm can guarantee that the output policy converges to the optimal robust policy.
% When the Equation~\ref{eq:inf} holds, there exists at least one condition the any algorithm can not let output policy converge at same time. 
% represent the probability of the output policy does not converge to the optimal policy under MDPs $\mathcal{M}_0$,  where the gap between optimal value function under MDPs $\mathcal{M}_0$ and the value function of algorithm output policy $\widehat{pi} $ is larger than $\epsilon $. 

Theorem \ref{thm:lb1} provides a lower bound for KL-defined robust RL when uncertainty level $\delta \in [1-\gamma, -\log(p_\wedge))$. For $\delta\in(0,1-\gamma)$, we borrow the lower bound of ${\Omega} \brac{\frac{|\mathcal{S}||\mathcal{A}| }{(1-\gamma)^3\epsilon^2}}$
from \cite{shi2022distributionally}. 
Combining our results with the one in \cite{shi2022distributionally}, we obtain a lower bound of $\Omega \brac{\frac{|\mathcal{S}||\mathcal{A}| }{(1-\gamma)^3\epsilon^2}\max\{\frac{\delta e^\delta}{1-\gamma},1 \}}$ that holds for any $\delta\in (0, -\log(p_\wedge))$.  
We further observe that when the uncertainty level is large, i.e., $\delta = \mathcal{O}(1)$, a sample complexity of
$\Omega\left( \frac{ e^\delta|\mathcal{S}||\mathcal{A}|}{(1-\gamma)^4 \epsilon^2} \right)$
is necessary. This lower bound is strictly larger than the minimax-optimal bound for standard RL, implying that KL-defined robust RL is intrinsically more challenging than standard RL.

\begin{theorem}
\label{thm:lb2}
Under the same conditions as in Theorem \ref{thm:lb1} and $p_\wedge \leq \frac{1}{8}$, \cref{eq:inf} holds   
 provided that the total number of samples satisfies
\vspace{-0.3in}
 
\begin{align}
    N |\mathcal{S}||\mathcal{A}|\leq \widetilde{\Omega} \brac{\frac{|\mathcal{S}||\mathcal{A}|\delta^2 }{(1-\gamma)^2p_\wedge \epsilon^2}}.
\end{align}
\end{theorem}     
Theorem \ref{thm:lb1} and Theorem \ref{thm:lb2} are derived using two different constructions of hard MDPs, and therefore, they hold simultaneously. Combining Theorem \ref{thm:lb1}, Theorem \ref{thm:lb2} and results from \cite{shi2022distributionally}, we obtain the following minimax lower bound. 
\begin{corollary}
    For any $\delta\in (1-\gamma, \log(1/p_\wedge))$, the overall minimax lower bound is given by
    \begin{align}
        \widetilde\Omega \brac{\frac{|\mathcal{S}||\mathcal{A}| }{(1-\gamma)^3\epsilon^2}\max\varbrac{1,\frac{\delta e^\delta}{1-\gamma}, \frac{\delta^2(1-\gamma)}{p_\wedge} }} .
    \end{align} 
\end{corollary}

% Similarly, using the lower bound results for $\delta\in(0,1-\gamma)$ in \citep{shi2022distributionally}, our results yield a lower bound of $ \Omega \brac{\frac{|\mathcal{S}||\mathcal{A}| }{(1-\gamma)^3\epsilon^2}\max\{\frac{\delta}{p_\wedge},1 \}}$ that covers the entire range of uncertainty levels. Moreover, both this bound and the lower bound in Theorem~\ref{thm:lb1} do not require any constraint on the relationship between the minimal support probability and the uncertainty level, implying that the two lower bounds can hold simultaneously. As a result, the overall minimax lower bound is given by $\Omega \brac{\frac{|\mathcal{S}||\mathcal{A}| }{(1-\gamma)^3\epsilon^2}\max\varbrac{1,\frac{1}{1-\gamma}, \frac{1-\gamma}{p_\wedge} }} $. 

The lower bound above implies that when the uncertainty level remains on the order of a constant, even a sufficiently small minimum positive transition probability can trigger a large variance, substantially impeding the convergence of learning algorithms. This observation not only sharpens the theoretical picture but also offers a principled explanation for the performance degradation often seen in practice.

% Again, the case when $\delta \geq -\log(p_\wedge)$ will be studied in \Cref{sec:worstpath}. 

\subsection{Worst-Path RL}\label{sec:worstpath}
In this section, we analyze the case when the uncertainty level $ \delta\geq \log\brac{p_\wedge^{-1}}$. Results here will provide insight about the fundamental difficulty of the problem when the uncertainty set is overly pessimistic, and the problem reduces to a worst-path RL problem \citep{du2022provably,deng2025near},  in which case the agent aims to maximize the minimum possible cumulative reward. Moreover, together with the lower bound results in \Cref{sec:lower}, our lower bound covers any value of $\delta.$

\begin{lemma}\label{lemma:worst}
For any $(s,a) \in \mathcal{S} \times \mathcal{A}$, if
$\delta \geq \log\!\left(\frac{1}{p_\wedge}\right),$ 
then the worst-case transition kernel concentrates all probability mass on the worst-case next state:
$$\inf_{P\in \mathcal{P}(\delta) }\E_{s'\sim P_{s,a}}[V(s')]=\min_{s' : P(s' \mid s,a) > 0} V(s').$$
\end{lemma}
Lemma \ref{lemma:worst} implies that when $\delta$ is large, the robust Bellman equation in \cref{eq:bell} reduces to
$Q^{*} (s,a)=r(s,a)+\gamma
\min_{s' : P^0(s' \mid s,a) > 0}\big( \max_{a' \in \mathcal{A}} Q^{*} (s',a') \big),$
which coincides with the worst-path RL Bellman equation (Equation~(31)) in \cite{deng2025near}.
Therefore, when $\delta \geq \log\!\left(\frac{1}{p_\wedge}\right)$, the KL-defined DR-RL problem reduces to a worst-path RL problem. 

In the following, we quote the minimax sample complexity results derived in \cite{deng2025near} for the worst-path RL problem. 
\begin{theorem}[\cite{deng2025near}, Theorems~3 and~4]\label{thm:deng}
For a worst-path RL problem with $|\mathcal{S}|$ states, $|\mathcal{A}|$ actions, and minimum positive transition probability $p_\wedge$, the minimax-optimal sample complexity order is
$\widetilde{\Theta}\!\left(\frac{|\mathcal{S}||\mathcal{A}|}{p_\wedge}\right).$
\end{theorem}

Combining Lemma \ref{lemma:worst} and Theorem \ref{thm:deng}, we have the following minimax sample complexity results for our KL-defined robust RL problem.
\begin{corollary}\label{corollary:worst}
    For any $\delta \geq \log\left(\frac{1}{p_\wedge}\right),$ the minimax optimal sample complexity for the KL-defined robust RL problem is 
$\widetilde{\Theta}\!\left(\frac{|\mathcal{S}||\mathcal{A}|}{p_\wedge}\right).$
\end{corollary}
Here, we note that the sample complexity lower bound established in \cite{shi2022distributionally}
applies to uncertainty levels such that
$\delta \in \big(\log(1/p_\wedge)-6,\; \log(1/p_\wedge)-5\big),$
and this set does not overlap with the set $[\log\left(\frac{1}{p_\wedge}\right),\infty)$ we consider here. Therefore, their results do not contradict ours. 
We also notice that the sample complexity in \Cref{corollary:worst} does not depend on $\epsilon$. As discussed in \cite{du2022provably}, this is because the worst-path RL problem only concerns the minimum
value of successor states, which is independent of specific transition probabilities. 
Corollary \ref{corollary:worst} provides both upper and lower bounds for any $\delta \geq \log\left(\frac{1}{p_\wedge}\right).$  Combined with the lower-bound results in Section~\ref{sec:lower}, we obtain the first complete characterization of the lower bound for any uncertainty level in the literature.

\subsection{Tightness of Upper and Lower Bounds}
In this section, we analyze the tightness of our upper and lower bounds. As will be shown below, for a wide range of uncertainty levels, our upper and lower bounds match with each other (up to logarithmic terms), and thus are minimax optimal. Moreover,  for regimes where the bounds do not match, we also quantify the gap between the upper and lower bounds.
\vspace{-0.12in}

\paragraph{Case 1: $p_\wedge\geq1-\gamma$.}
The upper bound is controlled by the standard-RL term and the term
involving $\min\{\sqrt\delta,1\}/(1-\gamma)$. Among the two new lower
bounds, the one proportional to
$|\mathcal S||\mathcal A|\delta/((1-\gamma)^4\epsilon^2)$ dominates.
The rates in Table~\ref{tab:tightness-large-support} match up to
logarithmic factors for $\delta\leq(1-\gamma)^2$, for
$1\leq\delta<\log(1/p_\wedge)$, and in the worst-path regime.
For $(1-\gamma)^2<\delta<1-\gamma$, their ratio is
$\sqrt\delta/(1-\gamma)$ up to logarithmic factors; for
$1-\gamma\leq\delta\le1$, it is $1/\sqrt\delta$.
Thus the largest polynomial gap in this case is
$(1-\gamma)^{-1/2}$.

\begin{table}[H]
 % \captionsetup{labelfont={color=red},textfont={color=red}}
 \centering
 \footnotesize
 \setlength{\tabcolsep}{2pt}
 \renewcommand{\arraystretch}{1.5}
 \begin{tabular}{p{0.26\linewidth}p{0.24\linewidth}p{0.22\linewidth}p{0.245\linewidth}}
 \toprule
 Uncertainty level & Upper bound & Lower bound & Gap\\
 \midrule
$0<\delta\leq(1-\gamma)^2$
 & $\widetilde{\mathcal O}\!\left(\frac{|\mathcal S||\mathcal A|}{(1-\gamma)^3\epsilon^2}\right)$
 & $\widetilde{\Omega}\!\left(\frac{|\mathcal S||\mathcal A|}{(1-\gamma)^3\epsilon^2}\right)$
 & $\widetilde {\mathcal O}\!\left(1\right)$\\
$(1-\gamma)^2<\delta<1-\gamma$
 & $\widetilde{\mathcal O}\!\left(\frac{|\mathcal S||\mathcal A|\sqrt\delta}{(1-\gamma)^4\epsilon^2}\right)$
 & $\widetilde{\Omega}\!\left(\frac{|\mathcal S||\mathcal A|}{(1-\gamma)^3\epsilon^2}\right)$
 & $\widetilde {\mathcal O}\!\left(\frac{\sqrt\delta}{1-\gamma}\right)$\\
$1-\gamma\leq\delta<1$
 & $\widetilde{\mathcal O}\!\left(\frac{|\mathcal S||\mathcal A|\sqrt\delta}{(1-\gamma)^4\epsilon^2}\right)$
 & $\widetilde{\Omega}\!\left(\frac{|\mathcal S||\mathcal A|\delta}{(1-\gamma)^4\epsilon^2}\right)$
 & $\widetilde{\mathcal O}\!\left(\frac1{\sqrt\delta}\right)$\\
$1\le\delta<\log(\frac{ 1}{ p_\wedge})$
 & $\widetilde{\mathcal O}\!\left(\frac{|\mathcal S||\mathcal A| e^\delta}{(1-\gamma)^4\epsilon^2}\right)$
 & $\widetilde{\Omega}\!\left(\frac{|\mathcal S||\mathcal A|e^\delta}{(1-\gamma)^4\epsilon^2}\right)$
 & $\widetilde{\mathcal O}\!\left(1\right)$\\
$\delta\geq\log(\frac{ 1}{ p_\wedge})$
 & $\widetilde{\mathcal O}\!\left(\frac{|\mathcal S||\mathcal A|}{p_\wedge}\right)$
 & $\widetilde{\Omega}\!\left(\frac{|\mathcal S||\mathcal A|}{p_\wedge}\right)$
 & $\widetilde{\mathcal O}(1)$\\
 \bottomrule
 \end{tabular}
 \caption{Upper bounds, lower bounds, and gaps for $p_\wedge\geq1-\gamma$.
The factor $\delta$ is retained in the lower bound even when
$1\leq\delta<\log(1/p_\wedge)$; its size is then at most logarithmic
in $1/p_\wedge$.}
 \label{tab:tightness-large-support}
\end{table}

The analysis above shows that our upper and lower bounds are near minimax optimal when the minimum positive transition probability satisfies $p_\wedge\geq 1-\gamma$ (see Figure \ref{fig:sota} (a)).

\paragraph{Case 2: $(1-\gamma)^2\leq p_\wedge<1-\gamma$.}
The dominant lower bound is the same as in Case~1. When
$\delta\leq p_\wedge$, the exponential branch of the upper bound is
bounded by an absolute constant. For $p_\wedge<\delta<1$, the full
minimum is retained in Table~\ref{tab:tightness-intermediate-support};
the exponential branch need not become inactive immediately above
$p_\wedge$. The rates match for $\delta\leq(1-\gamma)^2$ and in the
worst-path regime. When
$1\leq\delta<\log(1/p_\wedge)$, the ratio of the displayed rate
expressions is $(1-\gamma)/(\delta p_\wedge)$ before the logarithmic
factors in the upper bound are restored.

\begin{table}[H]

 \centering
 \footnotesize
 \setlength{\tabcolsep}{2pt}
 \renewcommand{\arraystretch}{1.5}
 \begin{tabular}{@{}p{0.30\linewidth}p{0.20\linewidth}p{0.20\linewidth}p{0.245\linewidth}@{}}
 \toprule
 Uncertainty level & Upper bound & Lower bound & Gap\\
 \midrule
$0<\delta\leq(1-\gamma)^2$
 & $\widetilde{\mathcal O}\!\left(\frac{|\mathcal S||\mathcal A|}{(1-\gamma)^3\epsilon^2}\right)$
 & $\widetilde{\Omega}\!\left(\frac{|\mathcal S||\mathcal A|}{(1-\gamma)^3\epsilon^2}\right)$
 & $\widetilde {\mathcal O}\!\left(1\right)$\\
$(1-\gamma)^2<\delta\leq p_\wedge $
 & $\widetilde{\mathcal O}\!\left(\frac{|\mathcal S||\mathcal A|\sqrt\delta}{(1-\gamma)^4\epsilon^2}\right)$
 & $\widetilde{\Omega}\!\left(\frac{|\mathcal S||\mathcal A|}{(1-\gamma)^3\epsilon^2}\right)$
 & $\widetilde {\mathcal O}\!\left(\frac{\sqrt\delta}{1-\gamma}\right)$\\ 
$p_\wedge<\delta<1-\gamma$
 & $\widetilde{\mathcal O}\!\left(
 \frac{|\mathcal S||\mathcal A|}{(1-\gamma)^3\epsilon^2} \frac{\sqrt{\delta}}{p_\wedge}\right)$
 & $\widetilde{\Omega}\!\left(\frac{|\mathcal S||\mathcal A|}{(1-\gamma)^3\epsilon^2}\right)$
 & $\mathcal O\!\left(\frac{\sqrt{\delta}}{p_\wedge}\right)$\\
$1-\gamma\leq\delta<1$
 & $\widetilde{\mathcal O}\!\left(
 \frac{|\mathcal S||\mathcal A|}{(1-\gamma)^3\epsilon^2}\frac{\sqrt{\delta}}{p_\wedge}\right)$
 & $\widetilde{\Omega}\!\left(\frac{|\mathcal S||\mathcal A|\delta}{(1-\gamma)^4\epsilon^2}\right)$
 & $\widetilde{\mathcal O}\!\left(\frac{1-\gamma}{\sqrt{\delta}p_\wedge}\right)$\\ 
$ 1\leq \delta<1+\log(\frac{1-\gamma}{p_\wedge}) $
 & $\widetilde{\mathcal O}\!\left(\frac{|\mathcal S||\mathcal A|}{(1-\gamma)^3p_\wedge\epsilon^2}\right)$
 & $\widetilde{\Omega}\!\left(\frac{ e^\delta|\mathcal S||\mathcal A| }{(1-\gamma)^4\epsilon^2}\right)$
 & $\widetilde{\mathcal O}\!\left(\frac{1-\gamma}{ p_\wedge e^\delta} \right)$\\
 $ 1+\log(\frac{1-\gamma}{p_\wedge}) \leq \delta<\log(1/p_\wedge) $
 & $\widetilde{\mathcal O}\!\left(\frac{e^\delta|\mathcal S||\mathcal A|}{(1-\gamma)^4\epsilon^2}\right)$
 & $\widetilde{\Omega}\!\left(\frac{e^\delta|\mathcal S||\mathcal A| }{(1-\gamma)^4\epsilon^2}\right)$
 & $\widetilde{\mathcal O}\!\left(1\right)$\\
$\delta\geq\log(1/p_\wedge)$
 & $\widetilde{\mathcal O}\!\left(\frac{|\mathcal S||\mathcal A|}{p_\wedge}\right)$
 & $\widetilde{\Omega}\!\left(\frac{|\mathcal S||\mathcal A|}{p_\wedge}\right)$
 & $\widetilde{\mathcal O}(1)$\\
 \bottomrule
 \end{tabular}
 \caption{Upper bounds, lower bounds, and gaps for
$(1-\gamma)^2\leq p_\wedge<1-\gamma$. Intermediate uncertainty
levels retain all three branches of the upper bound.}
 \label{tab:tightness-intermediate-support}
\end{table}

\paragraph{Case 3: $p_\wedge<(1-\gamma)^2$.}
The minimum involving $p_\wedge$ controls the upper bound throughout
$0<\delta\le1$. For $\delta\geq1-\gamma$, the dominant lower bound is
now the one proportional to
$|\mathcal S||\mathcal A|\delta/((1-\gamma)^2p_\wedge\epsilon^2)$.
Table~\ref{tab:tightness-small-support} uses this support-dependent
lower bound only on the range established by Theorem~\ref{thm:lb2}.
When $1\leq\delta<\log(1/p_\wedge)$, the ratio of the displayed
rate expressions is $1/(\delta(1-\gamma))$ before restoring
logarithmic factors.

\begin{table}[H]
 \centering
 \footnotesize
 \setlength{\tabcolsep}{2pt}
 \renewcommand{\arraystretch}{1.5}
 \begin{tabular}{@{}p{0.35\linewidth}p{0.20
 \linewidth}p{0.22\linewidth}p{0.22\linewidth}@{}}
 \toprule
 Uncertainty level & Upper bound & Lower bound & Gap\\
 \midrule
$0<\delta\leq p_\wedge$
 & $\widetilde{\mathcal O}\!\left(\frac{|\mathcal S||\mathcal A|}{(1-\gamma)^3\epsilon^2}\right)$
 & $\widetilde{\Omega}\!\left(\frac{|\mathcal S||\mathcal A|}{(1-\gamma)^3\epsilon^2}\right)$
 & $\widetilde{\mathcal O}  (1)$\\
$p_\wedge<\delta<\sqrt{1-\gamma}$
 & $\widetilde{\mathcal O}\!\left(
 \frac{|\mathcal S||\mathcal A|}{(1-\gamma)^3\epsilon^2}\frac{\sqrt{\delta}}{p_\wedge}\right)$
 & $\widetilde{\Omega}\!\left(\frac{|\mathcal S||\mathcal A|}{(1-\gamma)^3\epsilon^2}\right)$
 & $\widetilde{\mathcal O}\!\left(\frac{\sqrt{\delta}}{p_\wedge}\right)$\\ 
$\sqrt{1-\gamma}\leq\delta<1$
 & $\widetilde{\mathcal O}\!\left(
 \frac{|\mathcal S||\mathcal A|}{(1-\gamma)^3\epsilon^2}\frac{\sqrt{\delta}}{p_\wedge}\right)$
 & $\widetilde{\Omega}\!\left(\frac{|\mathcal S||\mathcal A|\delta^2}{(1-\gamma)^2p_\wedge\epsilon^2}\right)$
 & $\widetilde{\mathcal O}\!\left(\frac{1}{\delta\sqrt{\delta}(1-\gamma)}\right)$\\ 
$ 1\leq \delta<1+\log(\frac{(1-\gamma)^2}{p_\wedge}) $
 & $\widetilde{\mathcal O}\!\left(\frac{|\mathcal S||\mathcal A|}{(1-\gamma)^3p_\wedge\epsilon^2}\right)$
 & $\widetilde{\Omega}\!\left(\frac{|\mathcal S||\mathcal A| }{(1-\gamma)^2p_\wedge\epsilon^2}\right)$
 & $\widetilde{\mathcal O}\!\left(\frac{1}{1-\gamma} \right)$\\ 
  $ 1+\log(\frac{(1-\gamma)^2}{p_\wedge})\leq \delta<1+\log(\frac{1-\gamma}{p_\wedge}) $
 & $\widetilde{\mathcal O}\!\left(\frac{ |\mathcal S||\mathcal A|}{(1-\gamma)^3 p_\wedge \epsilon^2}\right)$
 & $\widetilde{\Omega}\!\left(\frac{e^\delta |\mathcal S||\mathcal A| }{(1-\gamma)^4\epsilon^2}\right)$
 & $\widetilde{\mathcal O}\!\left(\frac{1-\gamma}{p_\wedge e^\delta} \right)$\\ 
 
 $ 1+\log(\frac{1-\gamma}{p_\wedge})\leq \delta<\log(1/p_\wedge) $
 & $\widetilde{\mathcal O}\!\left(\frac{ e^\delta|\mathcal S||\mathcal A|}{(1-\gamma)^4 \epsilon^2}\right)$
 & $\widetilde{\Omega}\!\left(\frac{e^\delta|\mathcal S||\mathcal A| }{(1-\gamma)^4\epsilon^2}\right)$
 & $\widetilde{\mathcal O}\!\left(1 \right)$\\ 
$\delta\geq\log(1/p_\wedge)$
 & $\widetilde{\mathcal O}\!\left(\frac{|\mathcal S||\mathcal A|}{p_\wedge}\right)$
 & $\widetilde{\Omega}\!\left(\frac{|\mathcal S||\mathcal A|}{p_\wedge}\right)$
 & $\widetilde{\mathcal O}(1)$\\
 \bottomrule
 \end{tabular}
 \caption{Upper bounds, lower bounds, and gaps for $p_\wedge<(1-\gamma)^2$.
For $\delta<1-\gamma$, the lower-bound entries use only the cited
standard-RL result.}
 \label{tab:tightness-small-support}
\end{table}

The cases above are illustrated in \Cref{fig:sota} (b) and (c).

\section{Conclusion}

In this paper, we study the sample complexity of distributionally robust reinforcement learning under KL-defined uncertainty sets. We develop a comprehensive understanding of this problem with tighter upper bounds and, for the first time, establish minimax lower bounds that hold across the entire range of uncertainty levels. Our results close the gap between upper and lower bounds and recover the minimax-optimal rates of standard reinforcement learning in the vanishing-uncertainty regime. 
A key insight of our analysis is a precise characterization of how uncertainty level, discount factor, and minimum positive transition probability jointly govern the intrinsic difficulty of KL-defined robust RL. In particular, our lower bounds reveal that even under constant uncertainty, small support probabilities can induce large variance, fundamentally limiting learning efficiency. This phenomenon highlights a structural source of hardness that does not arise in standard RL and explains the degradation observed in robust settings.
Beyond KL-defined uncertainty sets, our analytical techniques, especially the transfer of robustness-induced variance into value-function variance and the hard-MDP constructions used for lower bounds, may be of independent interest and applicable to other divergence-based robust optimization and robust RL frameworks. 

% Above all, the main open directions are to move beyond the tabular/generative setting and the $(s,a)$-rectangular KL uncertainty set, e.g., extending these tight tradeoffs to offline data (distribution shift plus robustness), to function approximation, and to $s$-rectangular uncertainty set. 

\section*{Acknowledgements}
The work of Y. Wang, Z. Deng and S. Zou was supported in part by DARPA under Agreement No. D25AP00191, and by the National Science Foundation under Grants ECCS-2438392 (CAREER), ECCS-2622695, and CCF-2438429. The work of Y. Wang was additionally supported by UCF ECE department.
\newpage

\bibliography{camera_ready}
\bibliographystyle{abbrv}

%%%%%%%%%%%%%%%%%%%%%%%%%%%%%%%%%%%%%%%%%%%%%%%%%%%%%%%%%%%%
\newpage
\appendix
\onecolumn
\tableofcontents
\section{Robust RL with Other Uncertainty Sets}
In this section, we further compare our results with other sample-complexity results for robust RL across various uncertainty sets.
\begin{table*}[!h]
\vskip -0.05in
\begin{center}
\begin{small}
\begin{sc}
\begin{tabular}{p{14mm}|p{5mm}|p{40mm}|p{47mm}|p{12mm}}
\toprule
Reference   & Sets &Upper Bound &Lower bound &-  \\
% \midrule
\hline
\citep{wainwright2019variance}\newline\citep{li2020sample}\newline\citep{agarwal2020model}   & -   &$\widetilde{\mathcal{O }}\brac{\frac{|\mathcal{S}||\mathcal{A}|}{(1-\gamma)^3  \epsilon^2} }$&Same&Minimax  \\
\hline
\citep{shi2023curious}   &TV  & $\widetilde{\mathcal{O }}\brac{\frac{|\mathcal{S}||\mathcal{A}|}{ (1-\gamma)^2\epsilon^2} \min\varbrac{\frac{1}{\delta},\frac{1}{1-\gamma}} }$  &Same &Minimax \\
% & &$\widetilde{\mathcal{O }}\brac{\frac{|\mathcal{S}||\mathcal{A}|}{ (1-\gamma)^2\epsilon^2\delta} }$ & &  \\
\hline
\citep{shi2023curious}   &$\chi^2$   &$\widetilde{\mathcal{O }}\brac{\frac{|\mathcal{S}||\mathcal{A}|}{(1-\gamma)^3  \epsilon^2} \left(1+\frac{\max\{\delta, \sqrt{\delta}\}}{1-\gamma}\right) }$&$\widetilde{\Omega}\brac{\frac{|\mathcal{S}||\mathcal{A}|}{(1-\gamma)^3  \epsilon^2} \left(1+\frac{\delta}{1-\gamma}\right) }$&Near-Minimax \\
\hline
\citep{clavier2024near} &$L_p$ & $\widetilde{\mathcal{O }}\brac{\frac{|\mathcal{S}||\mathcal{A}| }{ (1-\gamma)^2 \epsilon^2 \max\{1-\gamma, \delta  \}} }$&Same&Minimax \\
\hline
\citep{shi2022distributionally}& KL & $\widetilde{\mathcal{O }}\brac{\frac{|\mathcal{S}||\mathcal{A}| }{ \delta p_\wedge (1-\gamma)^4 \epsilon^2} }$&$\widetilde{ {\Omega }}\brac{\frac{|\mathcal{S}||\mathcal{A}|}{(1-\gamma)^3  \epsilon^2} }$, $\delta\in (0,1-\gamma)$ &  \\
&&&$\widetilde{ {\Omega }}\brac{\frac{|\mathcal{S}||\mathcal{A}|}{ p_\wedge(1-\gamma)^2  \epsilon^2} }$,$\delta\sim\log(\frac{1}{p_\wedge}) $&\\
\hline
\citep{li2025near}&KL&  $\widetilde{\mathcal{O }}\brac{\frac{|\mathcal{S}||\mathcal{A}| }{p_\wedge (1-\gamma)^4 \epsilon^2} }$ &-&$s$-rec \\
\hline
Our work &KL& $\widetilde{\mathcal{O }}\big(\frac{|\mathcal{S}||\mathcal{A}|}{ (1-\gamma)^3 \epsilon^2}\frac{\min\{p_\wedge e^{\delta/p_\wedge},1\} }{p_\wedge}$  &
$ \widetilde{ {\Omega }}\brac{\frac{|\mathcal{S}||\mathcal{A}|}{ p_\wedge (1-\gamma)^2 \epsilon^2
} },\delta \in(1-\gamma,\log(\frac{1}{p_\wedge}))$
&Near-Minimax  \\
&&$+\frac{|\mathcal{S}||\mathcal{A}|}{ (1-\gamma)^3 \epsilon^2}\frac{e^\delta\min\{\sqrt{\delta},1\}}{1-\gamma} \big)$&$ \widetilde{ {\Omega }}\brac{\frac{\delta e^\delta|\mathcal{S}||\mathcal{A}|}{  (1-\gamma)^4 \epsilon^2
} }, \delta \in(1-\gamma,\log(\frac{1}{p_\wedge}))$&
\\
&& $\widetilde{\mathcal O}\left(\frac{|\mathcal{ S}||\mathcal A|}{p_{\wedge}}\right), \delta\geq \log(\frac{1}{p_\wedge})^{[1]}$& ${\Omega}\left(\frac{|\mathcal{ S}||\mathcal A|}{p_{\wedge}}\right), \delta\geq \log(1/p_{\wedge})^{[1]}$&\\
\bottomrule
\end{tabular}
\end{sc}
\end{small}
\end{center}
% \vskip -0.1in
\caption{Upper and Lower Bounds: standard RL, TV, $\chi^2$, $l_p$ and KL defined uncertainty sets. ${[1]}$: 1) When $ \delta\geq \log(1/p_\wedge)$, we prove that the KL-defined robust RL reduces to a worst-path RL problem and admits minimax-optimal sample complexity of $\widetilde{\Theta}(\frac{|\mathcal S||\mathcal A|}{p_{\wedge}})$ as shown in \cite{deng2025near}. 2) When $\delta\in (0,1-\gamma)$, we apply the lower bound result of $\widetilde{\Omega}\brac{\frac{|\mathcal{S}||\mathcal{A}|}{(1-\gamma)^3  \epsilon^2} }$ established in \cite{shi2022distributionally}.}\label{table:1}
\end{table*}

\section{Limitation}\label{sec:limit}
In this section, we discuss the limitations of our work. This paper is primarily theoretical, and our results provide a comprehensive understanding of KL-based DR-RL. However, further effort is required to develop exact minimax-optimal upper and lower bounds and to conduct simulations that validate these findings.

\section{Proof of \Cref{thm:up}: Upper Bound Analysis}\label{app:upper-analysis}
% BEGIN RED REVISION: reorganized upper-bound argument.

In this section, we prove \Cref{thm:up}.

We first introduce the notation and supporting lemmas in \Cref{sec:c1,sec:c2,sec:c3,sec:c4,sec:c5,sec:c6}, which also provide insight into the structure of DR-RL with KL uncertainty sets. We then use these lemmas to complete the proof in \Cref{sec:c7}.

\subsection{Dual representation and worst-case transitions}\label{sec:c1}
For a value vector $V$, recall the dual objective of strong duality in Lemma \ref{lemma:strongduality}:
\begin{align}\label{eq:20}
 f(P^0_{s,a},V,\lambda)
 =-\lambda\log\mathbb E_{P^0_{s,a}}[e^{-V/\lambda}]
  -\lambda\delta,\qquad \lambda>0.
\end{align}
At $\lambda=0$, $f(P^0_{s,a},V,\lambda)$ is defined by its limit:
$\lim_{\lambda\to 0^+}f(P^0_{s,a},V,\lambda)=\min_{s':P^0(s'|s,a)>0}V(s')$.

For $\delta>0$, denote population dual maximizers ${\lambda}^*$ by $ {\lambda}^*= \arg\max_{\lambda\geq0} f(  P_{s,a}^0,V,\lambda)$. 
Denote by empirical
dual maximizers $\widehat\lambda^*$ by $\widehat{\lambda}^*= \arg\max_{\lambda\geq0} f(\widehat P_{s,a}^0,V,\lambda)$, respectively. Here, we suppress the dependence of $\lambda^*$ and $\widehat{\lambda}^*$ on $s,a,V$  without causing any confusion.

If $\lambda^*>0$, define the weight function $h_{s,a}^{\lambda^*,V}(s')$ as the ratio of the worst-case distribution $\underline P^V (s'|s,a)\in \arg\min_{P_{s,a}\in \mathcal{P}_{s,a}(\delta)}\E_{P_{s,a}}[V]$ to the nominal distribution $P^0 (s'|s,a)$ . The closed-form of the population worst-case distribution $\underline P^V(s'|s,a)$ can then be written as follows from~\cite{hu2013kullback}:
\begin{align}
\label{eq: weight}
 h_{s,a}^{\lambda^*,V}(s')
 =\frac{e^{-V(s')/\lambda^*}}
 {\mathbb E_{P^0_{s,a}}[e^{-V/\lambda^*}]},\quad
 \underline P^V(s'|s,a)
 =P^0(s'|s,a)h_{s,a}^{\lambda^*,V}(s').
\end{align}
If $\widehat\lambda^*>0$, the empirical worst-case distribution $\underline {\widehat P}^V(s'|s,a)\in \arg\min_{P_{s,a}\in \widehat{ \mathcal{P}}_{s,a}(\delta)}\E_{P_{s,a}}[V]$ is defined in the same way as:
\begin{align}
\label{eq:weight2}
 \widehat {h}_{s,a}^{\widehat{\lambda}^*,V}(s')
 =\frac{e^{-V(s')/\widehat \lambda^*}}
 {\mathbb E_{\widehat {P}^0_{s,a}}[e^{-V/\widehat \lambda^*}]},\quad
 \underline{\widehat P}^V(s'|s,a)
 =\widehat {P}^0(s'|s,a)\widehat h_{s,a}^{\widehat{ \lambda}^*,V}(s'),
\end{align}
where the uncertainty set $ \widehat{ \mathcal{P}}_{s,a}$ is centered at empirical distribution $\widehat P^0_{s,a} $.

For given
policy $\pi$, define the transitions
$\underline {P}^{V,\pi}$ and $\underline{\widehat P}^{V,\pi}$ by averaging over $a\sim\pi(\cdot|s)$ as 
\begin{align}
    \underline{P}^{V,\pi}(s'|s)= \sum_{a \in\mathcal{A}} \pi(a|s) \underline{ P}^V(s'|s,a);\quad  \widehat{\underline{P}}^{V,\pi}(s'|s)= \sum_{a \in\mathcal{A}} \pi(a|s) \underline{\widehat P}^V(s'|s,a)
\end{align}

% The first lemma collects the identities that connect the dual optimizer to its worst-case transition row. Normalization makes the tilt a density ratio, while dual stationarity identifies the value of that density at the worst-case mean. These facts will be used when centering the concentration term in the variance-transfer argument.

Next, we introduce a lemma that characterizes the properties of the weight function and the worst-case distributions when $\lambda^*>0 $. 

\begin{lemma}\label{lm:hp}
For $\lambda^*>0 $, the weight function and the worst-case distribution satisfy
\begin{align}
&(a). \quad\mathbb E_{P^0_{s,a}}[h_{s,a}^{\lambda^*,V}]=1,\quad
 & (b).\quad\max_{s'}h_{s,a}^{\lambda^*,V}(s')\leq p_\wedge^{-1},\nonumber\\
 &(c).\quad\operatorname{KL}(\underline P^V_{s,a}\|P^0_{s,a})=\delta,
\quad &(d).\quad\frac{e^{-\overline{V}/ \lambda^*}}
 {\mathbb E_{  {P}^0_{s,a}}[e^{-V/  \lambda^*}]}=e^\delta,
 \label{eq:lm4}
\end{align}
where $\overline V=\mathbb E_{\underline P^V_{s,a}}[V]$.
\end{lemma}
\begin{proof}
First, \ref{eq:lm4}.(a) follows directly from the definition in \eqref{eq: weight}. Specifically, we have
\begin{align}\mathbb E_{P^0_{s,a}}[h_{s,a}^{\lambda^*,V}]=
 \sum_{s':P^0(s'|s,a)>0}P^0(s'|s,a)
 h_{s,a}^{\lambda^*,V}(s')
 =\frac{\sum_{s'}P^0(s'|s,a)e^{-V(s')/\lambda^*}}
 {\mathbb E_{P^0_{s,a}}\Fbrac{e^{-V/\lambda^*}}}=1.
\end{align}

Second, since $P^0(s'|s,a)h_{s,a}^{\lambda^*,V}(s')\leq1$, therefore $h_{s,a}^{\lambda^*,V}(s')\leq 1/P^0(s'|s,a)\leq p_\wedge^{-1}$, which completes the proof of \ref{eq:lm4}.(b).

Third, we provide the proof of \ref{eq:lm4}.(c). 
To differentiate the dual objective $ f(P^0_{s,a},V,\lambda)$, first note
that, for $\lambda>0$, 
\begin{align}
 \frac{d}{d\lambda}\mathbb E_{P^0_{s,a}}[e^{-V/\lambda}]
 =\frac{1}{\lambda^2}
   \mathbb E_{P^0_{s,a}}[Ve^{-V/\lambda}].
\end{align}
The product rule then gives
\begin{align}
 \partial_\lambda f(P^0_{s,a},V,\lambda)
 &=-\log\mathbb E_{P^0_{s,a}}[e^{-V/\lambda}]
   -\frac{\mathbb E_{P^0_{s,a}}[Ve^{-V/\lambda}]}
          {\lambda\mathbb E_{P^0_{s,a}}[e^{-V/\lambda}]}
   -\delta.
\end{align}
Consequently, by Karush–Kuhn–Tucker (KKT) conditions, differentiating the dual objective gives
\begin{align}
 \partial_\lambda f(P^0_{s,a},V,\lambda^*)=0,
\end{align}
at an interior maximizer $\lambda^*>0$. Therefore, we can conclude that
\begin{align}\label{eq:19}
    \delta= -\log\mathbb E_{P^0_{s,a}}[e^{-V/\lambda^*}]
   -\frac{\mathbb E_{P^0_{s,a}}[Ve^{-V/\lambda^*}]}
          {\lambda^*\mathbb E_{P^0_{s,a}}[e^{-V/\lambda^*}]}
\end{align}

Moreover, using $\log \brac{ h_{s,a}^{\lambda^*,V}}=-V/\lambda^*-\log\mathbb E_{P^0_{s,a}}[e^{-V/\lambda^*}]$, the \ref{eq:lm4}.(c) is equal to 
\begin{align}
    \operatorname{KL}(\underline P^V_{s,a}\|P^0_{s,a})
    &= \E_{\underline P^V_{s,a}}\Fbrac{ \log \brac{ h_{s,a}^{\lambda^*,V}}}
    \\& = 
   \E_{\underline P^V_{s,a}}\Fbrac{-\frac{V}{\lambda^*}}
          -\log\mathbb E_{P^0_{s,a}}[e^{-V/\lambda^*}]
          \\& = \frac{\mathbb E_{P^0_{s,a}}[-V/\lambda^* \cdot e^{-V/\lambda^*}]}
          {\mathbb E_{P^0_{s,a}}[e^{-V/\lambda^*}]}-\log\mathbb E_{P^0_{s,a}}[e^{-V/\lambda^*}]=\delta.
\end{align}

This completes the proof of \ref{eq:lm4}.(c).

Last, by \eqref{eq: weight}, we have
\begin{align}
\overline V
=\mathbb E_{\underline P^V_{s,a}}[V]
=\frac{\mathbb E_{P^0_{s,a}}[Ve^{-V/\lambda^*}]}
       {\mathbb E_{P^0_{s,a}}[e^{-V/\lambda^*}]}.
\end{align}
Substituting this identity into \eqref{eq:19} and rearranging gives
\begin{align}
-\log\mathbb E_{P^0_{s,a}}[e^{-V/\lambda^*}]
=\delta+\frac{\overline V}{\lambda^*}.
\end{align}
Exponentiating both sides and multiplying by
$e^{-\overline V/\lambda^*}$ yields
\begin{align}
\frac{e^{-\overline V/\lambda^*}}
     {\mathbb E_{P^0_{s,a}}[e^{-V/\lambda^*}]}
=e^\delta,
\end{align}
which proves \ref{eq:lm4}.(d).\end{proof}
% The above proof also implies that
% \begin{align}
%  -\log\brac{\mathbb E_{P^0_{s,a}}[e^{-V/\lambda^*}]}
%  &=\delta+\frac{\overline V}{\lambda^*}.
 % \\
 % \log h_{s,a}^{\lambda^*,V}(s')
 % &=\delta-\frac{V(s')-\overline V}{\lambda^*}.
% \end{align}
% Taking the expectation of the second identity under
% $\underline P^V_{s,a}$ yields
% \begin{align}
%  \operatorname{KL}(\underline P^V_{s,a}\|P^0_{s,a})
%  &=\sum_{s'}\underline P^V(s'|s,a)
%        \log h_{s,a}^{\lambda^*,V}(s')\\
%  &=\delta-\frac{\mathbb E_{\underline P^V_{s,a}}[V]-\overline V}
%                        {\lambda^*}=\delta.
% \end{align}
% Setting the real argument of $h$ equal to $\overline V$ instead gives
% $h(\overline V)=e^\delta$.
% These calculations also show why the sign
% in the exponential tilt is negative: smaller successor values receive
% larger density ratios.

% Therefore, exponentiation both sides yields the \eqref{eq:lm4}.(d):
% \begin{align}
%     \frac{e^{-\overline{V}/ \lambda^*}}
%  {\mathbb E_{  {P}^0_{s,a}}[e^{-V/  \lambda^*}]}=e^\delta.
% \end{align}

\subsection{Error decomposition: statistical error and planning error}\label{sec:c2}
Let $\widehat\pi^*$ be the optimal policy for the empirical robust MDP defined by $\widehat{\mathcal{P}}$, and let
$\pi_T$ be the policy returned by robust value iteration. Denote by $\widehat V_\pi$ the robust value function of policy $\pi$ under the empirical uncertainty set $\widehat{\mathcal{P}}$.  Write $\widehat V^*=\widehat V_{\widehat\pi^*}$, and let
$\widehat Q^*$ denote the optimal robust $Q$-function defined by $\widehat{\mathcal{P}}$.
We then decompose the error as follows and discuss each term one by one:
% Empirical optimality implies the following inequality:
\begin{align}\label{eq:errordecomp}
    V_{\pi^*} - {V}_{{\pi}_T}\nonumber
= &\underbrace{\big(V_{\pi^*} - \widehat{V}_{\pi^*}\big)}_{\text{Term I: statistical error under $\pi^*$}}
+ \underbrace{\big(\widehat{V}_{\pi^*} - \widehat V_{\widehat{\pi}^*}\big)}_{\text{Term II: optimality gap}}
\\&+ \underbrace{\big(\widehat V_{\widehat{\pi}^*} - \widehat{V}_{{\pi}_T}\big)}_{\text{Term III: planning error }}
+ \underbrace{\big(\widehat{V}_{{\pi}_T} - {V}_{{\pi}_T}\big)}_{\text{Term IV:statistical error under $\pi_T$}}.
\end{align}
Term II is componentwise nonpositive by the optimality of \(\widehat\pi^*\) in the empirical robust MDP. Term III is the planning error and can be bounded using the contraction property of the empirical Bellman operator. Terms I and IV compare population and empirical robust values under \(\pi^*\) and \(\pi_T\), respectively, which are referred to as the statistical error.

% The term III is non-positive since $\widehat \pi^*$ is the empirical optimal policy. Term II can be bounded using the $\gamma$-contraction property. The main challenge is to bound terms I and IV, which arise from estimation errors under the same policy.

First, we bound term III. By the $\gamma$-contraction property of the robust Bellman optimality operator and the initialization $Q_0=0$, we have $\|Q_T-\widehat Q^*\|_\infty\leq\gamma^T/(1-\gamma)$. % Indeed, a policy greedy with respect to $Q_T$ has a one-step suboptimality of at most $2\|Q_T-\widehat Q^*\|_\infty$. The contraction of the policy Bellman operator then yields the factor $(1-\gamma)^{-1}$.
Recall that $\widehat{\mathcal T}$ is the empirical robust Bellman optimality operator:
\begin{align}
(\widehat{\mathcal T}V)(s)
=\max_{a\in\mathcal A}
\left[
r(s,a)+\gamma
\inf_{q\in\widehat{\mathcal P}_{s,a}(\delta)}
\mathbb E_{s'\sim q}[V(s')]
\right].
\end{align}
Denote by $\widehat{\mathcal T}^{\pi_T}$ the empirical robust Bellman operator for a policy $\pi_T$:
\begin{equation}\label{eq:empiricalrobustBellmanoperator}
(\widehat{\mathcal T}^{\pi_T}V)(s)
=
\sum_{a\in\mathcal A}\pi_T(a\mid s)
\left[
r(s,a)
+\gamma\inf_{q\in\widehat{\mathcal P}_{s,a}(\delta)}
\mathbb E_{s'\sim q}[V(s')]
\right],
\qquad s\in\mathcal S.
\end{equation}

% and $\widehat{\mathcal T}^{\pi_T}$ and policy Bellman operators. 

Since $\pi_T$ is greedy with respect to $Q_T$, we have that
\begin{align}
0\leq\widehat{\mathcal T}\widehat V^*
-\widehat{\mathcal T}^{\pi_T}\widehat V^*
\leq2\|Q_T-\widehat Q^*\|_\infty\mathbf1.
\end{align}
Using the Bellman fixed-point equations and the $\gamma$-contraction property gives
\begin{align}
\|\widehat V^*-\widehat V_{\pi_T}\|_\infty
\leq2\|Q_T-\widehat Q^*\|_\infty
+\gamma\|\widehat V^*-\widehat V_{\pi_T}\|_\infty.
\end{align}
Rearranging and substituting the value-iteration error bound, we obtain that 
\begin{equation}\label{eq:polished-planning}
 \|\widehat V_{\widehat\pi^*}-\widehat V_{\pi_T}\|_\infty
 \leq\frac{2\gamma^T}{(1-\gamma)^2}.
\end{equation}

Therefore, using \eqref{eq:polished-planning} and the fact that term II is nonpositive, the error decomposition in \eqref{eq:errordecomp} can be bounded as follows:
\begin{align}
 \|V_{\pi^*}-V_{\pi_T}\|_\infty
 &\leq\|V_{\pi^*}-\widehat V_{\pi^*}\|_\infty
      +\|\widehat V_{\pi_T}-V_{\pi_T}\|_\infty+\|\widehat V_{\widehat\pi^*}-\widehat V_{\pi_T}\|_\infty
      \\& \leq \underbrace{\|V_{\pi^*}-\widehat V_{\pi^*}\|_\infty}_{\text{term I}}
      +\underbrace{\|\widehat V_{\pi_T}-V_{\pi_T}\|_\infty}_{\text{term IV}}+ \frac{2\gamma^T}{(1-\gamma)^2}.
\end{align}

For a fixed policy \(\pi\), the robust Bellman equations yield the following bound on the value-function statistical error:
\begin{align}
 \|V_\pi-\widehat V_\pi\|_\infty
 \leq\gamma\max_{W\in\{V_\pi,\widehat V_\pi\}}
 \Big\|&(I-\gamma\underline{\widehat P}^{W,\pi})^{-1}\left|\underline{\widehat P}^{V_\pi,\pi}V_\pi
       -\underline P^{V_\pi,\pi}V_\pi\right|\Big\|_\infty.
 \label{eq:polished-propagation}
\end{align}
Absolute value above is taken componentwise. %Evaluating each empirical robust expectation at the other value function's minimizing row bounds $\widehat V_\pi-V_\pi$ between the two resolvents applied to the same one-step error. Since both resolvents have nonnegative entries, \eqref{eq:polished-propagation} follows.

Recall the definition of empirical robust Bellman operator for a policy $\pi$ in \eqref{eq:empiricalrobustBellmanoperator}. Similarly, we can also define the robust Bellman operator for a policy $\pi$ using the true uncertainty set $\mathcal P$, denoted by $\mathcal T^\pi$:
\begin{equation}\label{eq:robustBellmanoperator}
({\mathcal T}^{\pi}V)(s)
=
\sum_{a\in\mathcal A}\pi(a\mid s)
\left[
r(s,a)
+\gamma\inf_{q\in{\mathcal P}_{s,a}(\delta)}
\mathbb E_{s'\sim q}[V(s')]
\right],
\qquad s\in\mathcal S.
\end{equation}

To show \eqref{eq:polished-propagation}, we first note that $\widehat V_\pi$ and $V_\pi$ are fixed-points of robust Bellman operators $\widehat{\mathcal T}^{\pi}$ and ${\mathcal T}^{\pi}$: $\widehat V_\pi=\widehat{\mathcal T}^{\pi}\widehat V_\pi$ and $V_\pi={\mathcal T}^{\pi}V_\pi$. Applying these, we obtain that
% 
% apply two robust Bellman equations:
% Here is the full Bellman comparison behind this bracketing argument.
% The two Bellman equations first imply
\begin{align}\label{eq:46}
 \widehat V_\pi-V_\pi
 &=\gamma\big(
   \underline{\widehat P}^{\widehat V_\pi,\pi}\widehat V_\pi
   -\underline{\widehat P}^{V_\pi,\pi}V_\pi\big)+\gamma\big(
   \underline{\widehat P}^{V_\pi,\pi}V_\pi
   -\underline P^{V_\pi,\pi}V_\pi\big).
\end{align}
For each state-action, since $\underline{\widehat P}^{\widehat V_\pi,\pi}$ corresponds to the worst-case transition kernel with respect to $\widehat V_\pi$, we have that
\begin{align}\label{eq:47}
 \underline{\widehat P}^{\widehat V_\pi,\pi}\widehat V_\pi
 \leq\underline{\widehat P}^{V_\pi,\pi}\widehat V_\pi.
\end{align}
Similarly,  for $V_\pi$, we have that
\begin{align}\label{eq:48}
 \underline{\widehat P}^{V_\pi,\pi}V_\pi
 \leq\underline{\widehat P}^{\widehat V_\pi,\pi}V_\pi.
\end{align}
% These inequalities remain valid after averaging over the fixed policy.
Substituting \eqref{eq:48} and \eqref{eq:47} into \eqref{eq:46}, respectively and  rearranging yields 
\begin{align}
 (I-\gamma\underline{\widehat P}^{\widehat V_\pi,\pi})
       (\widehat V_\pi-V_\pi)
 &\geq\gamma\big(
   \underline{\widehat P}^{V_\pi,\pi}V_\pi
   -\underline P^{V_\pi,\pi}V_\pi\big),\\
 (I-\gamma\underline{\widehat P}^{V_\pi,\pi})
       (\widehat V_\pi-V_\pi)
 &\leq\gamma\big(
   \underline{\widehat P}^{V_\pi,\pi}V_\pi
   -\underline P^{V_\pi,\pi}V_\pi\big).
\end{align}

For every stochastic matrix $P$,
$(I-\gamma P)^{-1}=\sum_{t=0}^\infty\gamma^tP^t$ has nonnegative
entries. Multiplying the two inequalities by their respective inverses
preserves inequality directions. Finally,
$|(I-\gamma P)^{-1}x|\leq(I-\gamma P)^{-1}|x|$ componentwise.
The positive and negative parts of $\widehat V_\pi-V_\pi$ are thus
bounded by the two quantities in~\eqref{eq:polished-propagation}.

\subsection{Support retention and a stable dual comparison}\label{sec:c3}

% In this subsection, we first ensure that sampling retains every nominal successor and a fixed fraction of its probability. The following event provides this control simultaneously for all transition rows.

We first establish a high-probability lower bound on the empirical
transition probabilities. With sufficiently many samples, each positive
nominal transition probability is estimated to be at least three quarters
of its true value, simultaneously for all state-action pairs and
successor states. Consequently, the empirical and nominal transition
kernels have the same support. This lower bound will be used to compare
the expectations appearing in the two dual objectives.
\begin{lemma}[Uniform lower bound on empirical transition probabilities]\label{lem:polished-retention}
If
$
 N\geq\frac{32}{p_\wedge}
       \log\frac{|\mathcal S|^2|\mathcal A|}{\eta},
$
then, with probability at least $1-\eta$,
\begin{align}\label{eq:51}
 \widehat P^0(s'|s,a)\geq\tfrac34P^0(s'|s,a)
 \quad\forall(s,a,s')\text{ with }P^0(s'|s,a)>0.
\end{align}
Moreover, the empirical and nominal transition distributions share the same support for every state-action pair $(s,a)$.

\end{lemma}

\begin{proof}
For any $(s,a,s')$ such that $P^0(s'|s,a)>0$, $N\widehat P^0(s'|s,a)$ is a binomial random variable with
mean $NP^0(s'|s,a)$. The multiplicative Chernoff inequality gives
\begin{align}
 \Pr\!\left(\widehat P^0(s'|s,a)<\tfrac34P^0(s'|s,a)\right)
 \leq e^{-NP^0(s'|s,a)/32}\leq e^{-Np_\wedge/32}.
\end{align}
A union bound over at most $|\mathcal S|^2|\mathcal A|$ entries proves \eqref{eq:51}. 

Sampling cannot create successors outside the nominal support, and the event prevents any nominal successor from disappearing, which proves the last claim in the lemma.
% 
% In particular, if the row samples are $s'_1,\ldots,s'_N$, then
% \begin{align}
%  N\widehat P^0(s'|s,a)
%  =\sum_{i=1}^N\mathbf1\{s'_i=s'\},
%  \qquad
%  \mathbb E\!\left[\sum_{i=1}^N\mathbf1\{s'_i=s'\}\right]
%  =NP^0(s'|s,a).
% \end{align}
% The lower-tail form of Chernoff, with relative deviation $1/4$,
% has exponent
% \begin{align}
%  -(1/4)^2NP^0(s'|s,a)/2=-NP^0(s'|s,a)/32.
% \end{align}
% Consequently, the probability that any positive entry fails is at most
% \begin{align}
%  \sum_{\substack{s,a,s':\\P^0(s'|s,a)>0}}
%        e^{-NP^0(s'|s,a)/32}
%  \leq|\mathcal S|^2|\mathcal A|e^{-Np_\wedge/32}
%  \leq\eta.
% \end{align}
% No independent sampling assumption between different rows is needed
% for this union bound; the binomial calculation uses independence within
% each sampled row. A zero-probability successor has empirical count zero
% almost surely, completing both directions of the support statement.
\end{proof}

Conditioned on the high probability event in Lemma \ref{lem:polished-retention}, we next bound the statistical errors in \eqref{eq:errordecomp}. Denote by $
h_{s,a}^{\lambda,V}(s')
:=\frac{e^{-V(s')/\lambda}}{\mathbb E_{P^0_{s,a}}[e^{-V/\lambda}]}, \forall \lambda>0.
$
% Evaluating each objective at the other problem’s optimizer reduces the Bellman error to empirical averages of the normalized exponential tilt.
% Here, we define the weight function by extending the function $h_{s,a}^{\lambda,V}(v) = \frac{e^{-\frac{v}{\lambda}} }{\E_{P^0_{s,a}}} $

\begin{lemma}\label{lm:5.4}
Conditioned on the event in Lemma~\ref{lem:polished-retention}, for any value
vector $V$, we have
\begin{align}
 \left|\inf_{P_{s,a}\in\widehat{\mathcal P}_{s,a}}
               \mathbb E_{P_{s,a}}[V]
       -\inf_{P_{s,a}\in\mathcal P_{s,a}}
               \mathbb E_{P_{s,a}}[V]\right| 
 &\quad\leq\max_{\lambda\in\{\lambda^*,\widehat\lambda^*\}}
 \lambda\left|\log
 \frac{\mathbb E_{\widehat P^0_{s,a}}[e^{-V/\lambda}]}
      {\mathbb E_{P^0_{s,a}}[e^{-V/\lambda}]}\right|\nonumber\\
 &\quad\leq\frac43
 \max_{\lambda\in\{\lambda^*,\widehat\lambda^*\}}
 \lambda\left|(\widehat P^0_{s,a}-P^0_{s,a})
                  h_{s,a}^{\lambda,V}\right|.
 \label{eq:bound}
\end{align}
If either of the dual maximizers $\lambda^*,\widehat\lambda^*$ equals zero, the corresponding terms on the right-hand side are interpreted by continuous extension and equal zero.
\end{lemma}
\begin{proof}
We first consider $\lambda^*,\widehat\lambda^*> 0$. By the definition of $\lambda^*,\widehat\lambda^*$, evaluating the dual objective $f$ in \eqref{eq:20} at $\widehat P^0_{s,a}, P^0_{s,a}$ with $\lambda^*,\widehat\lambda^*$ for the same $V$ gives
\begin{align}\label{eq:54}
 f(\widehat P^0_{s,a},V,\lambda^*)-f(P^0_{s,a},V,\lambda^*)
 &\leq f(\widehat P^0_{s,a},V,\widehat\lambda^*)
       -f(P^0_{s,a},V,\widehat\lambda^*)\nonumber\\
       &=
-\widehat\lambda^*\log
\frac{\mathbb E_{\widehat P^0_{s,a}}[e^{-V/\widehat\lambda^*}]}
     {\mathbb E_{P^0_{s,a}}[e^{-V/\widehat\lambda^*}]}.
\end{align}
Similarly, we have that
\begin{align}\label{eq:55}
 f(P^0_{s,a},V,\lambda^*)
       -f(\widehat P^0_{s,a},V,\widehat\lambda^*)
 &\leq f(P^0_{s,a},V,\lambda^*)
       -f(\widehat P^0_{s,a},V,\lambda^*)\nonumber\\
 &=\lambda^*\log
     \frac{\mathbb E_{\widehat P^0_{s,a}}[e^{-V/\lambda^*}]}
          {\mathbb E_{P^0_{s,a}}[e^{-V/\lambda^*}]}.
\end{align} 
Combining \eqref{eq:54} and \eqref{eq:55} proves the first inequality in \eqref{eq:bound}. 

% This proves the first inequality in \eqref{eq:bound}. 

Conditioned on the event in Lemma \ref{lem:polished-retention}, it follows that
$\mathbb E_{\widehat P^0_{s,a}}[Z]\geq(3/4)\mathbb E_{P^0_{s,a}}[Z]$
for any nonnegative $Z$. Then, if $x\geq \frac{3}{4} y >0$, it holds that $\min\{x,y\} \geq \frac{3}{4} y $.
We can conclude that
$|\log x-\log y|\leq|x-y|/\min\{x,y\}\leq \frac{4}{3}|x-y|/|x| $. Applying this on the right hand side of the first inequality in \eqref{eq:bound} shows the second inequality.

If $\lambda^*=\widehat\lambda^*=0$, 
it holds that
\begin{align}
    f(\widehat P^0_{s,a},V,\widehat\lambda^*)
                      -f(P^0_{s,a},V,0)= \min_{s'}V(s')- \min_{s'}V(s')=0.
\end{align}

If $\lambda^*=0<\widehat\lambda^*$, then, it holds that
\begin{align}
 0&\leq f(\widehat P^0_{s,a},V,\widehat\lambda^*)
                      -f(P^0_{s,a},V,0)\\
 &\leq f(\widehat P^0_{s,a},V,\widehat\lambda^*)
                      -f(P^0_{s,a},V,\widehat\lambda^*)\\
 &=-\widehat\lambda^*\log
     \frac{\mathbb E_{\widehat P^0_{s,a}}[e^{-V/\widehat\lambda^*}]}
          {\mathbb E_{P^0_{s,a}}[e^{-V/\widehat\lambda^*}]}\\
 &\leq\frac43\widehat\lambda^*
       \left|(\widehat P^0_{s,a}-P^0_{s,a})
                         h_{s,a}^{\widehat\lambda^*,V}\right|.
\end{align}
% The first difference is the difference between the empirical and
% nominal robust expectations. It is therefore controlled by the
% term at $\widehat\lambda^*>0$ in~\eqref{eq:bound}.

If $\widehat\lambda^*=0<\lambda^*$, the same argument gives
\begin{align}
 0&\leq f(P^0_{s,a},V,\lambda^*)
                      -f(\widehat P^0_{s,a},V,0)\\
 &\leq f(P^0_{s,a},V,\lambda^*)
                      -f(\widehat P^0_{s,a},V,\lambda^*)\\
 &=\lambda^*\log
     \frac{\mathbb E_{\widehat P^0_{s,a}}[e^{-V/\lambda^*}]}
          {\mathbb E_{P^0_{s,a}}[e^{-V/\lambda^*}]}\\
 &\leq\frac43\lambda^*
       \left|(\widehat P^0_{s,a}-P^0_{s,a})
                         h_{s,a}^{\lambda^*,V}\right|.
\end{align}
% Here the term at $\lambda^*>0$ controls the difference between
% the nominal and empirical robust expectations.

This completes the proof.
\end{proof}

\subsection{Concentration results}\label{sec:c4}

% The dual comparison has reduced the one-step estimation error to a difference of sample averages.
% We first record the absolute and the two relative forms of Bernstein’s inequality, because the dual
% expressions can involve either nominal or empirical normalizers. All three forms below follow from
% one concentration event for a function fixed independently of the row 

To bound the statistical error terms in \eqref{eq:errordecomp},
Lemma~\ref{lm:5.4} reduces the comparison of empirical and nominal
robust expectations to controlling the difference
$(\widehat P^0_{s,a}-P^0_{s,a})h$, where $h$ is the corresponding
weight vector. In this subsection, we use Bernstein's inequality
to derive concentration bounds for this difference. We first establish
absolute and relative bounds for a vector $h$ fixed independently
of the transition samples.

\begin{lemma}[Absolute and relative Bernstein bounds]\label{lem:bern-three-forms}
Let $X_1,\ldots,X_n$ be $n$ independent samples from a discrete distribution $P\in\Delta(\mathcal S)$, and let
$\widehat P$ be their empirical distribution. Fix a function $g:\mathcal S\rightarrow \mathbb R$
independently of these samples, with $0\leq g\leq B_g$, and suppose
$\mu_g=\mathbb E_P[g]>0$. Then, with probability at least $1-\eta$,
\begin{align}
 |(\widehat P-P)g|
 &\leq\sqrt{\frac{2\operatorname{Var}_P(g)\log(2/\eta)}{n}}
       +\frac{2B_g\log(2/\eta)}{3n},
 \label{eq:bern-general-absolute}\\
 \frac{|(\widehat P-P)g|}{\mathbb E_P[g]}
 &\leq\sqrt{\frac{2\operatorname{Var}_P(g)\log(2/\eta)}
                       {n(\mathbb E_P[g])^2}}
       +\frac{2B_g\log(2/\eta)}{3n\mathbb E_P[g]}.
 \label{eq:bern-rel-true}
\end{align}
Conditioned on the above high probability event, if the right-hand side of
\eqref{eq:bern-general-absolute} is strictly smaller than
$\mathbb E_P[g]$, then
\begin{equation}\label{eq:bern-rel-emp}
 \frac{|(\widehat P-P)g|}{\mathbb E_{\widehat P}[g]}
 \leq
 \frac{\displaystyle
       \sqrt{\frac{2\operatorname{Var}_P(g)\log(2/\eta)}{n}}
       +\frac{2B_g\log(2/\eta)}{3n}}
      {\displaystyle\mathbb E_P[g]
       -\sqrt{\frac{2\operatorname{Var}_P(g)\log(2/\eta)}{n}}
       -\frac{2B_g\log(2/\eta)}{3n}}.
\end{equation}
In particular, if the right-hand side of
\eqref{eq:bern-rel-true} is at most $1/2$, the empirical-relative
error is at most twice that right-hand side of \eqref{eq:bern-rel-true}.
\end{lemma}
\begin{proof}
Set $Y_i=g(X_i)-\mu_g$, and therefore $Y_i$'s are also independent and
\begin{align}
 \mathbb E [Y_i]=0,\qquad
 \mathbb E [Y_i^2]=\operatorname{Var}_P(g),\qquad |Y_i|\leq B_g.
\end{align}
% Tge last inequality uses nonnegativity: botg $g(X_i)$ and $\mu_g$
% belong to $[0,B_g]$. If only $|g|\leq B_g$ is assumed, tge centered
% envelope is $2B_g$, and tge linear remainder below must be enlarged
% accordingly.
The two-sided Bernstein inequality therefore states that, for $t>0$,
\begin{align}
 \Pr\!\left(\left|\frac1n\sum_{i=1}^nY_i\right|>t\right)
 \leq2\exp\!\left(
 -\frac{nt^2}{2\operatorname{Var}_P(g)+(2/3)B_gt}\right).
\end{align}
For $x>0$, the positive solution of
$nt^2-(2/3)B_gxt-2\operatorname{Var}_P(g)x=0$ is
\begin{align}
 \frac{B_gx}{3n}
 +\sqrt{\frac{2\operatorname{Var}_P(g)x}{n}
                   +\frac{B_g^2x^2}{9n^2}}
 \leq\sqrt{\frac{2\operatorname{Var}_P(g)x}{n}}
      +\frac{2B_gx}{3n}.
\end{align}
Taking $x=\log(2/\eta)$ proves
\eqref{eq:bern-general-absolute}. Further dividing by the positive
$\mu_g$ proves~\eqref{eq:bern-rel-true}.

For the empirical denominator, the same event gives
\begin{align}
 \mathbb E_{\widehat P}[g]
 &\geq\mathbb E_P[g]-|(\widehat P-P)g|\\
 &\geq\mathbb E_P[g]
       -\sqrt{\frac{2\operatorname{Var}_P(g)\log(2/\eta)}{n}}
       -\frac{2B_g\log(2/\eta)}{3n}>0.
\end{align}
We then divide \eqref{eq:bern-general-absolute} by this lower bound to obtain
\eqref{eq:bern-rel-emp}. 

If the nominal-relative upper bound is at
most $1/2$, the empirical mean is at least $\mu_g/2$, so replacing the
denominator by $\mu_g/2$ proves the final assertion. All three bounds follow from the same concentration event,
which holds with probability at least $1-\eta$.
\end{proof}

% We now specialize the concentration argument to the exponential
% tilt. Its nominal mean equals one, so centering the random variable
% as $\lambda(h-1)$ leaves the estimation error unchanged. The
% derivative of the tilt then bounds its centered range by
% $\operatorname{span}(V)/p_\wedge$, yielding the following remainder
% without an additional factor of $\lambda$.

We next apply Bernstein's inequality to the weight function
$h_{s,a}^{\lambda,V}$ for fixed $V$ and $\lambda>0$.
Since $\mathbb E_{P^0_{s,a}}[h_{s,a}^{\lambda,V}]=1$,
we consider the centered random variable
$\lambda(h_{s,a}^{\lambda,V}-1)$, whose empirical mean can be written as
$\lambda(\widehat P^0_{s,a}-P^0_{s,a})h_{s,a}^{\lambda,V}$.
We show that its absolute value is bounded by
$\operatorname{span}(V)/p_\wedge$ on the support of $P^0_{s,a}$.
This yields a Bernstein bound whose $1/N$ remainder has no
additional factor of $\lambda$.

\begin{lemma}\label{lm:bern}
Fix a value vector $V$ and a dual parameter $\lambda>0$, both chosen
independently of the samples drawn from $P^0_{s,a}$. With probability at least $1-\eta$,  we have
\begin{align}
 &\lambda\left|(\widehat P^0_{s,a}-P^0_{s,a})
                    h_{s,a}^{\lambda,V}\right|\leq\lambda\sqrt{
 \frac{2\operatorname{Var}_{P^0_{s,a}}(h_{s,a}^{\lambda,V})
                    \log(2/\eta)}{N}}
 +\frac{2\operatorname{span}(V)\log(2/\eta)}{3p_\wedge N},
 \label{eq:bern-abs}
\end{align}
where $\operatorname{span}(V)=\max_s V(s)-\min_s V(s)$.
\end{lemma}

\begin{proof}
% The random variable $\lambda(h_{s,a}^{\lambda,V}(s')-1)$ with $s'\sim P_{s,a}^0$ has mean
% zero. 
Throughout this proof, $\lambda>0$, $V$,  and $(s,a)$ are fixed.
Suppressing the dependence on these quantities, we define the scalar
function $h:\mathbb R\to\mathbb R$ by
\begin{align}
    h(v)=\frac{e^{-v/\lambda}}{\E_{P^0_{s,a}}\Fbrac{e^{-V/\lambda}}}, v\in\mathbb R.
\end{align}

Let $V_c=-\lambda\log\mathbb E_{P^0_{s,a}}[e^{-V/\lambda}]$.
Then $\min V\leq V_c\leq\max V$ and $h(V_c)=1$. 

Since $h'(v)=-h(v)/\lambda$, the function $h$ is decreasing.
For any $s'$ with $P^0(s'\mid s,a)>0$, the mean-value theorem
gives a point $\xi$ between $V(s')$ and $V_c$ such that
\begin{align}
h(V(s'))-1
=h(V(s'))-h(V_c)
=h'(\xi)(V(s')-V_c).
\end{align}
Consequently,
\begin{align}
\lambda|h(V(s'))-1|
=h(\xi)|V(s')-V_c|.
\end{align}
Because $h$ is decreasing and $h(V_c)=1$,
\begin{align}
h(\xi)\leq\max\{h(V(s')),h(V_c)\}
=\max\{h(V(s')),1\}.
\end{align}
Together with $|V(s')-V_c|\leq\operatorname{span}(V)$,
this gives
\begin{align}
\lambda|h(V(s'))-1|
\leq \operatorname{span}(V)\max\{h(V(s')),1\}.
\end{align}
Moreover, the denominator in the definition of $h$ satisfies
\begin{align}
\mathbb E_{P^0_{s,a}}[e^{-V/\lambda}]
\geq P^0(s'\mid s,a)e^{-V(s')/\lambda}.
\end{align}
Hence,
\begin{align}
h(V(s'))
\leq \frac{1}{P^0(s'\mid s,a)}
\leq \frac{1}{p_\wedge}.
\end{align}
Since $1\leq p_\wedge^{-1}$, we conclude that
\begin{align}\label{eq:81}
\lambda|h(V(s'))-1|
\leq \frac{\operatorname{span}(V)}{p_\wedge}
\end{align}
for every $s'$ in the support of $P^0_{s,a}$.

Let $s'_1,\ldots,s'_N$ be the independent samples drawn from
$P^0_{s,a}$. Using
$h(V(s'))=h_{s,a}^{\lambda,V}(s')$, define
\begin{align}
Y_i=\lambda\bigl(h(V(s'_i))-1\bigr)
=\lambda\bigl(h_{s,a}^{\lambda,V}(s'_i)-1\bigr).
\end{align}
These variables are independent and satisfy
\begin{align}
\mathbb E[Y_i]&=0, \nonumber\\
\operatorname{Var}(Y_i)
&=\lambda^2
  \operatorname{Var}_{P^0_{s,a}}(h_{s,a}^{\lambda,V}), \nonumber\\
|Y_i|&\leq\frac{\operatorname{span}(V)}{p_\wedge}.
\end{align}
Furthermore, since
$\mathbb E_{P^0_{s,a}}[h_{s,a}^{\lambda,V}]=1$,
\begin{align}
\frac1N\sum_{i=1}^N Y_i
=\lambda(\widehat P^0_{s,a}-P^0_{s,a})h_{s,a}^{\lambda,V}.
\end{align}
Applying the two-sided Bernstein inequality to this average
with failure probability $\eta$ proves \eqref{eq:bern-abs}.
\end{proof}

Lemma~\ref{lm:bern} requires both the value vector $V$ and the
dual parameter $\lambda>0$ to be chosen independently of the
samples drawn from $P^0_{s,a}$. It therefore does not directly
apply when either quantity depends on these samples. To address
this dependence, we construct auxiliary models in which a given
state is absorbing, so that their value functions, for each fixed
auxiliary parameter, are independent of the samples from that
state. We then apply Bernstein's inequality conditionally on the
samples from the other states and take a union bound over finite
grids of the auxiliary parameter and $\lambda$. Controlling the
discretization error extends these bounds to the required
sample-dependent value functions and their nominal and empirical
dual optimizers. The following lemma thus provides a bound on the
one-step estimation error that can be used in the subsequent  analysis.

Recall that $\lambda^*$ and $\widehat\lambda^*$ are the nominal and
empirical dual optimizers for the given $(s,a,V)$, respectively.
\begin{lemma}
\label{lem:small-delta-uniform}
Let $\delta,\eta\in(0,1)$. Suppose that $N$ samples are drawn
from each nominal distribution $P^0_{s,a}$, with all samples
mutually independent, and that
\begin{equation}\label{eq:original-small-delta-support-samples}
 N\geq\frac{192}{p_\wedge}
 \log\frac{36|\mathcal S|^2|\mathcal A|N}{\eta p_\wedge}.
\end{equation}
Then, with probability at least $1-\eta$, the following
inequalities hold simultaneously for all $(s,a,s')$ and
$V\in\{V_{\pi^*},\widehat V_{\pi^*},
\widehat V_{\widehat\pi^*}\}$:
\begin{equation}\label{eq:original-small-delta-relative}
 \left|\widehat P^0(s'|s,a)-P^0(s'|s,a)\right|
 \leq P^0(s'|s,a)\sqrt{\frac{12}{Np_\wedge}
 \log\frac{36|\mathcal S|^2|\mathcal A|N}{\eta p_\wedge}}
 \leq\frac14P^0(s'|s,a),
\end{equation}
and
\begin{align}
 &\left|\underline{\widehat P}^{V}(\cdot\mid s,a)V
       -\underline P^{V}(\cdot\mid s,a)V\right|
 \leq
 4\sqrt{\frac1N
 \log\frac{36|\mathcal S|^2|\mathcal A|N}{\eta p_\wedge}}
 \max_{\lambda\in\{\lambda^*,\widehat\lambda^*\}}
 \lambda\sqrt{\operatorname{Var}_{P^0_{s,a}}
                         (h_{s,a}^{\lambda,V})}
 \nonumber\\
 &\qquad+
 \frac{64}{p_\wedge N(1-\gamma)}
 \log\frac{36|\mathcal S|^2|\mathcal A|N}{\eta p_\wedge}.
 \label{eq:uniform-transition-concentration}
\end{align}
If either dual optimizer $\lambda^*$ or $\widehat\lambda^*$ equals zero, its contribution to the above
maximum is defined by
\[
 \left.
 \lambda\sqrt{\operatorname{Var}_{P^0_{s,a}}
                 (h_{s,a}^{\lambda,V})}
 \right|_{\lambda=0}
 :=
 \lim_{\lambda\downarrow0}
 \lambda\sqrt{\operatorname{Var}_{P^0_{s,a}}
                 (h_{s,a}^{\lambda,V})}
 =0.
\]
In particular,
\eqref{eq:original-small-delta-relative} implies
\[
 \operatorname{supp}(\widehat P^0_{s,a})
 =\operatorname{supp}(P^0_{s,a})
 \qquad\text{for every }(s,a).
\]
\end{lemma}
\begin{proof}
Set
$
 L=\log\frac{36|\mathcal S|^2|\mathcal A|N}
              {\eta p_\wedge},
 C=3+2\log(1/p_\wedge).
$
Fix a population-optimal policy $\pi^*$, and it is independent of the
samples.

We first establish the relative probability bound.
\Cref{eq:original-small-delta-support-samples} implies
$\sqrt{12L/(Np_\wedge)}\leq1/4$.
For every $(s,a,s')$ with $P^0(s'|s,a)>0$, the
multiplicative Chernoff inequality gives
\begin{align}
 &\Pr\!\left(
 \left|\widehat P^0(s'|s,a)-P^0(s'|s,a)\right|
 >
 P^0(s'|s,a)\sqrt{\frac{12L}{Np_\wedge}}
 \right)
 \nonumber\\
 &\qquad\leq
 2\exp\!\left(-\frac{4LP^0(s'|s,a)}{p_\wedge}\right)
 \leq2e^{-4L},
 \label{eq:c11-entry-tail}
\end{align}
where the last inequality uses
$P^0(s'|s,a)\geq p_\wedge$.
If $P^0(s'|s,a)=0$, then
$\widehat P^0(s'|s,a)=0$ almost surely.
A union bound over all $(s,a,s')$ therefore shows that
\eqref{eq:original-small-delta-relative} holds on an event
$\mathcal E_{\mathrm{rel}}$ with
\begin{equation}\label{eq:c11-relative-failure}
 \Pr(\mathcal E_{\mathrm{rel}}^c)
 \leq2|\mathcal S|^2|\mathcal A|e^{-4L}
 \leq\frac{\eta}{2},
\end{equation}
where 
\begin{equation}\label{eq:c11-relative-event}
 \mathcal E_{\mathrm{rel}}
 :=
 \left\{
 \left|\widehat P^0(s'|s,a)-P^0(s'|s,a)\right|
 \leq
 P^0(s'|s,a)\sqrt{\frac{12L}{Np_\wedge}}
 \quad\text{for all }(s,a,s')
 \right\}.
\end{equation}

On this event, the nominal and empirical supports agree, and
\begin{equation}\label{eq:c11-probability-comparison}
 \frac34P^0(s'|s,a)
 \leq\widehat P^0(s'|s,a)
 \leq\frac54P^0(s'|s,a)
 \qquad\text{for every }(s,a,s').
\end{equation}
Moreover,
\begin{align}
 d_{\mathrm{TV}}(P^0_{s,a},\widehat P^0_{s,a})
 &=\frac12\sum_{s'\in\mathcal S}
 \left|\widehat P^0(s'|s,a)-P^0(s'|s,a)\right|
 \nonumber\\
 &\leq\sqrt{\frac{3L}{Np_\wedge}}
 \qquad\text{for every }(s,a).
 \label{eq:original-small-delta-base-tv}
\end{align}

\noindent\textbf{Step 1: Bernstein's inequality for fixed functions.}
For $V\in[0,(1-\gamma)^{-1}]^{|\mathcal S|}$ and $\lambda>0$,
define the centered function
$\phi_{s,a}^{\lambda,V}:
\operatorname{supp}(P^0_{s,a})\to\mathbb R$ by
\begin{equation}\label{eq:original-centered-expression}
 \phi_{s,a}^{\lambda,V}(s')
 :=
 \lambda\bigl(h_{s,a}^{\lambda,V}(s')-1\bigr)
 =
 \lambda\left(
 \frac{e^{-V(s')/\lambda}}
      {\mathbb E_{P^0_{s,a}}[e^{-V/\lambda}]}-1
 \right).
\end{equation}
The definition of $\phi_{s,a}^{\lambda,V}$ and the normalization
$\mathbb E_{P^0_{s,a}}[h_{s,a}^{\lambda,V}]=1$ give the mean
and variance identities below
\begin{align}
 \mathbb E_{P^0_{s,a}}[\phi_{s,a}^{\lambda,V}]
 &=0,
 \nonumber\\
 \left|\phi_{s,a}^{\lambda,V}(s')\right|
 &\leq\frac{1}{p_\wedge(1-\gamma)}
 \qquad
 \text{for every }s'\in\operatorname{supp}(P^0_{s,a}),
 \nonumber\\
 \operatorname{Var}_{P^0_{s,a}}(\phi_{s,a}^{\lambda,V})
 &=\lambda^2
   \operatorname{Var}_{P^0_{s,a}}(h_{s,a}^{\lambda,V}).
 \label{eq:c11-centered-moments}
\end{align}
The absolute bound follows from  \eqref{eq:81},
together with $\operatorname{span}(V)\leq(1-\gamma)^{-1}$.

If $V$ and $\lambda$ are fixed independently of the samples
from $(s,a)$, Bernstein's inequality gives
\begin{align}
 &\left|
 (\widehat P^0_{s,a}-P^0_{s,a})\phi_{s,a}^{\lambda,V}
 \right|
 \leq
 \sqrt{\frac{8L}{N}
       \operatorname{Var}_{P^0_{s,a}}
       (\phi_{s,a}^{\lambda,V})}
 +\frac{16L}{3p_\wedge N(1-\gamma)},
 \label{eq:small-delta-fixed-bernstein}
\end{align}
which holds with probability at least $1-2e^{-4L}$.

The discretization argument in Step 2 also uses the limiting
cases $\lambda=0$ and $\lambda=\infty$. We therefore define
\begin{align}
 \phi_{s,a}^{0,V}(s')
 &:=\lim_{\lambda\downarrow0}\phi_{s,a}^{\lambda,V}(s')
 =0,
 \nonumber\\
 \phi_{s,a}^{\infty,V}(s')
 &:=\lim_{\lambda\to\infty}\phi_{s,a}^{\lambda,V}(s')
 =\mathbb E_{P^0_{s,a}}[V]-V(s').
 \label{eq:c11-centered-endpoints}
\end{align}
The bound \eqref{eq:small-delta-fixed-bernstein} is immediate
for $\phi_{s,a}^{0,V}=0$. For $\phi_{s,a}^{\infty,V}$,
it follows from Bernstein's inequality because this function
has mean zero under $P^0_{s,a}$ and absolute value at most
$(1-\gamma)^{-1}$.

\noindent\textbf{Step 2: Deriving uniform concentration.}
We first establish regularity estimates for the centered
function $\phi_{s,a}^{\lambda,V}$ to control the approximation
error of the finite grids constructed below.
For a fixed state--action pair $(s,a)$, any
$V,W\in[0,(1-\gamma)^{-1}]^{|\mathcal S|}$, and finite
$\lambda,\mu\geq0$, we have
\begin{align}
 \|\phi_{s,a}^{\lambda,V}
       -\phi_{s,a}^{\lambda,W}\|_\infty
 &\leq\frac{2}{p_\wedge}\|V-W\|_\infty,
 \label{eq:c11-value-lipschitz}\\
 \|\phi_{s,a}^{\lambda,V}
       -\phi_{s,a}^{\mu,V}\|_\infty
 &\leq\frac{C}{p_\wedge}|\lambda-\mu|.
 \label{eq:c11-lambda-lipschitz}
\end{align}
For positive dual parameters, these inequalities follow by
differentiating \eqref{eq:original-centered-expression}.
The calculation uses
$h_{s,a}^{\lambda,V}(s')\leq p_\wedge^{-1}$ whenever
$P^0(s'|s,a)>0$, the normalization
$\mathbb E_{P^0_{s,a}}[h_{s,a}^{\lambda,V}]=1$, and the bounds
\begin{align}
 0
 &\leq
 \mathbb E_{P^0_{s,a}}
 \left[h_{s,a}^{\lambda,V}
       \log h_{s,a}^{\lambda,V}\right]
 \leq\log(1/p_\wedge),
 \nonumber\\
 |x\log x|
 &\leq p_\wedge^{-1}\log(1/p_\wedge)+1/e,
 \qquad 0<x\leq p_\wedge^{-1}.
 \label{eq:c11-weight-moment-bounds}
\end{align}
The continuous extension at $\lambda=0$ gives the same
Lipschitz bounds when either parameter is zero.

For large $\lambda$, we approximate
$\phi_{s,a}^{\lambda,V}$ by its limiting function
$\phi_{s,a}^{\infty,V}$ defined in
\eqref{eq:c11-centered-endpoints}.
A Taylor expansion in $1/\lambda$ around zero yields
\begin{equation}\label{eq:c11-large-lambda}
 \|\phi_{s,a}^{\lambda,V}
       -\phi_{s,a}^{\infty,V}\|_\infty
 \leq\frac{e}{2\lambda(1-\gamma)^2}
 \leq\frac{2}{\lambda(1-\gamma)^2},
 \qquad
 \lambda\geq\frac1{1-\gamma}.
\end{equation}
Indeed, for $0\leq t\leq1-\gamma$, the second derivative of
$e^{-tV(s')}/\mathbb E_{P^0_{s,a}}[e^{-tV}]$ with respect
to $t$ has absolute value at most $e/(1-\gamma)^2$.
Thus, sufficiently large values of $\lambda$ can be covered
using the single limiting function
$\phi_{s,a}^{\infty,V}$.

% ------
% Differentiating~\eqref{eq:original-centered-expression} shows that it is
% $2/p_\wedge$-Lipschitz in $V$ under the supremum norm and
% $(3+2\log(1/p_\wedge))/p_\wedge$-Lipschitz in $\lambda$; use
% $h_{s,a}^{\lambda,V}\leq p_\wedge^{-1}$,
% $\mathbb E_Ph_{s,a}^{\lambda,V}=1$,
% $\mathbb E_P[h_{s,a}^{\lambda,V}\log h_{s,a}^{\lambda,V}]
% \leq\log(1/p_\wedge)$, and $|x\log x|\leq
% p_\wedge^{-1}\log(1/p_\wedge)+1/e$ for $0<x\leq p_\wedge^{-1}$.
% Taylor's theorem at $1/\lambda=0$ also gives
% \[
%  \|\lambda(h_{s,a}^{\lambda,V}-1)-(\mathbb E_PV-V)\|_\infty
%  \leq\frac2{\lambda(1-\gamma)^2},
%  \qquad\lambda\geq\frac1{1-\gamma},
% \]
% since the second derivative of the normalized exponential with
% respect to $1/\lambda$ is bounded by $e/(1-\gamma)^2$ on this interval.

% ------------------------------------

To handle the dependence of the empirical value functions on
the samples, we construct an auxiliary model for each state $s$.
For a fixed parameter $\theta\in[0,(1-\gamma)^{-1}]$, replace
the nominal transition distribution at $(s,a)$ by the point
mass $e_s$ and the reward by $(1-\gamma)\theta$, for every
action $a$. Keep the empirical model at all other states
unchanged. Since the forward-KL uncertainty set centered at
$e_s$ contains only $e_s$, state $s$ is absorbing and its
value is $\theta$.

Let $W_s^{\pi^*}(\theta)$ denote the robust value of the fixed
policy $\pi^*$ in this auxiliary model, and let
$W_s^*(\theta)$ denote its optimal robust value.
For each fixed $\theta$, both vectors depend only on samples
collected from state--action pairs $(x,a)$ with $x\neq s$.
Conditional on these samples, the two value vectors are fixed,
while the samples from $(s,a)$ remain independent samples from
$P^0_{s,a}$. Thus, the Bernstein bound from Step 1 applies to
the centered functions associated with either auxiliary value
and any fixed $\lambda$.

These auxiliary families recover the empirical value functions
at suitable choices of $\theta$:
\begin{align}
 W_s^{\pi^*}\!\left(\widehat V_{\pi^*}(s)\right)
 &=\widehat V_{\pi^*},
 \nonumber\\
 W_s^*\!\left(\widehat V_{\widehat\pi^*}(s)\right)
 &=\widehat V_{\widehat\pi^*}.
 \label{eq:c11-aux-reconstruction}
\end{align}
Indeed, the original empirical value satisfies the unchanged
Bellman equations at every state other than $s$. At state $s$,
setting $\theta$ equal to that value gives
$(1-\gamma)\theta+\gamma\theta=\theta$.
The identities in \eqref{eq:c11-aux-reconstruction} therefore
follow from uniqueness of the corresponding Bellman fixed points.

Moreover, changing $\theta$ to $\theta'$ changes the auxiliary
reward vector by at most $(1-\gamma)|\theta-\theta'|$.
The contraction properties of the policy and optimal Bellman
operators imply
\begin{align}
 \left\|W_s^{\pi^*}(\theta)
       -W_s^{\pi^*}(\theta')\right\|_\infty
 &\leq|\theta-\theta'|,
 \nonumber\\
 \left\|W_s^*(\theta)-W_s^*(\theta')\right\|_\infty
 &\leq|\theta-\theta'|.
 \label{eq:c11-aux-lipschitz}
\end{align}
Thus, discretizing the scalar parameter $\theta$ also controls
the approximation error of the corresponding value vectors.

We now choose deterministic grids for
$\theta\in[0,(1-\gamma)^{-1}]$ and
$\lambda\in[0,N/(1-\gamma)]$, including their endpoints, with
respective maximum spacings
\begin{equation}\label{eq:c11-grid-spacings}
 \frac1{N(1-\gamma)}
 \qquad\text{and}\qquad
 \frac1{NC(1-\gamma)}.
\end{equation}
We also include the limiting function at $\lambda=\infty$.
For each $(s,a)$ and each pair of grid points, form the centered
functions associated with $W_s^{\pi^*}(\theta)$ and
$W_s^*(\theta)$. Include also the centered functions associated
with the fixed value $V_{\pi^*}$ at every $\lambda$-grid point.

The $\theta$-grid contains at most $N+1$ points, and the
$\lambda$-grid, including $\infty$, contains at most
$\lceil N^2C\rceil+2$ points. Consequently, the total number
of centered grid functions over all $(s,a)$ is at most
\begin{equation}\label{eq:c11-grid-cardinality}
 |\mathcal S||\mathcal A|(2N+3)(\lceil N^2C\rceil+2)
 \leq12|\mathcal S||\mathcal A|N^3C.
\end{equation}

We next bound the approximation error of these grids.
Fix $(s,a)$ and
$V\in\{V_{\pi^*},\widehat V_{\pi^*},
\widehat V_{\widehat\pi^*}\}$.
By \eqref{eq:c11-aux-reconstruction} and
\eqref{eq:c11-aux-lipschitz}, there is a value vector $V_g$
included in the grid construction such that
\begin{equation}\label{eq:c11-value-grid-error}
 \|V-V_g\|_\infty\leq\frac1{N(1-\gamma)}.
\end{equation}
For $V=V_{\pi^*}$, we simply take $V_g=V_{\pi^*}$.

If $0\leq\lambda\leq N/(1-\gamma)$, choose a grid point
$\lambda_g$ with
$|\lambda-\lambda_g|\leq1/[NC(1-\gamma)]$.
The preceding Lipschitz bounds give
\begin{align}
 \|\phi_{s,a}^{\lambda,V}
       -\phi_{s,a}^{\lambda_g,V_g}\|_\infty
 &\leq
 \frac{2}{p_\wedge}\|V-V_g\|_\infty
 +\frac{C}{p_\wedge}|\lambda-\lambda_g|
 \nonumber\\
 &\leq\frac{3}{p_\wedge N(1-\gamma)}.
 \label{eq:c11-finite-lambda-grid-error}
\end{align}
If $\lambda>N/(1-\gamma)$, use the limiting grid function
$\phi_{s,a}^{\infty,V_g}$ instead.
The Taylor bound above and the $2$-Lipschitz property of
$V\mapsto\mathbb E_{P^0_{s,a}}[V]-V$ yield
\begin{align}
 \|\phi_{s,a}^{\lambda,V}
       -\phi_{s,a}^{\infty,V_g}\|_\infty
 &\leq
 \frac{2}{\lambda(1-\gamma)^2}
 +2\|V-V_g\|_\infty
 \nonumber\\
 &\leq\frac{4}{N(1-\gamma)}
 \leq\frac{4}{p_\wedge N(1-\gamma)}.
 \label{eq:c11-large-lambda-grid-error}
\end{align}
Therefore, for every required $V$ and every $\lambda\geq0$,
there is a centered grid function $\phi_{\mathrm{grid}}$
satisfying
\begin{equation}\label{eq:original-small-delta-grid-error}
 \|\phi_{s,a}^{\lambda,V}-\phi_{\mathrm{grid}}\|_\infty
 \leq\frac{4}{p_\wedge N(1-\gamma)}.
\end{equation}

% --------------------------------------------

Applying~\eqref{eq:small-delta-fixed-bernstein} conditional on the samples collected from the other states and taking the union bound gives simultaneous Bernstein
bounds on these grids with failure probability at most $\eta/2$.
Together with~\eqref{eq:original-small-delta-relative}, with probability
at least $1-\eta$, for every required $V,\lambda$,
\begin{align}
 &\lambda\left|
 (\widehat P^0_{s,a}-P^0_{s,a})h_{s,a}^{\lambda,V}
 \right|
 \nonumber\\
 &\overset{\mathrm{(i)}}{\leq}
 \left|
 (\widehat P^0_{s,a}-P^0_{s,a})\phi_{\mathrm{grid}}
 \right|
 +\frac{8}{p_\wedge N(1-\gamma)}
 \nonumber\\
 &\overset{\mathrm{(ii)}}{\leq}
 \sqrt{\frac{8L}{N}
       \operatorname{Var}_{P^0_{s,a}}(\phi_{\mathrm{grid}})}
 +\frac{16L}{3p_\wedge N(1-\gamma)}
 +\frac{8}{p_\wedge N(1-\gamma)}
 \nonumber\\
 &\overset{\mathrm{(iii)}}{\leq}
 \lambda\sqrt{\frac{8L}{N}
       \operatorname{Var}_{P^0_{s,a}}(h_{s,a}^{\lambda,V})}
 +\frac{16L}{3p_\wedge N(1-\gamma)}
 \nonumber\\
 &\qquad
 +\frac{4}{p_\wedge N(1-\gamma)}
  \left(2+\sqrt{\frac{8L}{N}}\right)
 \nonumber\\
 &\overset{\mathrm{(iv)}}{\leq}
 \lambda\sqrt{\frac{8L}{N}
       \operatorname{Var}_{P^0_{s,a}}(h_{s,a}^{\lambda,V})}
 +\frac{40L}{p_\wedge N(1-\gamma)}.
 \label{eq:original-small-delta-centered-remainder}
\end{align}
Here, (i) follows from
\eqref{eq:original-small-delta-grid-error},
the identity
$(\widehat P^0_{s,a}-P^0_{s,a})\mathbf 1=0$,
and $\|\widehat P^0_{s,a}-P^0_{s,a}\|_1\leq2$;
(ii) uses the simultaneous Bernstein bound for the grid
functions; (iii) combines
\eqref{eq:original-small-delta-grid-error} and
\eqref{eq:c11-centered-moments} with the fact that standard
deviation is $1$-Lipschitz under the supremum norm; and
(iv) uses \eqref{eq:original-small-delta-support-samples}
and $L\geq1$.

\noindent\textbf{Step 3: Comparing the empirical and nominal transitions.}
On the event established in Step 2, which has probability
at least $1-\eta$,
\eqref{eq:original-small-delta-relative} gives, simultaneously
for every $(s,a)$, value vector $V$, and $\lambda>0$,
\begin{equation}\label{eq:c11-normalizer-ratio}
 \frac34
 \leq
 \frac{\mathbb E_{\widehat P^0_{s,a}}[e^{-V/\lambda}]}
      {\mathbb E_{P^0_{s,a}}[e^{-V/\lambda}]}
 \leq\frac54.
\end{equation}
Using $|\log x|\leq\frac43|x-1|$ on $[3/4,5/4]$ and
evaluating each dual objective at the other objective's
maximizer yields
\begin{align}
 &\left|
 \underline{\widehat P}^{V}(\cdot\mid s,a)V
 -\underline P^{V}(\cdot\mid s,a)V
 \right|
 \nonumber\\
 &\qquad\leq
 \frac43
 \max_{\lambda\in\{\lambda^*,\widehat\lambda^*\}}
 \lambda
 \left|
 (\widehat P^0_{s,a}-P^0_{s,a})h_{s,a}^{\lambda,V}
 \right|.
 \label{eq:c11-dual-maxima-comparison}
\end{align}
For each value vector $V$ listed in the lemma, applying
\eqref{eq:original-small-delta-centered-remainder} and using
$4\sqrt8/3<4$ and $160/3<64$ gives
\eqref{eq:uniform-transition-concentration}.
If either optimizer equals zero, its corresponding term
in the dual comparison is defined by continuous extension
and equals zero, since the two dual objectives coincide
at $\lambda=0$ when the nominal and empirical supports agree.
\end{proof}
% --------------
% ---------------
Let $\widehat\pi:=\pi_T$ denote the deterministic policy returned by
robust value iteration, with ties resolved by a fixed rule. Choose
$\pi^*$ to be a deterministic population-optimal policy.

\begin{lemma}[Comparison of value variances]
\label{lem:small-delta-variance-comparison}
Suppose $0<\delta\le1$ and
\begin{equation}\label{eq:small-delta-variance-samples}
 N\geq\frac{192}{p_\wedge}
 \log\frac{36|\mathcal S|^2|\mathcal A|N}{\eta p_\wedge}.
\end{equation}
On the event of Lemma~\ref{lem:small-delta-uniform}, the following
inequalities hold at every state:
\begin{align}
 &\max\left\{
 \sqrt{\operatorname{Var}_{\underline P^{\widehat V_{\pi^*},\pi^*}}
                              (\widehat V_{\pi^*})},
 \sqrt{\operatorname{Var}_{\underline{\widehat P}
                                  ^{\widehat V_{\pi^*},\pi^*}}
                              (\widehat V_{\pi^*})}\right\}
 \nonumber\\
 &\leq\frac2{\sqrt3}
 \sqrt{\operatorname{Var}_{\underline P^{V_{\pi^*},\pi^*}}
                                  (V_{\pi^*})}
 +\left(\frac2{\sqrt3}\|\widehat V_{\pi^*}-V_{\pi^*}\|_\infty
                    +\frac{2\delta^{1/4}}{1-\gamma}\right)
                                  \mathbf1,
 \label{eq:small-delta-variance-comparison-one}\\
 &\max\left\{
 \sqrt{\operatorname{Var}_{\underline P^{V_{\pi^*},\pi^*}}
                              (V_{\pi^*})},
 \sqrt{\operatorname{Var}_{\underline{\widehat P}^{V_{\pi^*},\pi^*}}
                              (V_{\pi^*})}\right\}
 \nonumber\\
 &\leq\frac2{\sqrt3}
 \sqrt{\operatorname{Var}_{\underline{\widehat P}
                         ^{\widehat V_{\pi^*},\pi^*}}
                                  (\widehat V_{\pi^*})}
 +\left(\frac2{\sqrt3}\|V_{\pi^*}-\widehat V_{\pi^*}\|_\infty
                    +\frac{2\delta^{1/4}}{1-\gamma}\right)
                                  \mathbf1,
 \label{eq:small-delta-variance-comparison-two}\\
 &\max\left\{
 \sqrt{\operatorname{Var}_{\underline P
                         ^{\widehat V_{\widehat\pi^*},\widehat\pi}}
                              (\widehat V_{\widehat\pi^*})},
 \sqrt{\operatorname{Var}_{\underline{\widehat P}
                         ^{\widehat V_{\widehat\pi^*},\widehat\pi}}
                              (\widehat V_{\widehat\pi^*})}\right\}
 \nonumber\\
 &\leq\frac2{\sqrt3}
 \sqrt{\operatorname{Var}_{\underline P^{V_{\widehat\pi},\widehat\pi}}
                                  (V_{\widehat\pi})}
 +\left(\frac2{\sqrt3}\|\widehat V_{\widehat\pi^*}-V_{\widehat\pi}\|_\infty
                    +\frac{2\delta^{1/4}}{1-\gamma}\right)
                                  \mathbf1.
 \label{eq:small-delta-variance-comparison-three}
\end{align}
These conclusions also hold when a nominal or empirical dual
optimizer equals zero.
\end{lemma}
\begin{proof}
Fix a state-action pair $(s,a)$. By
\eqref{eq:original-small-delta-relative}, for every value function $V$,
\[
 \operatorname{Var}_{\widehat P^0_{s,a}}(V)
 \leq\frac54\operatorname{Var}_{P^0_{s,a}}(V),\qquad
 \operatorname{Var}_{P^0_{s,a}}(V)
 \leq\frac43\operatorname{Var}_{\widehat P^0_{s,a}}(V).
\]
To see this, center under the distribution on the right and use
$\frac34P^0_{s,a}\leq\widehat P^0_{s,a}\leq\frac54P^0_{s,a}$.
Pinsker's inequality also shows that replacing either base row by
any distribution in its KL uncertainty set changes the standard
deviation of a value function by at most
$\delta^{1/4}/[2^{1/4}(1-\gamma)]$. Indeed, for a center in
$[0,(1-\gamma)^{-1}]$, the squared deviation lies in
$[0,(1-\gamma)^{-2}]$, so its expectations differ by at most
$\sqrt{\delta/2}/(1-\gamma)^2$; minimizing over the center and
taking square roots gives the assertion. Consequently,
\begin{align*}
 &\sqrt{\operatorname{Var}_{\underline{\widehat P}^{\widehat V_{\pi^*}}_{s,a}}
                         (\widehat V_{\pi^*})}\\
 &\leq\sqrt{\operatorname{Var}_{\widehat P^0_{s,a}}(\widehat V_{\pi^*})}
       +\frac{\delta^{1/4}}{2^{1/4}(1-\gamma)}\\
 &\leq\frac2{\sqrt3}\sqrt{\operatorname{Var}_{P^0_{s,a}}(\widehat V_{\pi^*})}
       +\frac{\delta^{1/4}}{2^{1/4}(1-\gamma)}\\
 &\leq\frac2{\sqrt3}
       \sqrt{\operatorname{Var}_{\underline P^{V_{\pi^*}}_{s,a}}
                              (\widehat V_{\pi^*})}
       +\frac{(1+2/\sqrt3)\delta^{1/4}}{2^{1/4}(1-\gamma)}\\
 &\leq\frac2{\sqrt3}
       \sqrt{\operatorname{Var}_{\underline P^{V_{\pi^*}}_{s,a}}(V_{\pi^*})}
       +\frac2{\sqrt3}\|\widehat V_{\pi^*}-V_{\pi^*}\|_\infty
       +\frac{2\delta^{1/4}}{1-\gamma}.
\end{align*}
The last step uses the triangle inequality for standard deviations
and $(1+2/\sqrt3)2^{-1/4}<2$.
The same argument applies to the nominal worst-case row at
$\widehat V_{\pi^*}$, with both distributions compared to $P^0_{s,a}$.
Choosing $a=\pi^*(s)$ proves
\eqref{eq:small-delta-variance-comparison-one}.
Interchanging the nominal and empirical rows and the two value
functions proves~\eqref{eq:small-delta-variance-comparison-two}.
Replacing the two value functions by
$\widehat V_{\widehat\pi^*},V_{\widehat\pi}$ and choosing
$a=\widehat\pi(s)$ proves
\eqref{eq:small-delta-variance-comparison-three}.
The argument also covers zero dual optimizers because their
worst-case distributions belong to the same KL uncertainty sets.\qedhere
\end{proof}
\subsection{Variance transfer for small uncertainty}\label{sec:c5}
In the later part, we will provide a tighter bound when uncertainty level $\delta$ is small. 
Here, we first control the density ratio
$\underline P^V(s'|s,a)/P^0(s'|s,a)$. The next two lemmas give
tighter bounds from the KL constraint; combining
them with the minimum nominal probability will produce the three
branches used in the variance estimate. We begin with the
exponential bound.

\begin{lemma}[Probability ratio bound under a KL constraint]\label{lem:atom-exponential}
Let $p_1,p_2$ be probability distributions on a finite set $\mathcal X$ with
$\operatorname{KL}(p_1\|p_2)\leq\delta$. 
 Then, for every $x$ with
$p_2(x)>0$,
\begin{align}
p_1(x)\leq p_2(x)\exp\!\left(1+\frac{\delta}{p_2(x)}\right).
\end{align}
% If
% $p_1(x)>p_2(x)>0$, then
% \begin{align}
%  p_1(x)\leq p_2(x)\exp\!\left(1+\frac{\delta}{p_2(x)}\right).
% \end{align}
\end{lemma}

\begin{proof}
Fix $x$ with $p_2(x)>0$. Since $\delta\geq0$, the claim is
immediate when $p_1(x)\leq p_2(x)$. We therefore assume
$p_1(x)>p_2(x)>0$.

We first note that, for $u,v>0$, the inequality
$\log t\leq t-1$ gives
\begin{align}
u\log(u/v)
=-u\log(v/u)
\geq -u\left(\frac{v}{u}-1\right)
=u-v.
\end{align}
This inequality also holds for $u=0$ by continuity, with the
convention $0\log(0/v)=0$.

Let $\mathcal X_2=\operatorname{supp}(p_2)$, where $\operatorname{supp}$ denotes the support. Finite KL divergence
implies that $p_1(y)=0$ for every $y\notin\mathcal X_2$.
Applying the preceding scalar inequality to each
$y\in\mathcal X_2\setminus\{x\}$, we obtain
\begin{align}
\delta
&\geq \operatorname{KL}(p_1\|p_2)\\
&=p_1(x)\log\frac{p_1(x)}{p_2(x)}
  +\sum_{y\in\mathcal X_2\setminus\{x\}}
     p_1(y)\log\frac{p_1(y)}{p_2(y)}\\
&\geq p_1(x)\log\frac{p_1(x)}{p_2(x)}
  +\sum_{y\in\mathcal X_2\setminus\{x\}}
     \bigl(p_1(y)-p_2(y)\bigr)\\
&=p_1(x)\log\frac{p_1(x)}{p_2(x)}
  +p_2(x)-p_1(x).
\end{align}
Dividing by $p_1(x)>0$ and rearranging yields
\begin{align}
\log\frac{p_1(x)}{p_2(x)}
&\leq \frac{\delta}{p_1(x)}
       +1-\frac{p_2(x)}{p_1(x)}\\
&\leq \frac{\delta}{p_2(x)}+1,
\end{align}
where the last inequality uses $\delta\geq0$ and
$p_1(x)>p_2(x)>0$. Exponentiating gives
\begin{align}
p_1(x)\leq
p_2(x)\exp\!\left(1+\frac{\delta}{p_2(x)}\right),
\end{align}
which proves the claim.
\end{proof}

% The exponential estimate is useful when $\delta$ is small relative
% to the nominal atom probability. A second estimate controls the
% additive change of each atom through total variation. After
% division by the nominal probability, it yields the
% $\sqrt{\delta}/p_\wedge$ branch and complements the exponential
% branch.

The preceding exponential bound is useful when $\delta$ is small
relative to the nominal probability $p_2(x)$. We next derive an
alternative bound using Pinsker's inequality, which controls
$|p_1(x)-p_2(x)|$. Dividing by $p_2(x)$ yields a bound of
$1+\sqrt{\delta/2}/p_2(x)$ on the probability ratio.
For nominal transition probabilities satisfying
$p_2(x)\geq p_\wedge$, this gives the $\sqrt{\delta}/p_\wedge$
term in the subsequent density ratio estimate.

\begin{lemma}[Pinsker bound on probability differences]\label{lm:prob2}
Let $p_1$ and $p_2$ be probability distributions on a finite set
$\mathcal X$ satisfying
$\operatorname{KL}(p_1\|p_2)\leq\delta$.
Then, for every $x\in\mathcal X$,
\begin{align}
 |p_1(x)-p_2(x)|\leq\sqrt{\delta/2}.
\end{align}
\end{lemma}
\begin{proof}
For finite distributions, total variation has the equivalent form:
\begin{align}
 d_{\mathrm{TV}}(p_1,p_2)
 =\frac12\sum_y|p_1(y)-p_2(y)|
 =\sup_B|p_1(B)-p_2(B)|.
\end{align}
Applying Pinsker's inequality yields
\begin{align}
 |p_1(x)-p_2(x)|
 \leq d_{\mathrm{TV}}(p_1,p_2)
 \leq\sqrt{\frac{\operatorname{KL}(p_1\|p_2)}{2}}
 \leq\sqrt{\delta/2}.
\end{align}
This completes the proof.
% In particular, when $p_1(x)>p_2(x)$ this is the    one-sided
% bound $p_1(x)\leq p_2(x)+\sqrt{\delta/2}$.
\end{proof}

% We now apply the two atomwise estimates to the nominal and
% worst-case transition rows. Since a transition probability is also
% at most one, the probability floor supplies a third bound. Taking
% the minimum retains whichever of these three controls is strongest
% in the parameter regime under consideration.

We now apply Lemmas~\ref{lem:atom-exponential} and~\ref{lm:prob2}
to the worst-case transition distribution $\underline P^V_{s,a}$
and the nominal distribution $P^0_{s,a}$. On the support of
$P^0_{s,a}$, the inequalities
$\underline P^V(s'\mid s,a)\leq1$ and
$P^0(s'\mid s,a)\geq p_\wedge$ give an additional upper bound
of $1/p_\wedge$ on the density ratio. Combining these three
bounds yields the following result.

\begin{lemma}[Density ratio bounds for worst-case transitions]
\label{lm:prob}
Fix a state-action pair $(s,a)\in\mathcal S\times\mathcal A$
and a value vector $V$. Suppose that
$\operatorname{KL}(\underline P^V_{s,a}\|P^0_{s,a})\leq\delta$.
Then the maximum density ratio on the support of $P^0_{s,a}$
satisfies
\begin{align}
 \max_{\substack{s'\in\mathcal S:\\ P^0(s'\mid s,a)>0}}
 \frac{\underline P^V(s'\mid s,a)}{P^0(s'\mid s,a)}
 \leq
 \min\left\{
   \exp\!\left(1+\frac{\delta}{p_\wedge}\right),
   1+\frac{\sqrt{\delta/2}}{p_\wedge},
   \frac{1}{p_\wedge}
 \right\}.
\end{align}
\end{lemma}

% \begin{lemma}[Density ratio bounds for worst-case transitions]\label{lm:prob}
% If $\operatorname{KL}(\underline P^V_{s,a}\|P^0_{s,a})\leq\delta$,
% then
% \begin{align}
%  \max_{s':P^0(s'|s,a)>0}
%  \frac{\underline P^V(s'|s,a)}{P^0(s'|s,a)}
%  \leq\min\left\{e^{1+\delta/p_\wedge},
%                   1+\frac{\sqrt{\delta/2}}{p_\wedge},
%                   \frac1{p_\wedge}\right\}.
% \end{align}
% \end{lemma}

\begin{proof}
Fix any $s'\in\mathcal S$ with $P^0(s'\mid s,a)>0$.
By the definition of $p_\wedge$,
$P^0(s'\mid s,a)\geq p_\wedge>0$.

Applying Lemma~\ref{lem:atom-exponential} with
$p_1=\underline P^V_{s,a}$ and $p_2=P^0_{s,a}$ gives
\begin{align}
 \frac{\underline P^V(s'\mid s,a)}{P^0(s'\mid s,a)}
 &\leq
 \exp\!\left(1+\frac{\delta}{P^0(s'\mid s,a)}\right)\\
 &\leq
 \exp\!\left(1+\frac{\delta}{p_\wedge}\right).
\end{align}
Next, Lemma~\ref{lm:prob2}, applied to the same distributions,
yields
\begin{align}
 \underline P^V(s'\mid s,a)
 \leq P^0(s'\mid s,a)+\sqrt{\delta/2}.
\end{align}
Dividing by $P^0(s'\mid s,a)$, we obtain
\begin{align}
 \frac{\underline P^V(s'\mid s,a)}{P^0(s'\mid s,a)}
 &\leq 1+\frac{\sqrt{\delta/2}}{P^0(s'\mid s,a)}\\
 &\leq 1+\frac{\sqrt{\delta/2}}{p_\wedge}.
\end{align}
Finally, since $\underline P^V(s'\mid s,a)\leq1$,
\begin{align}
 \frac{\underline P^V(s'\mid s,a)}{P^0(s'\mid s,a)}
 \leq \frac{1}{P^0(s'\mid s,a)}
 \leq \frac{1}{p_\wedge}.
\end{align}

These three bounds hold for every $s'$ in the support of
$P^0_{s,a}$, and their right-hand sides are independent of $s'$.
Taking their minimum and then maximizing over
$s'$ in this support proves the claim.
\end{proof}
% \begin{proof}
% The last bound follows because the numerator is at most one.
% Pinsker's inequality bounds the difference between the two
% probabilities of any singleton by $\sqrt{\delta/2}$, giving the
% second bound. For the first, the inequality
% $x\log(x/y)\geq x-y$ applied to every other atom gives
% \begin{align}
%  \delta&\geq\underline P^V(s'|s,a)
%        \log\frac{\underline P^V(s'|s,a)}{P^0(s'|s,a)}
%        +P^0(s'|s,a)-\underline P^V(s'|s,a).
% \end{align}
% If the density ratio exceeds one, division by the numerator yields a
% logarithm at most $1+\delta/P^0(s'|s,a)\leq1+\delta/p_\wedge$.
% If the ratio is at most one, the bound is immediate.

% More explicitly, Lemma~\ref{lem:atom-exponential} applied with
% $p_1=\underline P^V_{s,a}$ and $p_2=P^0_{s,a}$ gives
% \begin{align}
%  \frac{\underline P^V(s'|s,a)}{P^0(s'|s,a)}
%  \leq\exp\!\left(1+\frac{\delta}{P^0(s'|s,a)}\right)
%  \leq e^{1+\delta/p_\wedge}
% \end{align}
% whenever the ratio exceeds one; the same upper bound also covers
% ratios at most one. Lemma~\ref{lm:prob2} gives, on every positive
% nominal atom,
% \begin{align}
%  \frac{\underline P^V(s'|s,a)}{P^0(s'|s,a)}
%  \leq1+\frac{\sqrt{\delta/2}}{P^0(s'|s,a)}
%  \leq1+\frac{\sqrt{\delta/2}}{p_\wedge}.
% \end{align}
% Finally, $\underline P^V(s'|s,a)\leq1$ gives the third bound.
% Since all three hold simultaneously, their minimum is a valid upper
% bound on the maximum density ratio.
% \end{proof}

We next use Lemma~\ref{lm:prob} to relate the variance of
$h_{s,a}^{\lambda^*,V}$ under the nominal transition distribution
to the variance of $V$ under the worst-case transition distribution.
The proof compares the exponential weight at $V(s')$ with its
value at $\overline V=\mathbb E_{\underline P^V_{s,a}}[V]$.
The mean-value theorem then cancels the explicit factor
$\lambda^*$ against the reciprocal factor from the derivative.
A change of measure, together with the density ratio bound,
controls the resulting weighted second moment in terms of
$\operatorname{Var}_{\underline P^V_{s,a}}(V)$, while total
variation bounds the remaining discrepancy between the nominal
and worst-case second moments. The following lemma states this
variance comparison and its simplified form for $0<\delta\le1$.

% With the density ratio controlled, we can establish the central
% variance-transfer estimate. Centering the tilt at the worst-case
% mean allows its derivative to cancel the dual multiplier. A change
% of measure then produces the local value variance under the
% worst-case row, while total variation controls the remaining
% nominal second moment. The following lemma states the resulting
% general estimate and its sharper specialization to $0<\delta\le1$.

% \begin{lemma}[Variance transfer]\label{lem:polished-variance}
% Suppose $\lambda^*>0$ is a population dual maximizer.
% % For every
% % $1\ge\delta>0$,
% % \begin{align}
% %  &\lambda^*\sqrt{\operatorname{Var}_{P^0_{s,a}}
% %                          (h_{s,a}^{\lambda^*,V})}\nonumber\\
% %  &\leq\left[\sqrt{\min\left\{e^{1+\delta/p_\wedge},
% %                  1+\frac{\sqrt{\delta/2}}{p_\wedge},
% %                  \frac1{p_\wedge}\right\}}+4\right]
% %        \sqrt{\operatorname{Var}_{\underline P^V_{s,a}}(V)}+4\operatorname{span}(V)
% %                      \min\{(\delta/2)^{1/4},1\}.
% %  \label{eq:polished-all-delta}
% % \end{align}
% Consequently, for $0<\delta\le1$,
% \begin{align}
%  &\lambda^*\sqrt{\operatorname{Var}_{P^0_{s,a}}
%                          (h_{s,a}^{\lambda^*,V})}\nonumber\\
%  &\leq8\sqrt{\left(1+\min\left\{e^{\delta/p_\wedge},
%                          \frac{\sqrt\delta}{p_\wedge},
%                          \frac1{p_\wedge}\right\}\right)
%                     \operatorname{Var}_{\underline P^V_{s,a}}(V)}
%        +4\delta^{1/4}\operatorname{span}(V).
%  \label{eq:29}
% \end{align}
% \end{lemma}
\begin{lemma}[Variance transfer]\label{lem:polished-variance}
Let $0<\delta\le1$, and fix a state-action pair
$(s,a)\in\mathcal S\times\mathcal A$ and a value vector $V$.
Suppose that $\lambda^*>0$ is a maximizer of the population
dual objective $f(P^0_{s,a},V,\lambda)$ over $\lambda\geq0$.
Then
\begin{align}
 &\lambda^*\sqrt{
   \operatorname{Var}_{P^0_{s,a}}
   \bigl(h_{s,a}^{\lambda^*,V}\bigr)}
 \nonumber\\
 &\leq
 8\sqrt{
   \left(
     1+\min\left\{
       \exp\!\left(\frac{\delta}{p_\wedge}\right),
       \frac{\sqrt{\delta}}{p_\wedge},
       \frac{1}{p_\wedge}
     \right\}
   \right)
   \operatorname{Var}_{\underline P^V_{s,a}}(V)}
 +4\delta^{1/4}\operatorname{span}(V).
 \label{eq:29}
\end{align}
% where $\operatorname{span}(V)=\max_s V(s)-\min_s V(s)$.
\end{lemma}
\begin{proof}
As in the proof of Lemma~\ref{lm:bern}, suppress the dependence
on the fixed quantities $(s,a)$, $V$, and $\lambda^*$, and define
\begin{align}
 h(v)=\frac{e^{-v/\lambda^*}}
           {\mathbb E_{P^0_{s,a}}[e^{-V/\lambda^*}]},
 \qquad v\in\mathbb R.
\end{align}
Then
$h(V(s'))=h_{s,a}^{\lambda^*,V}(s')$ and
$h'(v)=-h(v)/\lambda^*$.
Recall $\overline V=\mathbb E_{\underline P^V_{s,a}}[V]$.
By Lemma~\ref{lm:hp}, $h(\overline V)=e^\delta$.

For each $s'$ in the support of $P^0_{s,a}$, the mean-value
theorem gives a point $\xi(s')$ between $V(s')$ and
$\overline V$ such that
\begin{align}\label{eq:106}
 \lambda^*\bigl(h(V(s'))-h(\overline V)\bigr)
 =-h(\xi(s'))\bigl(V(s')-\overline V\bigr).
\end{align}

Since $h$ is positive and decreasing,
\begin{align}\label{eq:107}
 h(\xi(s'))^2
 \leq\max\{h(V(s'))^2,e^{2\delta}\}
 \leq h(V(s'))^2+e^{2\delta}.
\end{align}
Since variance is bounded above by the mean squared deviation
from any constant, using
$h(V(s'))=h_{s,a}^{\lambda^*,V}(s')$ gives
\begin{align}
 (\lambda^*)^2\operatorname{Var}_{P^0_{s,a}}
 \bigl(h_{s,a}^{\lambda^*,V}\bigr)
 &\leq
 (\lambda^*)^2\mathbb E_{s'\sim P^0_{s,a}}
 \bigl[(h(V(s'))-h(\overline V))^2\bigr]\\
 &\leq
 \mathbb E_{s'\sim P^0_{s,a}}
 \bigl[h(V(s'))^2(V(s')-\overline V)^2\bigr]\nonumber\\
 &\quad+
 e^{2\delta}\mathbb E_{s'\sim P^0_{s,a}}
 \bigl[(V(s')-\overline V)^2\bigr].\label{eq:109}
\end{align}
\Cref{eq:109} follows from squaring \eqref{eq:106}, applying \eqref{eq:107}, and taking expectations with respect to $s'\sim P^0_{s,a}$.

To bound the first term in \eqref{eq:109}, let
\begin{align}
 H=\max_{s':P^0(s'\mid s,a)>0}h(V(s')).
\end{align}
Using the identity
$\underline P^V(s'\mid s,a)
 =P^0(s'\mid s,a)h(V(s'))$, we have
 \begin{align}
 \mathbb E_{s'\sim P^0_{s,a}}
 \bigl[h(V(s'))^2(V(s')-\overline V)^2\bigr]
 &=\mathbb E_{s'\sim\underline P^V_{s,a}}
 \bigl[h(V(s'))(V(s')-\overline V)^2\bigr]\\
 &\leq H\operatorname{Var}_{\underline P^V_{s,a}}(V).\label{eq:weighted-second-moment}
\end{align}

% \begin{align}
%  \mathbb E_{P^0_{s,a}}
%  \bigl[h(V)^2(V-\overline V)^2\bigr]
%  &=\mathbb E_{\underline P^V_{s,a}}
%  \bigl[h(V)(V-\overline V)^2\bigr]\\
%  &\leq H\operatorname{Var}_{\underline P^V_{s,a}}(V).
% \end{align}

% For the second term in \eqref{eq:109}, note that
% $\min_s V(s)\leq\overline V\leq\max_s V(s)$, so
% % $0\leq(V-\overline V)^2\leq\operatorname{span}(V)^2$.

For the second term in \eqref{eq:109}, note that
$\min_s V(s)\leq\overline V\leq\max_s V(s)$.
Hence, for every $s'\in\mathcal S$,
$ 0\leq(V(s')-\overline V)^2\leq\operatorname{span}(V)^2.$

The bounded-function inequality for total variation and
Pinsker's inequality therefore give
\begin{align}
 \mathbb E_{s'\sim P^0_{s,a}}
 \bigl[(V(s')-\overline V)^2\bigr]
 &\leq
 \operatorname{Var}_{\underline P^V_{s,a}}(V)
 +\operatorname{span}(V)^2
 d_{\mathrm{TV}}(P^0_{s,a},\underline P^V_{s,a})
 \nonumber\\
 &\leq
 \operatorname{Var}_{\underline P^V_{s,a}}(V)
 +\operatorname{span}(V)^2\sqrt{\delta/2},
 \label{eq:nominal-second-moment}
\end{align}
where we used
$\operatorname{KL}(\underline P^V_{s,a}\|P^0_{s,a})=\delta$.

% \begin{align}
%  \mathbb E_{P^0_{s,a}}\bigl[(V-\overline V)^2\bigr]
%  &\leq
%  \operatorname{Var}_{\underline P^V_{s,a}}(V)+
%  \operatorname{span}(V)^2
%  d_{\mathrm{TV}}(P^0_{s,a},\underline P^V_{s,a})\nonumber\\
%  &\leq
%  \operatorname{Var}_{\underline P^V_{s,a}}(V)
%  +\operatorname{span}(V)^2\sqrt{\delta/2},\label{eq:nominal-second-moment}
% \end{align}
% where we used
% $\operatorname{KL}(\underline P^V_{s,a}\|P^0_{s,a})=\delta$.

Substituting \eqref{eq:weighted-second-moment} into
\eqref{eq:109} and taking square roots yields the first
inequality below. Applying \eqref{eq:nominal-second-moment}
then gives the second. In both steps, we use
$\sqrt{x+y}\leq\sqrt{x}+\sqrt{y}$ for $x,y\geq0$:
\begin{align}
 \lambda^*\sqrt{\operatorname{Var}_{P^0_{s,a}}
 \bigl(h_{s,a}^{\lambda^*,V}\bigr)}
 &\leq
 \sqrt H\,
 \sqrt{\operatorname{Var}_{\underline P^V_{s,a}}(V)}
 +e^\delta
 \sqrt{\mathbb E_{s'\sim P^0_{s,a}}
 \bigl[(V(s')-\overline V)^2\bigr]}\\
 &\leq
 (\sqrt H+e^\delta)
 \sqrt{\operatorname{Var}_{\underline P^V_{s,a}}(V)}
 +e^\delta(\delta/2)^{1/4}\operatorname{span}(V).
\end{align}

% Combining these estimates, taking square roots, and using
% $\sqrt{x+y}\leq\sqrt{x}+\sqrt{y}$ yields
% \begin{align}
%  \lambda^*\sqrt{\operatorname{Var}_{P^0_{s,a}}(h(V))}
%  &\leq
%  \sqrt H\,
%  \sqrt{\operatorname{Var}_{\underline P^V_{s,a}}(V)}
%  +e^\delta
%  \sqrt{\mathbb E_{P^0_{s,a}}[(V-\overline V)^2]}\\
%  &\leq
%  (\sqrt H+e^\delta)
%  \sqrt{\operatorname{Var}_{\underline P^V_{s,a}}(V)}
%  +e^\delta(\delta/2)^{1/4}\operatorname{span}(V).
% \end{align}

Finally, set
\begin{align}
 K_\delta
 =1+\min\left\{
   \exp\!\left(\frac{\delta}{p_\wedge}\right),
   \frac{\sqrt{\delta}}{p_\wedge},
   \frac{1}{p_\wedge}
 \right\}.
\end{align}
Lemma~\ref{lm:prob} implies
\begin{align}
 H\leq
 \min\left\{
   \exp\!\left(1+\frac{\delta}{p_\wedge}\right),
   1+\frac{\sqrt{\delta/2}}{p_\wedge},
   \frac{1}{p_\wedge}
 \right\}
 \leq eK_\delta.
\end{align}
The last inequality follows by comparing the three terms
with $e$ times one plus their respective counterparts in
the definition of $K_\delta$.
Since $0<\delta\le1$ and $K_\delta\geq1$,
\begin{align}
 \sqrt H+e^\delta
 \leq\sqrt{eK_\delta}+e
 \leq(\sqrt e+e)\sqrt{K_\delta}
 \leq8\sqrt{K_\delta}.
\end{align}
Also,
\begin{align}
 e^\delta(\delta/2)^{1/4}
 \leq e\,2^{-1/4}\delta^{1/4}
 \leq4\delta^{1/4}.
\end{align}
Substituting these bounds and recalling
$h(V(s'))=h_{s,a}^{\lambda^*,V}(s')$
proves~\eqref{eq:29}.
\end{proof}

\paragraph{Empirical version of the variance bound in
Lemma~\ref{lem:polished-variance}.}
The preceding proof also yields a variance bound for the
empirical transition kernel.

Let $0<\delta\le1$, and fix a state-action pair $(s,a)$ and a
value vector $V$. Suppose that $\widehat\lambda^*>0$ maximizes
the empirical dual objective
$f(\widehat P^0_{s,a},V,\lambda)$ over $\lambda\geq0$.
Recall that the corresponding weight function and worst-case
transition distribution are given by
\begin{align}
 \widehat h_{s,a}^{\widehat\lambda^*,V}(s')
 &=
 \frac{e^{-V(s')/\widehat\lambda^*}}
      {\mathbb E_{s''\sim\widehat P^0_{s,a}}
       [e^{-V(s'')/\widehat\lambda^*}]},\\
 \underline{\widehat P}^{V}(s'\mid s,a)
 &=
 \widehat P^0(s'\mid s,a)
 \widehat h_{s,a}^{\widehat\lambda^*,V}(s').
\end{align}
Define the minimum positive empirical transition probability as
\begin{equation}
 \widehat p_\wedge
 =
 \min_{\substack{s,a,s':\\\widehat P^0(s'\mid s,a)>0}}
 \widehat P^0(s'\mid s,a).
\end{equation}

The normalization and stationarity arguments in
Lemma~\ref{lm:hp} give
\begin{align}
 \mathbb E_{s'\sim\widehat P^0_{s,a}}
 \bigl[\widehat h_{s,a}^{\widehat\lambda^*,V}(s')\bigr]
 &=1,\\
 \operatorname{KL}
 \bigl(\underline{\widehat P}^{V}_{s,a}
       \|\widehat P^0_{s,a}\bigr)
 &=\delta.
\end{align}
Thus, for each realization of the transition samples,
Lemma~\ref{lem:polished-variance} applies with
$P^0_{s,a}$, $\lambda^*$, and $p_\wedge$ replaced by
$\widehat P^0_{s,a}$, $\widehat\lambda^*$, and
$\widehat p_\wedge$, respectively.

On the event in \eqref{eq:51}, we further have
$\widehat p_\wedge\geq(3/4)p_\wedge$.
Consequently, on this event,
\begin{align}
 &\widehat\lambda^*
 \sqrt{\operatorname{Var}_{\widehat P^0_{s,a}}
 \bigl(\widehat h_{s,a}^{\widehat\lambda^*,V}\bigr)}
 \nonumber\\
 &\leq
 8\sqrt{
 \left(
 1+\min\left\{
   \exp\!\left(\frac{\delta}{\widehat p_\wedge}\right),
   \frac{\sqrt{\delta}}{\widehat p_\wedge},
   \frac{1}{\widehat p_\wedge}
 \right\}
 \right)
 \operatorname{Var}_{\underline{\widehat P}^{V}_{s,a}}(V)}
 +4\delta^{1/4}\operatorname{span}(V)\\
 &\leq
 8\sqrt{
 \left(
 1+\min\left\{
   \exp\!\left(\frac{4\delta}{3p_\wedge}\right),
   \frac{4\sqrt{\delta}}{3p_\wedge},
   \frac{4}{3p_\wedge}
 \right\}
 \right)
 \operatorname{Var}_{\underline{\widehat P}^{V}_{s,a}}(V)}
 +4\delta^{1/4}\operatorname{span}(V).
\end{align}
The second inequality follows from
$1/\widehat p_\wedge\leq4/(3p_\wedge)$ and the monotonicity
of each term inside the minimum.

This variance bound holds deterministically for each realization
of the empirical model. Here,
$\widehat h_{s,a}^{\widehat\lambda^*,V}$ is normalized under
$\widehat P^0_{s,a}$, whereas
$h_{s,a}^{\widehat\lambda^*,V}$ in \eqref{eq:bound} is normalized
under $P^0_{s,a}$. Applying the present bound in the statistical
error analysis therefore requires a comparison of these
normalizations and the corresponding variances, together with
a valid concentration argument for quantities selected using
the samples.

\subsection{A resolvent bound for value variance}\label{sec:c6}
\begin{lemma}\label{lem:polished-resolvent}
Let $0<\gamma<1$, let $P$ be a transition matrix on $\mathcal S$,
and let $r\in[0,1]^{|\mathcal S|}$ be a reward vector.
Let $V$ be the unique solution to the Bellman equation: $V=r+\gamma PV$, and define
\[
\operatorname{Var}_P(V)=P(V^2)-(PV)^2.
\]
Then
\begin{equation}\label{eq:polished-resolvent}
\left\|(I-\gamma P)^{-1}
\sqrt{\operatorname{Var}_P(V)}\right\|_\infty
\leq
\frac{\sqrt{1+\gamma}}{2\gamma(1-\gamma)^{3/2}},
\end{equation}
where the squares and square root are taken componentwise.
\end{lemma}

\begin{proof}
Let $(s_t)_{t\geq0}$ be a Markov chain with transition matrix $P$,
and define the discounted return
\[
G=\sum_{t=0}^{\infty}\gamma^t r(s_t).
\]
The Bellman equation implies that
$\mathbb E[G\mid s_0=s]=V(s)$.
Write
\[
W(s)=\operatorname{Var}(G\mid s_0=s),
\qquad
u=\operatorname{Var}_P(V).
\]
Since $r(s_0)$ is deterministic conditional on $s_0$, the Markov
property and the law of total variance give
\begin{align}
W(s)
&=\gamma^2\mathbb E[W(s_1)\mid s_0=s]
  +\gamma^2\operatorname{Var}(V(s_1)\mid s_0=s).
\end{align}
Equivalently, $W=\gamma^2PW+\gamma^2u$, and hence
\[
W=\gamma^2(I-\gamma^2P)^{-1}u.
\]

Because $0\leq r\leq1$, the return satisfies
$0\leq G\leq(1-\gamma)^{-1}$. Therefore, for every state $s$,
\[
W(s)
\leq
\mathbb E\!\left[
\left(G-\frac{1}{2(1-\gamma)}\right)^2
\middle|s_0=s
\right]
\leq\frac{1}{4(1-\gamma)^2}.
\]
It follows that
\begin{equation}\label{eq:return-variance-bound}
(I-\gamma^2P)^{-1}u
\leq\frac{\mathbf1}{4\gamma^2(1-\gamma)^2},
\end{equation}
where the inequality is componentwise.

Next, the matrix inverse identity gives
\begin{align}
(I-\gamma P)^{-1}
&=(I-\gamma^2P)^{-1} \nonumber\\
&\quad+\gamma(1-\gamma)
(I-\gamma P)^{-1}P(I-\gamma^2P)^{-1}.
\end{align}
The inverse matrices have nonnegative entries, and
$P\mathbf1=\mathbf1$ implies
\[
(I-\gamma P)^{-1}P\mathbf1
=\frac{\mathbf1}{1-\gamma}.
\]
Applying the matrix identity to $u$ and using
\eqref{eq:return-variance-bound}, we obtain
\begin{align}
(I-\gamma P)^{-1}u
&\leq
\frac{\mathbf1+
\gamma(1-\gamma)(I-\gamma P)^{-1}P\mathbf1}
     {4\gamma^2(1-\gamma)^2} \nonumber\\
&=
\frac{1+\gamma}{4\gamma^2(1-\gamma)^2}\mathbf1.
\label{eq:propagated-variance-bound}
\end{align}

Finally, write $R=(I-\gamma P)^{-1}$.
Its entries are nonnegative, and each row sums to
$(1-\gamma)^{-1}$. Thus, for every state $s$, the
Cauchy--Schwarz inequality gives
\begin{align}
\bigl[(R\sqrt{u})(s)\bigr]^2
&=
\left(\sum_{s'}R(s,s')\sqrt{u(s')}\right)^2
\nonumber\\
&\leq
\left(\sum_{s'}R(s,s')\right)
\left(\sum_{s'}R(s,s')u(s')\right)
\nonumber\\
&=\frac{(Ru)(s)}{1-\gamma}
\nonumber\\
&\leq
\frac{1+\gamma}{4\gamma^2(1-\gamma)^3},
\end{align}
where the last inequality follows from
\eqref{eq:propagated-variance-bound}.
Taking square roots and maximizing over $s$ proves
\eqref{eq:polished-resolvent}.
\end{proof}

The bound requires $V=r+\gamma PV$ with
$r\in[0,1]^{|\mathcal S|}$, where the same matrix $P$ appears
in both the resolvent and $\operatorname{Var}_P(V)$.
When $P$ is an empirical worst-case transition matrix and $V$
is a population robust value function, this Bellman equation
need not hold. Applying the bound to such a pair therefore
requires an additional comparison argument.

% This lemma applies when $V$ is the value function for the same
% transition matrix that appears in the resolvent. Applying it to an
% empirical worst-case matrix and a population value requires a separate
% comparison; the two matrices cannot be interchanged directly.

\subsection{Sample-complexity analysis for small uncertainty level: $\delta\le 1$.}
\label{sec:small-delta-completion}\label{sec:c7}

\begin{theorem}[Policy guarantee for $0<\delta\le1$]
\label{thm:small-delta-policy}
Let $0<\gamma<1$ and $\delta,\eta\in(0,1)$.
Suppose that rewards are known and belong to $[0,1]$,
and $N$ samples are drawn independently from the nominal transition
distribution for every state--action pair, with independence between
the samples from different pairs. Let $\widehat\pi$ be
the output policy after $T$ iterations of robust value iteration with
$Q_0=0$, using the empirical transition kernel and a fixed rule to
choose among actions attaining the same maximum. If
\begin{align}
 N\geq\max\Bigg\{&\frac{192}{p_\wedge}
       \log\frac{36|\mathcal S|^2|\mathcal A|N}{\eta p_\wedge},
       \nonumber\\
 &\frac{65536}{(1-\gamma)^2}
 \left(1+\min\left\{e^{2\delta/p_\wedge},
       \frac{2\sqrt\delta}{p_\wedge}+1,\frac2{p_\wedge}\right\}\right)
       \log\frac{36|\mathcal S|^2|\mathcal A|N}{\eta p_\wedge}
       \Bigg\},
 \label{eq:small-delta-policy-sample-floor}
\end{align}
then, with probability at least $1-\eta$,
\begin{align}
 \|V_{\pi^*}-V_{\widehat\pi}\|_\infty
 &\leq1024
 \sqrt{\frac{1}{N}
       \log\frac{36|\mathcal S|^2|\mathcal A|N}{\eta p_\wedge}}
       \left(\frac{\sqrt{\min\left\{e^{2\delta/p_\wedge},
       \frac{2\sqrt\delta}{p_\wedge}+1,\frac2{p_\wedge}\right\}}}{(1-\gamma)^{3/2}}
                   +\frac{\delta^{1/4}}{(1-\gamma)^2}\right)
       \nonumber\\
 &\quad+\frac{512}{p_\wedge N(1-\gamma)^2}
       \log\frac{36|\mathcal S|^2|\mathcal A|N}{\eta p_\wedge}
       +\frac{8\gamma^T}{(1-\gamma)^2}.
 \label{eq:polished-conditional-rate}
\end{align}
\end{theorem}
\begin{proof}
Work on the event of Lemma~\ref{lem:small-delta-uniform}.
Nominal and empirical optimal policies are denoted by $\pi^*$
and $\widehat\pi^*$, respectively. Every value function below belongs
to $[0,(1-\gamma)^{-1}]^{|\mathcal S|}$.

\textbf{One-step estimation error.}\par\nopagebreak[4]

Fix $(s,a)$ and a value vector $V$ listed in
Lemma~\ref{lem:small-delta-uniform}.
For $\widehat\lambda^*>0$, the relative probability bound
\eqref{eq:original-small-delta-relative} implies
\begin{align*}
 &\widehat\lambda^*
 \sqrt{\operatorname{Var}_{P^0_{s,a}}
       (h_{s,a}^{\widehat\lambda^*,V})}
 \leq\frac5{2\sqrt3}\widehat\lambda^*
 \sqrt{\operatorname{Var}_{\widehat P^0_{s,a}}
       (\widehat h_{s,a}^{\widehat\lambda^*,V})}.
\end{align*}
Indeed, the relative probability bound gives
$\operatorname{Var}_{P^0_{s,a}}(h_{s,a}^{\widehat\lambda^*,V})
\leq\tfrac43\operatorname{Var}_{\widehat P^0_{s,a}}
(h_{s,a}^{\widehat\lambda^*,V})$.
Moreover, $h_{s,a}^{\widehat\lambda^*,V}$ equals
$\widehat h_{s,a}^{\widehat\lambda^*,V}$ multiplied by the ratio
of the empirical to nominal normalizing expectations, which is
at most $5/4$.
Apply Lemma~\ref{lem:polished-variance} to the nominal and empirical
distributions at their respective positive dual optimizers, using
$\widehat p_\wedge\geq(3/4)p_\wedge$ for the empirical distribution.
Combining these bounds with the preceding comparison and
\eqref{eq:uniform-transition-concentration} gives, for every required
$V$ and $(s,a)$, equations (83), (106), and (107), together with
$|\log x|\leq(4/3)|x-1|$ on $[3/4,5/4]$, give
\begin{align}
 &|f(\widehat P^0_{s,a},V,\lambda)-f(P^0_{s,a},V,\lambda)|
 \leq\sqrt{\frac1N
 \log\frac{36|\mathcal S|^2|\mathcal A|N}{\eta p_\wedge}}
 \min\Bigg\{
 4\lambda\sqrt{\operatorname{Var}_{P^0_{s,a}}
                         (h_{s,a}^{\lambda,V})},
 \nonumber\\
 &\hspace{42mm}
 \frac{11}{2}\lambda\sqrt{\operatorname{Var}_{\widehat P^0_{s,a}}
                         (\widehat h_{s,a}^{\lambda,V})}
 \Bigg\}
 +\frac{54}{p_\wedge N(1-\gamma)}
 \log\frac{36|\mathcal S|^2|\mathcal A|N}{\eta p_\wedge}.
 \label{eq:short-objective}
\end{align}
The normalization comparison in (83) and (107) contributes
$5/(2\sqrt3)$ to the empirical coefficient. Terms at $\lambda=0$
use continuity; scalar weights follow (133).

For every $\lambda>0$,
\begin{align}
 &\lambda\sqrt{\operatorname{Var}_{P^0_{s,a}}
                         (h_{s,a}^{\lambda,V})}
 \leq
 3\lambda^*\sqrt{\operatorname{Var}_{P^0_{s,a}}
                         (h_{s,a}^{\lambda^*,V})}
 +\frac{3\delta^{1/4}}{1-\gamma}+
 \sqrt{\frac{2}{p_\wedge(1-\gamma)}
 \left[\underline P^V_{s,a}V-f(P^0_{s,a},V,\lambda)\right]}.
 \label{eq:complete-temperature-transfer}
\end{align}
The bracket is nonnegative.

Since $x(\log x)^2\leq(x-1)^2$, normalization and the
exponential-tilt identity imply
\begin{equation}
 \sqrt{\operatorname{Var}_{\underline P^V_{s,a}}(V)}
 \leq\lambda^*
 \sqrt{\operatorname{Var}_{P^0_{s,a}}(h_{s,a}^{\lambda^*,V})}.
 \label{eq:complete-reverse-transfer}
\end{equation}
Both sides vanish when $\lambda^*=0$.

For $\lambda\geq\lambda^*>0$, Lemma C.1 gives
$h_{s,a}^{\lambda,V}(\underline P^V_{s,a}V)\leq e^\delta
=h_{s,a}^{\lambda^*,V}(\underline P^V_{s,a}V)$.
Below $\underline P^V_{s,a}V$, monotonicity of
$\lambda(e^{x/\lambda}-1)$ bounds the centered weight difference, multiplied by $\lambda$,
by $\lambda^*|h_{s,a}^{\lambda^*,V}-1|$; above it, the
mean-value theorem bounds that difference by
$e^\delta|V-\underline P^V_{s,a}V|$.
Centering the variance at
$h_{s,a}^{\lambda,V}(\underline P^V_{s,a}V)$ and applying (141)
and \eqref{eq:complete-reverse-transfer} therefore give
\begin{align*}
 \lambda\sqrt{\operatorname{Var}_{P^0_{s,a}}(h_{s,a}^{\lambda,V})}
 &\leq\sqrt{1+e^{2\delta}}\,
 \lambda^*\sqrt{\operatorname{Var}_{P^0_{s,a}}(h_{s,a}^{\lambda^*,V})}
 +\frac{e^\delta(\delta/2)^{1/4}}{1-\gamma}.
\end{align*}
Since $\sqrt{1+e^2}<3$, this proves
\eqref{eq:complete-temperature-transfer} in this case.
For $\lambda^*=0$, the first region contributes zero and the
same argument applies.

Next suppose that $0<\lambda<\lambda^*$ and
$\lambda\leq(1-\gamma)^{-1}$. The exponential-tilt identities give
\begin{align*}
 &\lambda\operatorname{KL}
 \left(\underline P^V_{s,a}\,\middle\Vert\,
                  P^0_{s,a}h_{s,a}^{\lambda,V}\right)\\
 &\qquad=\underline P^V_{s,a}V-f(P^0_{s,a},V,\lambda).
\end{align*}
The triangle inequality in $L^2(P^0_{s,a})$, followed by
Pinsker's inequality and $P^0(s'\mid s,a)\geq p_\wedge$ on the
support, yields
\begin{align*}
 &\lambda\sqrt{\operatorname{Var}_{P^0_{s,a}}
                         (h_{s,a}^{\lambda,V})}\\
 &\quad\leq\frac{\lambda}{\lambda^*}\,
 \lambda^*\sqrt{\operatorname{Var}_{P^0_{s,a}}
                         (h_{s,a}^{\lambda^*,V})}\\
 &\qquad+\sqrt{\frac{2\lambda}{p_\wedge}
 \left[\underline P^V_{s,a}V-f(P^0_{s,a},V,\lambda)\right]}.
\end{align*}
This proves \eqref{eq:complete-temperature-transfer} in this case.
For $\lambda>(1-\gamma)^{-1}$, the weight is at most $e$;
the mean-value theorem, (141), and
\eqref{eq:complete-reverse-transfer} bound the left side by the
first two terms in \eqref{eq:complete-temperature-transfer}.
This completes the estimate, whose empirical version uses
$\widehat p_\wedge\geq3p_\wedge/4$.

Retain the dual optimization loss:
\begin{align*}
 &\underline{\widehat P}^V_{s,a}V-\underline P^V_{s,a}V\\
 &=f(\widehat P^0_{s,a},V,\widehat\lambda^*)
       -f(P^0_{s,a},V,\widehat\lambda^*)\\
 &\quad-\left[\underline P^V_{s,a}V
                   -f(P^0_{s,a},V,\widehat\lambda^*)\right].
\end{align*}
Apply \eqref{eq:short-objective} and
\eqref{eq:complete-temperature-transfer}, and complete the square
in the square root of the bracket. For the opposite sign,
evaluate the empirical objective at $\lambda^*$.
Interchanging the centers gives the empirical bound; its quadratic
remainder has coefficient $121/6$, compared with $8$ for the
nominal bound. Since $54+121/6<75$, both yield
\begin{align}
 &\left|\underline{\widehat P}^V_{s,a}V-\underline P^V_{s,a}V\right|
 \nonumber\\
 &\quad\leq\frac{33}{2}\sqrt{\frac1N
 \log\frac{36|\mathcal S|^2|\mathcal A|N}{\eta p_\wedge}}
 \Bigg[\min\Bigg\{
 \lambda^*\sqrt{\operatorname{Var}_{P^0_{s,a}}(h_{s,a}^{\lambda^*,V})},
 \nonumber\\
 &\hspace{45mm}
 \widehat\lambda^*\sqrt{\operatorname{Var}_{\widehat P^0_{s,a}}
 (\widehat h_{s,a}^{\widehat\lambda^*,V})}\Bigg\}
 +\frac{\delta^{1/4}}{1-\gamma}\Bigg]
 \nonumber\\
 &\qquad+\frac{75}{p_\wedge N(1-\gamma)}
 \log\frac{36|\mathcal S|^2|\mathcal A|N}{\eta p_\wedge}.
 \label{eq:complete-influence}
\end{align}

\textbf{Comparison at different value vectors.}\par\nopagebreak[4]

For every $V,W\in[0,(1-\gamma)^{-1}]^{|\mathcal S|}$,
\begin{align}
 &\lambda^*\sqrt{\operatorname{Var}_{P^0_{s,a}}
                         (h_{s,a}^{\lambda^*,V})}
 \nonumber\\
 &\quad\leq5\sqrt{1+\min\left\{
 e^{2\delta/p_\wedge},\,
 1+\frac{2\sqrt\delta}{p_\wedge},\,
 \frac2{p_\wedge}\right\}}
 \left[\sqrt{\operatorname{Var}_{\underline P^W_{s,a}}(W)}
       +2\|V-W\|_\infty\right]
 +\frac{7\delta^{1/4}}{1-\gamma}.
 \label{eq:complete-value-influence}
\end{align}
The same estimate holds with empirical centers, weights, and
worst-case rows.

By Lemma C.10 and $\widehat p_\wedge\geq3p_\wedge/4$
from (83), the maximum optimized density for either center is
bounded by
\begin{equation}
 e\left(1+\min\left\{
 e^{2\delta/p_\wedge},\,
 1+\frac{2\sqrt\delta}{p_\wedge},\,
 \frac2{p_\wedge}\right\}\right).
 \label{eq:complete-density}
\end{equation}
We use the refined variance-transfer estimate (143) from the
proof of Lemma C.11, with $\operatorname{span}(V)\leq(1-\gamma)^{-1}$.
At a zero optimizer the optimized influence and value variance
are zero, and any density is interpreted by its zero-temperature limit.

Consider $V+t(W-V)$, $0\leq t\leq1$, and let $\lambda_t^*$
be its dual optimizer. Wherever this optimizer is positive and
the value variance is positive, differentiating the active KL
constraint gives
\begin{align*}
 \frac{d\lambda_t^*}{dt}
 =\lambda_t^*\,
 \frac{\operatorname{Cov}_{\underline P^{V+t(W-V)}_{s,a}}
                     (V+t(W-V),W-V)}
      {\operatorname{Var}_{\underline P^{V+t(W-V)}_{s,a}}
                     (V+t(W-V))}.
\end{align*}
Differentiate the $L^2(P^0_{s,a})$ norm of
$\lambda_t^*(h_{s,a}^{\lambda_t^*,\,V+t(W-V)}-1)$.
Cauchy--Schwarz and the preceding identity yield
\begin{align}
 &\left|\frac d{dt}\left[
 \lambda_t^*\sqrt{\operatorname{Var}_{P^0_{s,a}}
       (h_{s,a}^{\lambda_t^*,\,V+t(W-V)})}\right]\right|
 \nonumber\\
 &\quad\leq
 \left[
 \frac{\lambda_t^*\sqrt{\operatorname{Var}_{P^0_{s,a}}
       (h_{s,a}^{\lambda_t^*,\,V+t(W-V)})}}
      {\sqrt{\operatorname{Var}_{\underline P^{V+t(W-V)}_{s,a}}
       (V+t(W-V))}}
 +2\sqrt{\max_{s'}h_{s,a}^{\lambda_t^*,\,V+t(W-V)}(s')}
 \right]\|W-V\|_\infty.
 \label{eq:complete-influence-derivative}
\end{align}
Here the weight derivative contributes one maximum-density term;
the optimizer derivative contributes the ratio and the second
maximum-density term.

Whenever the influence on the left of
equation~(143) exceeds
$7\delta^{1/4}/(1-\gamma)$, its ratio to the worst-case standard
deviation is at most
\[
 \frac32\left(
 \sqrt{\max_{s'}h_{s,a}^{\lambda_t^*,\,V+t(W-V)}(s')}
 +e^\delta\right),
\]
because $3e^\delta2^{-1/4}<7$.
Equations \ref{eq:complete-density} and
\eqref{eq:complete-influence-derivative}, together with
$(7/2)\sqrt e+(3/2)e<10$, therefore bound the derivative in
that region by
\[
 10\sqrt{1+\min\left\{
 e^{2\delta/p_\wedge},\,
 1+\frac{2\sqrt\delta}{p_\wedge},\,
 \frac2{p_\wedge}\right\}}\,\|W-V\|_\infty.
\]
The optimizer is unique in $[0,1/(\delta(1-\gamma))]$.
The influence is continuous, including at zero optimizers, since
the density is bounded by the reciprocal of the minimum positive
base probability. Hence clipping it from below at
$7\delta^{1/4}/(1-\gamma)$ gives the same Lipschitz bound on
the entire segment. At $W$, (143), $\sqrt e+e<5$, and
$e2^{-1/4}<7$ give the endpoint bound in
\eqref{eq:complete-value-influence}, completing the estimate for both centers.

\textbf{Bellman comparisons.}\par\nopagebreak[4]

Monotonicity, contraction, and $Q_0=0$ give
$0\leq Q_T\leq\widehat Q^*$ and
$\|\widehat Q^*-Q_T\|_\infty\leq\gamma^T/(1-\gamma)$.
Since $\widehat\pi$ is greedy for $Q_T$, its optimal empirical
Bellman residual satisfies
\[
 0\leq\widehat V_{\widehat\pi^*}
 -r(\cdot,\widehat\pi(\cdot))
 -\gamma\underline{\widehat P}^{\widehat V_{\widehat\pi^*},\widehat\pi}
 \widehat V_{\widehat\pi^*}
 \leq\frac{\gamma^T}{1-\gamma}\mathbf1.
\]
Subtract the population Bellman equation for $V_{\widehat\pi}$
and use the nominal minimizing property to obtain
\begin{align}
 &\widehat V_{\widehat\pi^*}-V_{\widehat\pi}
 \nonumber\\
 &\quad\leq
 (I-\gamma\underline P^{V_{\widehat\pi},\widehat\pi})^{-1}
 \Bigg[\frac{\gamma^T}{1-\gamma}\mathbf1
 \nonumber\\
 &\hspace{32mm}+\gamma\left(
 \underline{\widehat P}^{\widehat V_{\widehat\pi^*},\widehat\pi}
 -\underline P^{\widehat V_{\widehat\pi^*},\widehat\pi}\right)
                         \widehat V_{\widehat\pi^*}\Bigg].
 \label{eq:complete-bellman-three}
\end{align}

\textbf{Propagation and absorption.}\par\nopagebreak[4]

Apply \eqref{eq:complete-influence} and
\eqref{eq:complete-value-influence} to these two comparisons and
\eqref{eq:complete-bellman-three}
with the respective pairs
\[
 (\widehat V_{\pi^*},V_{\pi^*}),\qquad
 (V_{\pi^*},\widehat V_{\pi^*}),\qquad
 (\widehat V_{\widehat\pi^*},V_{\widehat\pi}).
\]
Choose the nominal term in the first and third comparisons
and the empirical term in the second. The resulting coefficients
are bounded by $83$, $165$, $132$, and $75$, using
$(33/2)\cdot5<83$, $(33/2)\cdot10=165$, and
$(33/2)(7+1)=132$. Lemma C.12 bounds
each matched variance after discounted propagation by
$1/(\sqrt2(1-\gamma)^{3/2})$; the inverses map $\mathbf1$ to
$\mathbf1/(1-\gamma)$. Population and empirical optimality give
$V_{\widehat\pi}-\widehat V_{\widehat\pi^*}
\leq V_{\pi^*}-\widehat V_{\pi^*}$, so the two signed
comparisons for $\pi^*$ also control the negative part of the
third comparison.
Taking the maximum yields
\begin{align}
 &\max\left\{
 \|V_{\pi^*}-\widehat V_{\pi^*}\|_\infty,
 \|\widehat V_{\widehat\pi^*}-V_{\widehat\pi}\|_\infty
 \right\}
 \nonumber\\
 &\quad\leq
 \sqrt{\frac{1+\min\left\{
 e^{2\delta/p_\wedge},\,
 1+\dfrac{2\sqrt\delta}{p_\wedge},\,
 \dfrac2{p_\wedge}\right\}}N
 \log\frac{36|\mathcal S|^2|\mathcal A|N}{\eta p_\wedge}}
 \nonumber\\
 &\qquad\cdot\Bigg[\frac{83}{\sqrt2(1-\gamma)^{3/2}}
 +\frac{165}{1-\gamma}\max\left\{
 \|V_{\pi^*}-\widehat V_{\pi^*}\|_\infty,
 \|\widehat V_{\widehat\pi^*}-V_{\widehat\pi}\|_\infty
 \right\}\Bigg]
 \nonumber\\
 &\qquad+\frac{132\delta^{1/4}}{(1-\gamma)^2}
 \sqrt{\frac1N
 \log\frac{36|\mathcal S|^2|\mathcal A|N}{\eta p_\wedge}}
 +\frac{75}{p_\wedge N(1-\gamma)^2}
 \log\frac{36|\mathcal S|^2|\mathcal A|N}{\eta p_\wedge}
 +\frac{\gamma^T}{(1-\gamma)^2}.
 \label{eq:complete-bootstrap}
\end{align}
The coefficient of the repeated error
maximum in \eqref{eq:complete-bootstrap} is at most
$165/256<1$. Moving it to the left multiplies the remaining
terms by at most $256/91$.

Empirical optimality bounds the policy loss by
twice this error maximum, giving a total multiplier $512/91$.
Since
\[
 \frac{512\cdot83}{91}<1024,\qquad
 \frac{512\cdot132}{91}<1024,\qquad
 \frac{512\cdot75}{91}<512,\qquad
 \frac{512}{91}<8.
\]
the claimed bound follows on the event of probability at least $1-\eta$.
\end{proof}

\subsection{Sample-complexity analysis for large uncertainty level: $\delta>1$}
\label{sec:large-delta-completion}
\begin{lemma}[Variance bound for $\delta>1$]
\label{lem:large-delta-variance}
Fix $(s,a)$ and $\delta>1$. Suppose every positive probability in
$P^0_{s,a}$ is at least $p_\wedge$, and let $\lambda^*$ maximize
$f(P^0_{s,a},V,\lambda)$ over $\lambda\geq0$.
Use the worst-case transition $\underline P^V_{s,a}$ and weight
$h_{s,a}^{\lambda^*,V}$ defined in~\eqref{eq: weight}.
Then
\begin{align}
 &(\lambda^*)^2\operatorname{Var}_{P^0_{s,a}}
(h_{s,a}^{\lambda^*,V})\leq\frac2{p_\wedge}\operatorname{Var}_{\underline P^V_{s,a}}(V)
       +8e^\delta\operatorname{span}(V)^2.
 \label{eq:large-delta-combined-variance}
\end{align}
At $\lambda^*=0$, interpret the left-hand side as its limit as
$\lambda\downarrow0$. That limit is zero, so
\eqref{eq:large-delta-combined-variance} remains valid.
\end{lemma}
\begin{proof}
Suppose first that $\lambda^*>0$. The dual first-order condition
and the definition of $h_{s,a}^{\lambda^*,V}$ imply
\begin{align}
 \log h_{s,a}^{\lambda^*,V}
 &=\delta-\frac{V-\mathbb E_{\underline P^V_{s,a}}[V]}{\lambda^*},
 \nonumber\\
 \mathbb E_{\underline P^V_{s,a}}
       [\log h_{s,a}^{\lambda^*,V}]&=\delta,
 \label{eq:large-delta-stationarity}\\
 \operatorname{Var}_{P^0_{s,a}}(h_{s,a}^{\lambda^*,V})
 &=\mathbb E_{\underline P^V_{s,a}}[h_{s,a}^{\lambda^*,V}]-1.
 \label{eq:large-delta-density-identities}
\end{align}
The last identity uses
$\mathbb E_{P^0_{s,a}}[h_{s,a}^{\lambda^*,V}]=1$ and
$P^0_{s,a}h_{s,a}^{\lambda^*,V}=\underline P^V_{s,a}$.
Moreover, $h_{s,a}^{\lambda^*,V}\leq p_\wedge^{-1}$ on the
nominal support. Jensen's inequality applied to the dual normalizer gives
\begin{align}
 \lambda^*\delta
 &=-\lambda^*\log\mathbb E_{P^0_{s,a}}[e^{-V/\lambda^*}]
                     -\mathbb E_{\underline P^V_{s,a}}[V]
 \nonumber\\
 &\leq\mathbb E_{P^0_{s,a}}[V]
                     -\mathbb E_{\underline P^V_{s,a}}[V]
 \leq\operatorname{span}(V).
 \label{eq:large-delta-multiplier-span}
\end{align}

Define
\[
 \psi=\log\mathbb E_{\underline P^V_{s,a}}
                   [h_{s,a}^{\lambda^*,V}]-\delta.
\]
Jensen's inequality and~\eqref{eq:large-delta-stationarity} give
$\psi\geq0$. We consider the following two cases.

\noindent\textbf{Case 1: $\psi>2$.}
Then $\mathbb E_{\underline P^V_{s,a}}[h_{s,a}^{\lambda^*,V}]
=e^{\delta+\psi}>e^{\delta+2}$.
Splitting this expectation at half its value gives
\begin{align*}
 &\mathbb E_{\underline P^V_{s,a}}[h_{s,a}^{\lambda^*,V}]\\
 &\leq\frac12\mathbb E_{\underline P^V_{s,a}}
                         [h_{s,a}^{\lambda^*,V}]
 +\frac1{p_\wedge}\Pr_{\underline P^V_{s,a}}\!\left(
 h_{s,a}^{\lambda^*,V}\geq
 \frac12\mathbb E_{\underline P^V_{s,a}}
                         [h_{s,a}^{\lambda^*,V}]\right).
\end{align*}
Hence
\begin{equation}\label{eq:large-delta-moment-tail}
 \Pr_{\underline P^V_{s,a}}\!\left(
 h_{s,a}^{\lambda^*,V}\geq
 \frac12\mathbb E_{\underline P^V_{s,a}}
                         [h_{s,a}^{\lambda^*,V}]\right)
 \geq\frac{p_\wedge}{2}
       \mathbb E_{\underline P^V_{s,a}}[h_{s,a}^{\lambda^*,V}].
\end{equation}
On this event,
$\log h_{s,a}^{\lambda^*,V}-\delta\geq\psi-\log2>1$.
By~\eqref{eq:large-delta-stationarity},
\begin{align*}
 \frac{\operatorname{Var}_{\underline P^V_{s,a}}(V)}{(\lambda^*)^2}
 &=\mathbb E_{\underline P^V_{s,a}}
              [(\log h_{s,a}^{\lambda^*,V}-\delta)^2]\\
 &\geq\Pr_{\underline P^V_{s,a}}\!\left(
 h_{s,a}^{\lambda^*,V}\geq
 \frac12\mathbb E_{\underline P^V_{s,a}}
                         [h_{s,a}^{\lambda^*,V}]\right)\\
 &\geq\frac{p_\wedge}{2}
       \mathbb E_{\underline P^V_{s,a}}[h_{s,a}^{\lambda^*,V}].
\end{align*}
Combining this with~\eqref{eq:large-delta-density-identities}
bounds $(\lambda^*)^2\operatorname{Var}_{P^0_{s,a}}
(h_{s,a}^{\lambda^*,V})$ by
$2\operatorname{Var}_{\underline P^V_{s,a}}(V)/p_\wedge$.

\noindent\textbf{Case 2: $0\leq\psi\leq2$.}
Since $\mathbb E_{\underline P^V_{s,a}}[h_{s,a}^{\lambda^*,V}]
=e^{\delta+\psi}\leq e^{\delta+2}$, we obtain
\begin{align*}
 (\lambda^*)^2\operatorname{Var}_{P^0_{s,a}}
                  (h_{s,a}^{\lambda^*,V})
 &\leq(\lambda^*)^2(e^{\delta+2}-1)\\
 &\leq\frac{e^{\delta+2}-1}{\delta^2}\operatorname{span}(V)^2
 \leq8e^\delta\operatorname{span}(V)^2,
\end{align*}
where the second inequality follows from
\eqref{eq:large-delta-multiplier-span}, and the last uses
$\delta>1$ and $e^2<8$.
In either case, the sum of the two nonnegative terms on the
right-hand side of~\eqref{eq:large-delta-combined-variance}
bounds the left-hand side.

For every $\lambda>0$, the nominally normalized weight satisfies
$h_{s,a}^{\lambda,V}\leq p_\wedge^{-1}$, and therefore
\[
 0\leq\lambda^2\operatorname{Var}_{P^0_{s,a}}
                  (h_{s,a}^{\lambda,V})
 \leq\lambda^2(p_\wedge^{-1}-1)\longrightarrow0
 \qquad\text{as }\lambda\downarrow0.
\]
If $\lambda^*=0$, strong duality gives
\[
 \mathbb E_{\underline P^V_{s,a}}[V]
 =f(P^0_{s,a},V,0)
 =\min_{s':P^0(s'\mid s,a)>0}V(s').
\]
Finite KL divergence restricts the worst-case distribution to the
nominal support. The displayed equality therefore forces it to
assign mass only to minimizing states, so
$\operatorname{Var}_{\underline P^V_{s,a}}(V)=0$.
This argument does not use stationarity at zero.
\end{proof}

\begin{lemma}[Uniform estimation for $\delta>1$]
\label{lem:large-delta-uniform}
Fix $\delta>1$ and $0<\gamma<1$. Suppose rewards are known and
in $[0,1]$, and $N$ samples are drawn independently from each
nominal transition $P^0_{s,a}$, independently across state--action
pairs. For $0<\eta<1$, assume
\begin{equation}\label{eq:large-delta-support-samples}
 N\geq\frac{192}{p_\wedge}
 \log\frac{36|\mathcal S|^2|\mathcal A|N}{\eta p_\wedge}.
\end{equation}
With probability at least $1-\eta$, the supports agree,
$3P^0/4\leq\widehat P^0\leq5P^0/4$, and
\eqref{eq:uniform-transition-concentration} holds simultaneously
for every $(s,a)$ and
$V\in\{V_{\pi^*},\widehat V_{\pi^*},
\widehat V_{\widehat\pi^*}\}$. On this event,
\begin{align}
 &\left|\underline{\widehat P}^{V}(\cdot\mid s,a)V
                    -\underline P^{V}(\cdot\mid s,a)V\right|
 \nonumber\\
 &\leq16\sqrt{\frac1{Np_\wedge}
 \log\frac{36|\mathcal S|^2|\mathcal A|N}{\eta p_\wedge}}
 \min\left\{
 \sqrt{\operatorname{Var}_{\underline P^V(\cdot\mid s,a)}(V)},
 \sqrt{\operatorname{Var}_{\underline{\widehat P}^V(\cdot\mid s,a)}(V)}
 \right\}
 \nonumber\\
 &\quad+21\sqrt{\frac{e^\delta}{N}
 \log\frac{36|\mathcal S|^2|\mathcal A|N}{\eta p_\wedge}}
             \operatorname{span}(V)
 \nonumber\\
 &\quad+\frac{652}{p_\wedge N(1-\gamma)}
 \log\frac{36|\mathcal S|^2|\mathcal A|N}{\eta p_\wedge}.
 \label{eq:large-delta-uniform-local}
\end{align}
\end{lemma}
\begin{proof}
The proof of Lemma~\ref{lem:small-delta-uniform} applies to each
fixed $\delta>1$: its support estimate, Bernstein inequality, and
covering grids do not use $\delta\le1$. The absorbing-state values
remain bounded by $(1-\gamma)^{-1}$ and satisfy the same Bellman
contraction. In particular, the grid argument is uniform in
$\lambda\geq0$ and covers both data-dependent optimizers.
Thus its stated support and concentration conclusions hold on a
single event of probability at least $1-\eta$.

For $\widehat\lambda^*>0$, the relative probability bounds give
\begin{equation}\label{eq:large-delta-normalizer-comparison}
 \frac34\leq
 \frac{\mathbb E_{\widehat P^0_{s,a}}[e^{-V/\widehat\lambda^*}]}
      {\mathbb E_{P^0_{s,a}}[e^{-V/\widehat\lambda^*}]}
 \leq\frac54.
\end{equation}
The nominal and empirical weights at this optimizer satisfy
\[
 h_{s,a}^{\widehat\lambda^*,V}
 =\frac{\mathbb E_{\widehat P^0_{s,a}}[e^{-V/\widehat\lambda^*}]}
       {\mathbb E_{P^0_{s,a}}[e^{-V/\widehat\lambda^*}]}
       \widehat h_{s,a}^{\widehat\lambda^*,V}.
\]
Since
$\mathbb E_{\widehat P^0_{s,a}}[
\widehat h_{s,a}^{\widehat\lambda^*,V}]=1$ and
$P^0_{s,a}\leq(4/3)\widehat P^0_{s,a}$,
\begin{align*}
 \operatorname{Var}_{P^0_{s,a}}
              (\widehat h_{s,a}^{\widehat\lambda^*,V})
 &\leq\mathbb E_{P^0_{s,a}}
              [(\widehat h_{s,a}^{\widehat\lambda^*,V}-1)^2]\\
 &\leq\frac43\mathbb E_{\widehat P^0_{s,a}}
              [(\widehat h_{s,a}^{\widehat\lambda^*,V}-1)^2]\\
 &=\frac43\operatorname{Var}_{\widehat P^0_{s,a}}
              (\widehat h_{s,a}^{\widehat\lambda^*,V}).
\end{align*}
Together with~\eqref{eq:large-delta-normalizer-comparison}, this yields
\begin{align}
 &\widehat\lambda^*\sqrt{\operatorname{Var}_{P^0_{s,a}}
                  (h_{s,a}^{\widehat\lambda^*,V})}\nonumber\\
 &\leq\frac5{2\sqrt3}\widehat\lambda^*
       \sqrt{\operatorname{Var}_{\widehat P^0_{s,a}}
                  (\widehat h_{s,a}^{\widehat\lambda^*,V})}.
 \label{eq:large-delta-empirical-variance-conversion}
\end{align}
Apply Lemma~\ref{lem:large-delta-variance} to the nominal and
empirical optimizers, using the probability floors $p_\wedge$
and $3p_\wedge/4$, respectively. Taking square roots and using
$\sqrt{x+y}\leq\sqrt{x}+\sqrt{y}$ for $x,y\geq0$ gives
\begin{align*}
 &\lambda^*\sqrt{\operatorname{Var}_{P^0_{s,a}}
                      (h_{s,a}^{\lambda^*,V})}\\
 &\leq\sqrt{\frac2{p_\wedge}}
       \sqrt{\operatorname{Var}_{\underline P^V_{s,a}}(V)}
       +\sqrt{8e^\delta}\operatorname{span}(V),\\
 &\widehat\lambda^*\sqrt{\operatorname{Var}_{P^0_{s,a}}
                      (h_{s,a}^{\widehat\lambda^*,V})}\\
 &\leq\frac5{2\sqrt3}\left(
       \sqrt{\frac8{3p_\wedge}}
       \sqrt{\operatorname{Var}_{\underline{\widehat P}^V_{s,a}}(V)}
       +\sqrt{8e^\delta}\operatorname{span}(V)\right),
\end{align*}
where the second estimate also uses
\eqref{eq:large-delta-empirical-variance-conversion}.
The empirical coefficients dominate the nominal ones. Substituting
both bounds into~\eqref{eq:uniform-transition-concentration} gives
\begin{align*}
 &\left|\underline{\widehat P}^{V}(\cdot\mid s,a)V
                    -\underline P^{V}(\cdot\mid s,a)V\right|\\
 &\leq\frac{20\sqrt2}{3}\sqrt{\frac1{Np_\wedge}
 \log\frac{36|\mathcal S|^2|\mathcal A|N}{\eta p_\wedge}}
 \max\left\{
 \sqrt{\operatorname{Var}_{\underline P^V(\cdot\mid s,a)}(V)},
 \sqrt{\operatorname{Var}_{\underline{\widehat P}^V(\cdot\mid s,a)}(V)}
 \right\}\\
 &\quad+20\sqrt{\frac{2e^\delta}{3N}
 \log\frac{36|\mathcal S|^2|\mathcal A|N}{\eta p_\wedge}}
             \operatorname{span}(V)\\
 &\quad+\frac{64}{p_\wedge N(1-\gamma)}
 \log\frac{36|\mathcal S|^2|\mathcal A|N}{\eta p_\wedge}.
\end{align*}

We next compare the two value variances in this estimate.
Suppose first that both optimizers are positive. The distribution
$P^0_{s,a}h_{s,a}^{\widehat\lambda^*,V}$ uses the nominal base
and the empirical optimizer. The relative probability bound in
Lemma~\ref{lem:small-delta-uniform} gives
\[
 \frac35\leq
 \frac{\underline{\widehat P}^{V}(s'\mid s,a)}
 {P^0(s'\mid s,a)h_{s,a}^{\widehat\lambda^*,V}(s')}
 \leq\frac53.
\]
The ratio equals $\widehat P^0(s'\mid s,a)/P^0(s'\mid s,a)$
divided by its expectation under
$P^0_{s,a}h_{s,a}^{\widehat\lambda^*,V}$.
For a positive function, its variance divided by the square of its
mean is at most the squared difference between its upper and lower
bounds divided by four times their product. This follows by
expanding the nonnegative product of its distances to these bounds
and completing the square. Since the base ratio differs from one
by at most the relative error in
\eqref{eq:original-small-delta-relative}, we obtain
\begin{align*}
 &\operatorname{Var}_{P^0_{s,a}h_{s,a}^{\widehat\lambda^*,V}}
 \!\left(\frac{\underline{\widehat P}^{V}(\cdot\mid s,a)}
 {P^0(\cdot\mid s,a)h_{s,a}^{\widehat\lambda^*,V}(\cdot)}\right)\\
 &\leq
 \frac{\dfrac{12}{Np_\wedge}
 \log\dfrac{36|\mathcal S|^2|\mathcal A|N}{\eta p_\wedge}}
 {1-\dfrac{12}{Np_\wedge}
 \log\dfrac{36|\mathcal S|^2|\mathcal A|N}{\eta p_\wedge}}
 \leq\frac{16}{15}\frac{12}{Np_\wedge}
 \log\frac{36|\mathcal S|^2|\mathcal A|N}{\eta p_\wedge}.
\end{align*}
The last inequality uses~\eqref{eq:large-delta-support-samples}.
Centering under $P^0_{s,a}h_{s,a}^{\widehat\lambda^*,V}$ and
using Cauchy--Schwarz therefore gives
\begin{align*}
 \operatorname{Var}_{\underline{\widehat P}^{V}_{s,a}}(V)
 &\leq\frac53
 \operatorname{Var}_{P^0_{s,a}h_{s,a}^{\widehat\lambda^*,V}}(V),\\
 \left|\mathbb E_{\underline{\widehat P}^{V}_{s,a}}[V]
       -\mathbb E_{P^0_{s,a}h_{s,a}^{\widehat\lambda^*,V}}[V]\right|
 &\leq\frac4{\sqrt{15}}\sqrt{\frac{12}{Np_\wedge}
 \log\frac{36|\mathcal S|^2|\mathcal A|N}{\eta p_\wedge}}
 \sqrt{\operatorname{Var}_{P^0_{s,a}h_{s,a}^{\widehat\lambda^*,V}}(V)}.
\end{align*}

The distributions $P^0_{s,a}h_{s,a}^{\widehat\lambda^*,V}$ and
$\underline P^V_{s,a}=P^0_{s,a}h_{s,a}^{\lambda^*,V}$
are exponential tilts of the same base. Their density ratio is
monotone in $V$, so their distributions of $V$ are stochastically
ordered. This follows directly from the covariance inequality
for an increasing function and a decreasing function.
The squared deviation from $\mathbb E_{\underline P^V_{s,a}}[V]$
is $2\operatorname{span}(V)$-Lipschitz on the range of $V$.
Adding or subtracting this squared deviation from
$2\operatorname{span}(V)V$ gives increasing functions.
Applying the expectation order to these two functions yields
\begin{align*}
 &\operatorname{Var}_{P^0_{s,a}h_{s,a}^{\widehat\lambda^*,V}}(V)\\
 &\leq\operatorname{Var}_{\underline P^V_{s,a}}(V)
 +2\operatorname{span}(V)
 \left|\mathbb E_{P^0_{s,a}h_{s,a}^{\widehat\lambda^*,V}}[V]
       -\mathbb E_{\underline P^V_{s,a}}[V]\right|\\
 &\leq\operatorname{Var}_{\underline P^V_{s,a}}(V)
 +2\operatorname{span}(V)
 \left|(\underline{\widehat P}^{V}_{s,a}-\underline P^V_{s,a})V\right|\\
 &\quad+\frac8{\sqrt{15}}\operatorname{span}(V)
 \sqrt{\frac{12}{Np_\wedge}
 \log\frac{36|\mathcal S|^2|\mathcal A|N}{\eta p_\wedge}}
 \sqrt{\operatorname{Var}_{P^0_{s,a}h_{s,a}^{\widehat\lambda^*,V}}(V)}.
\end{align*}
Solving this quadratic inequality for the last standard deviation,
and then using the preceding factor $5/3$, gives
\begin{align*}
 \sqrt{\operatorname{Var}_{\underline{\widehat P}^{V}_{s,a}}(V)}
 &\leq\sqrt{\frac53}
 \sqrt{\operatorname{Var}_{\underline P^V_{s,a}}(V)}
 +\sqrt{\frac{10}{3}\operatorname{span}(V)
 \left|(\underline{\widehat P}^{V}_{s,a}-\underline P^V_{s,a})V\right|}\\
 &\quad+\frac83\operatorname{span}(V)
 \sqrt{\frac{12}{Np_\wedge}
 \log\frac{36|\mathcal S|^2|\mathcal A|N}{\eta p_\wedge}}.
\end{align*}
The same inequality with the two worst-case variances interchanged
also holds. Use the empirical base with the nominal optimizer:
the reciprocal base ratios, after normalization, obey the same
pointwise and variance bounds, so the remaining argument is unchanged.

Substitute either comparison into the preceding transition estimate
and apply Young's inequality:
\begin{align*}
 &\frac{40}{3}\sqrt{\frac{5\operatorname{span}(V)}{3Np_\wedge}
 \log\frac{36|\mathcal S|^2|\mathcal A|N}{\eta p_\wedge}
 \left|(\underline{\widehat P}^{V}_{s,a}-\underline P^V_{s,a})V\right|}\\
 &\leq\frac15
 \left|(\underline{\widehat P}^{V}_{s,a}-\underline P^V_{s,a})V\right|
 +\frac{10000\operatorname{span}(V)}{27Np_\wedge}
 \log\frac{36|\mathcal S|^2|\mathcal A|N}{\eta p_\wedge}.
\end{align*}
Move one fifth of the absolute difference to the left. Since
$\operatorname{span}(V)\leq(1-\gamma)^{-1}$ and
\[
 \frac{25}{3}\sqrt{\frac{10}{3}}<16,\qquad
 25\sqrt{\frac23}<21,\qquad
 80+\frac{400\sqrt6}{9}+\frac{12500}{27}<652,
\]
we obtain
\begin{align*}
 &\left|(\underline{\widehat P}^{V}_{s,a}-\underline P^V_{s,a})V\right|
 \\
 &\leq16\sqrt{\frac1{Np_\wedge}
 \log\frac{36|\mathcal S|^2|\mathcal A|N}{\eta p_\wedge}}
 \min\left\{
 \sqrt{\operatorname{Var}_{\underline P^V_{s,a}}(V)},
 \sqrt{\operatorname{Var}_{\underline{\widehat P}^{V}_{s,a}}(V)}
 \right\}\\
 &\quad+21\sqrt{\frac{e^\delta}{N}
 \log\frac{36|\mathcal S|^2|\mathcal A|N}{\eta p_\wedge}}
 \operatorname{span}(V)\\
 &\quad+\frac{652}{Np_\wedge(1-\gamma)}
 \log\frac{36|\mathcal S|^2|\mathcal A|N}{\eta p_\wedge}.
\end{align*}
This proves~\eqref{eq:large-delta-uniform-local}.
When an optimizer equals zero, take the corresponding exponential
tilt limit. Its value equals the minimum on the common support
almost surely. Every worst-case distribution at that zero
optimizer has this same distribution of $V$, so all the mean
and variance inequalities above remain valid.
\end{proof}

\begin{theorem}[Policy guarantee for $\delta>1$]
\label{thm:large-delta-policy}
Suppose $0<\gamma<1$, $\delta>1$, $0<\eta<1$, rewards are known
and in $[0,1]$, and the sampling assumptions and sample condition
of Lemma~\ref{lem:large-delta-uniform} hold. Let $\widehat\pi$ be
obtained after $T$ iterations of empirical robust value iteration
from $Q_0=0$, with a fixed rule for choosing among maximizing actions.
On the event of Lemma~\ref{lem:large-delta-uniform}, whose probability
is at least $1-\eta$,
\begin{align}
 \|V_{\pi^*}-V_{\widehat\pi}\|_\infty
 &\leq34\sqrt{\frac1{Np_\wedge(1-\gamma)^3}
 \log\frac{36|\mathcal S|^2|\mathcal A|N}{\eta p_\wedge}}
 \nonumber\\
 &\quad+63\sqrt{\frac{e^\delta}{N(1-\gamma)^4}
 \log\frac{36|\mathcal S|^2|\mathcal A|N}{\eta p_\wedge}}
 \nonumber\\
 &\quad+\frac{2304\bigl(1+\log(4/(3p_\wedge))\bigr)}
 {Np_\wedge(1-\gamma)^3}
 \log\frac{36|\mathcal S|^2|\mathcal A|N}{\eta p_\wedge}
 +\frac{6\gamma^T}{(1-\gamma)^2}.
 \label{eq:large-delta-policy-bound}
\end{align}
If $\delta\geq\log(4/(3p_\wedge))$, then on the same event
\begin{equation}\label{eq:large-delta-support-policy}
 \|V_{\pi^*}-V_{\widehat\pi}\|_\infty
 \leq\frac{2\gamma^T}{(1-\gamma)^2}.
\end{equation}
\end{theorem}
\begin{proof}
Work on the event of Lemma~\ref{lem:large-delta-uniform}.
All values belong to $[0,(1-\gamma)^{-1}]^{|\mathcal S|}$.
The proof uses the value variances in
\eqref{eq:large-delta-uniform-local} before bounding their discounted
propagation.

For every fixed nominal row and
$V,W\in[0,(1-\gamma)^{-1}]^{|\mathcal S|}$, we first prove
\begin{align}
 &\left|\operatorname{Var}_{\underline P^V_{s,a}}(V)
             -\operatorname{Var}_{\underline P^W_{s,a}}(W)\right|
 \nonumber\\
 &\leq\frac{4\log(1/p_\wedge)}{1-\gamma}
                       \|V-W\|_\infty.
 \label{eq:large-delta-value-variance-comparison}
\end{align}
Suppose first that the dual optimizer at $V$ is positive, and let
$\lambda_t$ denote the dual optimizer at $V+t(W-V)$ for $t$ near
zero. The worst-case distribution is then unique and assigns
positive probability to every state in the nominal support.
Moreover, $\operatorname{Var}_{\underline P^V_{s,a}}(V)>0$:
otherwise $V$ would be constant on that support, and the dual
objective would be strictly decreasing for $\lambda>0$.
Differentiating the active constraint
\[
 \text{KL}\!\left(\underline P^{V+t(W-V)}_{s,a}
                         \,\middle\|\,P^0_{s,a}\right)=\delta
\]
and the exponential formula for the worst-case distribution gives
\[
 \left.\frac{d\lambda_t}{dt}\right|_{t=0}
 =\frac{\lambda^*\operatorname{Cov}_{\underline P^V_{s,a}}(V,W-V)}
              {\operatorname{Var}_{\underline P^V_{s,a}}(V)}.
\]
The derivative exists by the implicit function theorem, since the
derivative of the KL divergence with respect to $\lambda$ is
$-\operatorname{Var}_{\underline P^V_{s,a}}(V)/(\lambda^*)^3<0$.
Differentiating the value variance therefore yields
\begin{align*}
 &\left.\frac{d}{dt}
 \operatorname{Var}_{\underline P^{V+t(W-V)}_{s,a}}
                       \bigl(V+t(W-V)\bigr)\right|_{t=0}\\
 &=2\operatorname{Cov}_{\underline P^V_{s,a}}(V,W-V)\\
 &\quad-\frac1{\lambda^*}
 \operatorname{Cov}_{\underline P^V_{s,a}}
 \!\left(\bigl(V-\mathbb E_{\underline P^V_{s,a}}[V]\bigr)^2,W-V\right)\\
 &\quad+\frac{
 \mathbb E_{\underline P^V_{s,a}}
       [\bigl(V-\mathbb E_{\underline P^V_{s,a}}[V]\bigr)^3]}
 {\lambda^*\operatorname{Var}_{\underline P^V_{s,a}}(V)}
       \operatorname{Cov}_{\underline P^V_{s,a}}(V,W-V).
\end{align*}
By~\eqref{eq:large-delta-stationarity} and
$h_{s,a}^{\lambda^*,V}\leq p_\wedge^{-1}$,
\begin{align*}
 &\mathbb E_{\underline P^V_{s,a}}
       \!\left[\left|V-\mathbb E_{\underline P^V_{s,a}}[V]\right|\right]\\
 &=2\mathbb E_{\underline P^V_{s,a}}
       \!\left[\max\left\{\mathbb E_{\underline P^V_{s,a}}[V]-V,0\right\}\right]
 \leq2\lambda^*\bigl(\log(1/p_\wedge)-\delta\bigr).
\end{align*}
The expected positive and negative deviations from the mean are
equal. Bounding their squares by their respective endpoint
distances times the absolute deviations therefore gives
\[
 \operatorname{Var}_{\underline P^V_{s,a}}(V)
 \leq\frac{\operatorname{span}(V)}2
 \mathbb E_{\underline P^V_{s,a}}
 \!\left[\left|V-\mathbb E_{\underline P^V_{s,a}}[V]\right|\right].
\]
Cauchy--Schwarz then bounds the displayed mean absolute deviation
by $\operatorname{span}(V)/2$.
Combining these facts with the preceding multiplier bound and
$\operatorname{span}(V)\leq(1-\gamma)^{-1}$ gives
\begin{align*}
 \operatorname{Var}_{\underline P^V_{s,a}}(V)
 &\leq\frac{\lambda^*\bigl(\log(1/p_\wedge)-\delta\bigr)}{1-\gamma},\\
 \left|\mathbb E_{\underline P^V_{s,a}}
       [\bigl(V-\mathbb E_{\underline P^V_{s,a}}[V]\bigr)^3]\right|
 &\leq\frac{\operatorname{Var}_{\underline P^V_{s,a}}(V)}{1-\gamma}.
\end{align*}
These inequalities bound the absolute values of the three derivative
terms, respectively, by
\[
 \frac{1}{1-\gamma}\|W-V\|_\infty,\qquad
 \frac{2\bigl(\log(1/p_\wedge)-\delta\bigr)}{1-\gamma}\|W-V\|_\infty,\qquad
 \frac{2\bigl(\log(1/p_\wedge)-\delta\bigr)}{1-\gamma}\|W-V\|_\infty.
\]
For the middle bound, center $W-V$ in the covariance and use
$\left|W-V-\mathbb E_{\underline P^V_{s,a}}[W-V]\right|
\leq2\|W-V\|_\infty$.
A positive optimizer implies $\delta\leq\log(1/p_\wedge)$.
Since $\delta>1$, summing the three bounds gives a derivative
of absolute value at most
$4\log(1/p_\wedge)\|W-V\|_\infty/(1-\gamma)$.

This derivative bound extends to the entire segment
$V+t(W-V)$, $0\leq t\leq1$, including zero optimizers.
Indeed, the KL uncertainty set on the nominal support is compact.
Whenever value vectors converge, every limit of their minimizing
distributions minimizes the limiting objective. At a positive
optimizer that distribution is unique; at a zero optimizer every
minimizer is supported on states attaining the minimum value and
has zero value variance, by Lemma~\ref{lem:large-delta-variance}.
Consequently, the value variance is continuous along the segment.
It is differentiable with the preceding bound on every open
interval where it is positive, and is zero elsewhere. Integrating
on such intervals, and using continuity at their endpoints, proves
\eqref{eq:large-delta-value-variance-comparison}.

The same argument applies to the empirical row on the event of
Lemma~\ref{lem:large-delta-uniform}, with probability floor
$3p_\wedge/4$. Hence
\begin{align}
 &\left|\operatorname{Var}_{\underline{\widehat P}^V_{s,a}}(V)
             -\operatorname{Var}_{\underline{\widehat P}^W_{s,a}}(W)\right|
 \nonumber\\
 &\leq\frac{4\log(4/(3p_\wedge))}{1-\gamma}
                       \|V-W\|_\infty.
 \label{eq:large-delta-empirical-value-variance-comparison}
\end{align}
Taking square roots in these comparisons gives, in particular,
\begin{align*}
 \sqrt{\operatorname{Var}_{\underline P^W_{s,a}}(W)}
 &\leq\sqrt{\operatorname{Var}_{\underline P^V_{s,a}}(V)}
 +\sqrt{\frac{4\log(1/p_\wedge)}{1-\gamma}
                                  \|V-W\|_\infty},\\
 \sqrt{\operatorname{Var}_{\underline{\widehat P}^W_{s,a}}(W)}
 &\leq\sqrt{\operatorname{Var}_{\underline{\widehat P}^V_{s,a}}(V)}
 +\sqrt{\frac{4\log(4/(3p_\wedge))}{1-\gamma}
                                  \|V-W\|_\infty}.
\end{align*}

The nominal and empirical Bellman equations give the following
three componentwise comparisons:
\begin{align}
 \widehat V_{\pi^*}-V_{\pi^*}
 &\leq\gamma
 (I-\gamma\underline P^{V_{\pi^*},\pi^*})^{-1}
 \big(\underline{\widehat P}^{\widehat V_{\pi^*},\pi^*}
              -\underline P^{\widehat V_{\pi^*},\pi^*}\big)
                         \widehat V_{\pi^*},
 \label{eq:large-delta-fixed-policy-one}\\
 V_{\pi^*}-\widehat V_{\pi^*}
 &\leq\gamma
 (I-\gamma\underline{\widehat P}^{\widehat V_{\pi^*},\pi^*})^{-1}
 \big(\underline P^{V_{\pi^*},\pi^*}
              -\underline{\widehat P}^{V_{\pi^*},\pi^*}\big)
                         V_{\pi^*},
 \label{eq:large-delta-fixed-policy-two}\\
 \widehat V_{\widehat\pi^*}-V_{\widehat\pi}
 &\leq (I-\gamma\underline P^{V_{\widehat\pi},\widehat\pi})^{-1}
 \Big[\frac{2\gamma^T}{1-\gamma}\mathbf1
 +\gamma\big(\underline{\widehat P}^{\widehat V_{\widehat\pi^*},\widehat\pi}
              -\underline P^{\widehat V_{\widehat\pi^*},\widehat\pi}\big)
                         \widehat V_{\widehat\pi^*}\Big].
 \label{eq:large-delta-output-propagation}
\end{align}
For the first inequality, split the difference of the Bellman
equations at the nominal robust expectation of $\widehat V_{\pi^*}$.
The nominal robust-expectation difference is at most
$\underline P^{V_{\pi^*},\pi^*}
(\widehat V_{\pi^*}-V_{\pi^*})$, because this transition minimizes
the expectation of $V_{\pi^*}$. Rearranging and using the
nonnegative inverse proves the inequality. Interchanging the two
models gives the second inequality. For the third, contraction
gives $\|Q_T-Q^*_{\widehat\pi^*}\|_\infty\leq\gamma^T/(1-\gamma)$.
Greediness of $\widehat\pi$ therefore implies
\[
 0\leq\widehat V_{\widehat\pi^*}
 -r(\cdot,\widehat\pi(\cdot))
 -\gamma\underline{\widehat P}^{\widehat V_{\widehat\pi^*},\widehat\pi}
                       \widehat V_{\widehat\pi^*}
 \leq\frac{2\gamma^T}{1-\gamma}\mathbf1.
\]
Subtracting the nominal Bellman equation for $V_{\widehat\pi}$
and using its minimizing transition gives
\eqref{eq:large-delta-output-propagation}.

Apply Lemma~\ref{lem:large-delta-uniform} at
$\widehat V_{\pi^*}$, $V_{\pi^*}$, and
$\widehat V_{\widehat\pi^*}$, respectively. In the first and
third comparisons, use the nominal variance in
\eqref{eq:large-delta-uniform-local}; in the second, use the
empirical variance. The value-variance comparison proved above
then bounds each required standard deviation by the standard
deviation of the value function of the matrix in its inverse,
plus
\[
 \sqrt{\frac{4\log(4/(3p_\wedge))}{1-\gamma}
 \max\left\{
   \|V_{\pi^*}-\widehat V_{\pi^*}\|_\infty,
   \|\widehat V_{\widehat\pi^*}-V_{\widehat\pi}\|_\infty
 \right\}}.
\]
Here the empirical probability floor is at least $3p_\wedge/4$,
which explains the common logarithm. The three pairs of value
functions are, respectively,
$(\widehat V_{\pi^*},V_{\pi^*})$ under the nominal base,
$(V_{\pi^*},\widehat V_{\pi^*})$ under the empirical base, and
$(\widehat V_{\widehat\pi^*},V_{\widehat\pi})$ under the nominal
base. Thus the variance comparison is deterministic and requires
no concentration at $V_{\widehat\pi}$.

For each resulting variance, the corresponding value function
satisfies its own Bellman equation with rewards in $[0,1]$.
Lemma~\ref{lem:polished-resolvent} consequently bounds its
discounted propagation, including the factor $\gamma$, by
$1/[\sqrt2(1-\gamma)^{3/2}]$. Each inverse maps $\mathbf1$ to
$\mathbf1/(1-\gamma)$, and every span is at most
$(1-\gamma)^{-1}$.
Moreover, $V_{\widehat\pi}\leq V_{\pi^*}$ and
$\widehat V_{\pi^*}\leq\widehat V_{\widehat\pi^*}$, so the
negative part of $\widehat V_{\widehat\pi^*}-V_{\widehat\pi}$ is
bounded by the positive part of $V_{\pi^*}-\widehat V_{\pi^*}$.
The three comparisons therefore control both absolute errors and give
\begin{align*}
 &\max\left\{
   \|V_{\pi^*}-\widehat V_{\pi^*}\|_\infty,
   \|\widehat V_{\widehat\pi^*}-V_{\widehat\pi}\|_\infty
 \right\}\\
 &\leq8\sqrt2
 \sqrt{\frac1{Np_\wedge(1-\gamma)^3}
 \log\frac{36|\mathcal S|^2|\mathcal A|N}{\eta p_\wedge}}\\
 &\quad+16
 \sqrt{\frac{4\log(4/(3p_\wedge))}{Np_\wedge(1-\gamma)^3}
 \log\frac{36|\mathcal S|^2|\mathcal A|N}{\eta p_\wedge}}
 \sqrt{\max\left\{
   \|V_{\pi^*}-\widehat V_{\pi^*}\|_\infty,
   \|\widehat V_{\widehat\pi^*}-V_{\widehat\pi}\|_\infty
 \right\}}\\
 &\quad+21
 \sqrt{\frac{e^\delta}{N(1-\gamma)^4}
 \log\frac{36|\mathcal S|^2|\mathcal A|N}{\eta p_\wedge}}
 +\frac{652}{Np_\wedge(1-\gamma)^2}
 \log\frac{36|\mathcal S|^2|\mathcal A|N}{\eta p_\wedge}
 +\frac{2\gamma^T}{(1-\gamma)^2}.
\end{align*}
Applying Young's inequality to the term containing the square
root of the repeated maximum absorbs one third of that maximum
into the left-hand side. The additional term before division by
$2/3$ is
\[
 \frac{768\log(4/(3p_\wedge))}
 {Np_\wedge(1-\gamma)^3}
 \log\frac{36|\mathcal S|^2|\mathcal A|N}{\eta p_\wedge}.
\]
Finally,
\[
 0\leq V_{\pi^*}-V_{\widehat\pi}
 \leq (V_{\pi^*}-\widehat V_{\pi^*})
       +(\widehat V_{\widehat\pi^*}-V_{\widehat\pi}),
\]
so the policy loss is at most twice the maximum. Using
$24\sqrt2<34$ and
\[
 2304\log(4/(3p_\wedge))+1956
 \leq2304\bigl(1+\log(4/(3p_\wedge))\bigr)
\]
proves~\eqref{eq:large-delta-policy-bound}.

Finally, if $\delta\geq\log(4/(3p_\wedge))$, a point mass at any
state in the common support has KL divergence at most $\delta$
from either transition row. Hence, for every $V$ and $(s,a)$,
\[
 \underline P^V_{s,a}V
 =\underline{\widehat P}^V_{s,a}V
 =\min_{s':P^0(s'\mid s,a)>0}V(s').
\]
The nominal and empirical robust Bellman operators therefore agree,
so $V_{\pi^*}=\widehat V_{\widehat\pi^*}$. The empirical Bellman
residual of $\widehat\pi$ proved above is also its nominal residual.
Contraction for this policy consequently gives
\[
 \|V_{\pi^*}-V_{\widehat\pi}\|_\infty
 =\|\widehat V_{\widehat\pi^*}-V_{\widehat\pi}\|_\infty
 \leq\frac1{1-\gamma}\frac{2\gamma^T}{1-\gamma},
\]
which proves~\eqref{eq:large-delta-support-policy}.
\end{proof}

\begin{corollary}[Sample requirements for $\delta>1$]
\label{cor:large-delta-sample}
Under the assumptions of Theorem~\ref{thm:large-delta-policy}, let
$0<\epsilon\leq(1-\gamma)^{-1}$ and choose
\begin{equation}\label{eq:large-delta-planning-iterations}
 T\geq\left\lceil
 \frac{\log(24/((1-\gamma)^2\epsilon))}{-\log\gamma}
 \right\rceil.
\end{equation}
For every $\delta>1$, the sample condition
\begin{align}
 N\geq\max\Bigg\{&\frac{192}{p_\wedge},\,
 \frac{9216}{\epsilon^2}
 \left(\frac1{p_\wedge(1-\gamma)^3}
       +\frac{e^\delta}{(1-\gamma)^4}\right)
 \nonumber\\
 &+\frac{6144\bigl(1+\log(4/(3p_\wedge))\bigr)}
      {p_\wedge(1-\gamma)^3\epsilon}\Bigg\}
 \log\frac{36|\mathcal S|^2|\mathcal A|N}{\eta p_\wedge}
 \label{eq:large-delta-sample-requirement}
\end{align}
ensures $\|V_{\pi^*}-V_{\widehat\pi}\|_\infty\leq\epsilon$
with probability at least $1-\eta$. The sufficient total sample
count is
\begin{equation}\label{eq:large-delta-general-complexity}
 \widetilde{\mathcal O}\!\left(
 |\mathcal S||\mathcal A|\left[
 \frac1{p_\wedge}
 +\frac1{p_\wedge(1-\gamma)^3\epsilon^2}
 +\frac{e^\delta}{(1-\gamma)^4\epsilon^2}
 +\frac1{p_\wedge(1-\gamma)^3\epsilon}\right]\right).
\end{equation}
For $0<\epsilon\leq1$, this simplifies to
\begin{equation}\label{eq:large-delta-mixed-complexity}
 \widetilde{\mathcal O}\!\left(
 \frac{|\mathcal S||\mathcal A|}{\epsilon^2}
 \left[\frac1{p_\wedge(1-\gamma)^3}
       +\frac{e^\delta}{(1-\gamma)^4}\right]\right).
\end{equation}
For $\delta\geq\log(4/(3p_\wedge))$, only
\eqref{eq:large-delta-support-samples} is required, giving total
sample count $\widetilde{\mathcal O}(|\mathcal S||\mathcal A|/p_\wedge)$.
\end{corollary}
\begin{proof}
For~\eqref{eq:large-delta-policy-bound}, Cauchy--Schwarz and
$34^2+63^2=5125<72^2$ bound the sum of the two square-root terms by
\[
 72\sqrt{\frac1N
 \left(\frac1{p_\wedge(1-\gamma)^3}
       +\frac{e^\delta}{(1-\gamma)^4}\right)
 \log\frac{36|\mathcal S|^2|\mathcal A|N}{\eta p_\wedge}}.
\]
Young's inequality bounds this display by
\[
 \frac{3\epsilon}{8}
 +\frac{3456}{\epsilon N}
 \left(\frac1{p_\wedge(1-\gamma)^3}
       +\frac{e^\delta}{(1-\gamma)^4}\right)
 \log\frac{36|\mathcal S|^2|\mathcal A|N}{\eta p_\wedge}.
\]
Adding the $N^{-1}$ term in~\eqref{eq:large-delta-policy-bound}
and applying~\eqref{eq:large-delta-sample-requirement} bounds
the sum by $3\epsilon/4$: the identities
$(3/8)\cdot9216=3456$ and $(3/8)\cdot6144=2304$
account for its two sample coefficients.
The choice of $T$ bounds the planning term
$6\gamma^T/(1-\gamma)^2$ by $\epsilon/4$.

Multiplication by $|\mathcal S||\mathcal A|$ gives
the total sample requirements; the displayed logarithms contribute
only logarithmic factors. If $\epsilon\leq1$, the
$p_\wedge^{-1}$ and $[p_\wedge(1-\gamma)^3\epsilon]^{-1}$
terms are dominated by $[p_\wedge(1-\gamma)^3\epsilon^2]^{-1}$,
which proves~\eqref{eq:large-delta-mixed-complexity}.
The last assertion follows directly
from~\eqref{eq:large-delta-support-policy}.
\end{proof}

% \subsection{Lower Bound Design of Theorem \ref{thm:lb1}.}

\section{Lower Bound Proof}
In this section, we present our proof of the lower bounds.
% \subsection{Lower Bound Design of Theorem \ref{thm:lb1}.}

\subsection{Hard MDP design and lower bound proof for Theorem \ref{thm:lb1}}

\begin{theorem}[Restatement of Theorem \ref{thm:lb1}]
Suppose $|\mathcal S|\geq2$, $|\mathcal A|\geq2$,
$\gamma\in(3/4,1)$, $\delta\geq1-\gamma$, $0<\epsilon<1$, and
\[
 0<\epsilon\leq
 \frac1
 {754(1-\gamma)\left(\delta+\log\dfrac2{1-\gamma}\right)}.
\]
There exist two DR-MDPs $\mathcal M_0,\mathcal M_1$ with the same
known rewards in $[0,1]$ and KL-defined $(s,a)$-rectangular
uncertainty sets of radius $\delta$ such that, given $N$ independent
samples per state-action pair, independently across pairs,
\begin{align}
 \inf_{\widehat\pi}\max_{\phi\in\{0,1\}}
 \mathbb P_\phi\!\left(
 \|V_\phi^*-V_\phi^{\widehat\pi}\|_\infty>\epsilon\right)
 >\frac18,
\end{align}
provided that
\[
 N|\mathcal S||\mathcal A|
 \leq\frac1{8457}\,
 \frac{\delta e^\delta|\mathcal S||\mathcal A|}
 {(1+\delta)(1-\gamma)^4\epsilon^2
  \left(\delta+\log\dfrac2{1-\gamma}\right)^2}.
\]
The infimum is over all learning algorithms and their returned
policies, including randomization. The failure probability is
at least $3/8$. The minimal non-zero nominal transition probability
is determined by the construction and satisfies
$\delta<\log(1/p_\wedge)$. Consequently, the total sample
complexity is lower bounded by
\[
 \widetilde{\Omega}\!\left(
 \frac{|\mathcal S||\mathcal A|\delta e^\delta}
 {(1-\gamma)^4\epsilon^2}\right),
\]
where the suppressed factors are logarithmic in $1/p_\wedge$
and $1/(1-\gamma)$ and are displayed explicitly in the sample
condition above.
\end{theorem}

We construct a family of hard instances to establish the minimax lower bound for KL-defined robust RL.
Consider a class of two-state MDPs $\{\mathcal{M}_\phi\}$ parameterized by $p$ with an uncertainty level $\delta$.

The state space includes a state $s_0$ and an absorbing state $s_1$, with two available actions
$a \in \{0,1\}$:
\begin{align}
s_0 &: r(s_0)=0, \qquad
s_1 : r(s_1)=1. \nonumber
\end{align}

The nominal transition probabilities from $s_0$ are defined as follows:
% probability $p$ and transitions to $s_1$ otherwise; for action $a=1-\phi$, the corresponding probabilities are
% $q$ and $1-q$:
\begin{align}
P(s_0 \mid s_0, a=\phi) &= p, &
P(s_1 \mid s_0, a=\phi) &= 1-p, \nonumber \\
P(s_0 \mid s_0, a=1-\phi) &= q, &
P(s_1 \mid s_0, a=1-\phi) &= 1-q. \nonumber
\end{align}

State $s_1$ is absorbing for both actions:
\begin{align}
P(s_1 \mid s_1, a) = 1, \qquad \forall a \in \{0,1\}. \nonumber
\end{align}

% Under the nominal (vanilla) MDP, the transition probabilities are given by
% \begin{align}
% P(s_1 \mid s_0, a=\phi) &= 1-p, &
% P(s_0 \mid s_0, a=\phi) &= p, \nonumber \\
% P(s_1 \mid s_0, a=1-\phi) &= 1-q, &
% P(s_0 \mid s_0, a=1-\phi) &= q. \nonumber
% \end{align}

In the robust formulation, the adversary selects a worst-case transition kernel within the KL uncertainty set.
The resulting worst-case transitions preserve the same structure:
\begin{align}
P^{\mathrm{wc}}(s_1 \mid s_0, a=\phi) &= 1-\tilde{p}, &
P^{\mathrm{wc}}(s_0 \mid s_0, a=\phi) &= \tilde{p}, \nonumber \\
P^{\mathrm{wc}}(s_1 \mid s_0, a=1-\phi) &= 1-\tilde{q}, &
P^{\mathrm{wc}}(s_0 \mid s_0, a=1-\phi) &=\tilde{q}. \nonumber
\end{align}

Since $s_1$ is absorbing and yields unit reward at every step, its value function is
\begin{align}
V(s_1) = \frac{1}{1-\gamma}. \nonumber
\end{align}

For a policy that always selects action $\phi$, the robust Bellman equation at $s_0$ is
\begin{align}
V_\phi(s_0)
&= \gamma \tilde{p} V_\phi(s_0) + \gamma (1-\tilde{p}) V(s_1) \nonumber \\
&= \gamma \tilde{p} V_\phi(s_0) + \frac{\gamma(1-\tilde{p})}{1-\gamma}, \nonumber
\end{align}
which yields the closed-form expression
\begin{align}
V_\phi(s_0)
= \frac{\gamma(1-\tilde{p})}{(1-\gamma)(1-\gamma \tilde{p})}. \nonumber
\end{align}

Similarly, for the policy that always selects action $1-\phi$, we obtain
\begin{align}
V_{1-\phi}(s_0)
&= \gamma \tilde{q} V_{1-\phi}(s_0) + \gamma (1-\tilde{q}) V(s_1) \nonumber \\
&= \gamma \tilde{q} V_{1-\phi}(s_0) + \frac{\gamma(1-\tilde{q})}{1-\gamma}, \nonumber
\end{align}
and hence
\begin{align}
V_{1-\phi}(s_0)
= \frac{\gamma(1-\tilde{q})}{(1-\gamma)(1-\gamma \tilde{q})}. \nonumber
\end{align}

We now compare the values induced by the two deterministic policies $\pi(s_0)=\phi$ and $\pi(s_0)=1-\phi$. 
Define the value gap as
\begin{align}
\Delta V = \big| V_\phi(s_0) - V_{1-\phi}(s_0) \big|. \nonumber
\end{align}

Substituting the closed-form expressions for the value functions, we obtain
\begin{align}
\Delta V
&= \left|
\frac{\gamma(1-\tilde{q})}{(1-\gamma)(1-\gamma\tilde{q})}
-
\frac{\gamma(1-\tilde{p})}{(1-\gamma)(1-\gamma\tilde{p})}
\right| \nonumber \\
&= \left|
\frac{\gamma\big((1-\tilde{q})(1-\gamma \tilde{p})-(1-\tilde{p})(1-\gamma \tilde{q})\big)}
{(1-\gamma)(1-\gamma \tilde{q})(1-\gamma \tilde{p})}
\right| \nonumber \\
&= \left|
\frac{\gamma(1-\gamma)(\tilde{p}-\tilde{q})}
{(1-\gamma)(1-\gamma\tilde{q})(1-\gamma\tilde{p})}
\right| \nonumber \\
&= \frac{\gamma|\tilde{p}-\tilde{q}|}{(1-\gamma \tilde{q})(1-\gamma \tilde{p})}.
\label{eq:value_gap}
\end{align}

We set
\begin{align}
 p&=\gamma e^{-\delta/\gamma},\qquad q=pe^{\varepsilon},\nonumber\\
 \varepsilon
 &=\frac{8(1+\gamma)^2(1-\gamma)^2\epsilon}{\gamma^2}
   \left(\delta+\log\frac2{1-\gamma}\right).
 \label{eq:lb1-perturbation}
\end{align}
Since $\delta\geq1-\gamma$ and
$e^{(1-\gamma)/\gamma}\geq1+(1-\gamma)/\gamma=1/\gamma$,
we have $p\leq\gamma^2$. To verify the perturbation range,
first note that $\gamma>3/4$ gives
\[
 128\frac{(1+\gamma)^2}{\gamma^2}
 \leq\frac{6272}{9}<754.
\]
Convexity of the exponential on $[-1/3,0]$ also gives
\begin{align*}
 p&\leq\gamma e^{-(1-\gamma)/\gamma}\\
  &\leq\gamma-3(1-e^{-1/3})(1-\gamma),
\end{align*}
since $0<(1-\gamma)/\gamma<1/3$. Consequently,
\[
 \frac{1-\gamma}{1-p}
 \leq\frac1{4-3e^{-1/3}}\leq\frac{113}{209},
\]
where the last inequality follows from
$e^{1/3}\geq1+1/3+1/18+1/162=113/81$. Hence
\[
 256\frac{(1+\gamma)^2(1-\gamma)}{\gamma^2(1-p)}
 \leq\frac{1417472}{1881}<754.
\]
Substituting the theorem's accuracy condition
into~\eqref{eq:lb1-perturbation} now yields
\begin{equation}\label{eq:lb1-perturbation-range}
 0<\varepsilon\leq\frac{1-\gamma}{16},
 \qquad \varepsilon\leq\frac{1-p}{32}.
\end{equation}
In particular,
\begin{align}
 0<q-p=p(e^\varepsilon-1)
 &\leq2p\varepsilon\leq\frac{1-p}{2},\qquad
 1-q\geq\frac{1-p}{2},
 \label{eq:pq_gap}\\
 q=\gamma e^{-\delta/\gamma+\varepsilon}&<\gamma,
 \nonumber
\end{align}
since $\delta\geq1-\gamma$. Moreover,
\begin{align}
 \log\frac1p-\delta
 &=-\log\gamma+\frac{\delta(1-\gamma)}{\gamma}
 \geq(1-\gamma)\left(1+\frac\delta\gamma\right),
 \label{eq:lb1-endpoint-slack}\\
 \log\frac1x-\delta
 &\geq\log\frac1p-\delta-\varepsilon
 \geq\frac{15}{16}\left(\log\frac1p-\delta\right)>0,
 \qquad x\in[p,q].
 \label{eq:lb1-interior}
\end{align}
For each $x\in[p,q]$, the worst-case self-loop probability $y$
is therefore the unique solution in $(x,1)$ of
\[
 \text{KL}(y\|x)
 =y\log\frac yx+(1-y)\log\frac{1-y}{1-x}=\delta.
\]
Indeed, this divergence is zero at $y=x$, equals
$\log(1/x)>\delta$ at $y=1$, and is strictly increasing for $y>x$.
At $x=p$ and $x=q$, the solutions are $\widetilde p$ and
$\widetilde q$, respectively, so
\begin{equation}\label{eq:kl_equal}
 \text{KL}(\widetilde p\|p)
 =\text{KL}(\widetilde q\|q)=\delta.
\end{equation}

We next bound the difference between the worst-case probabilities.
The partial derivatives of the binary KL divergence are
\[
 \frac{\partial}{\partial x}\text{KL}(y\|x)
 =\frac{x-y}{x(1-x)},\qquad
 \frac{\partial}{\partial y}\text{KL}(y\|x)
 =\log\frac{y(1-x)}{x(1-y)}>0.
\]
The implicit function theorem gives
\begin{equation}\label{eq:implicit_derivative}
 x\frac{dy}{dx}
 =\frac{y-x}
 {(1-x)\log\!\left(\dfrac{y(1-x)}{x(1-y)}\right)}>0.
\end{equation}
By convexity of $\text{KL}(\cdot\|x)$ between $x$ and $1$,
\[
 \delta\leq\frac{y-x}{1-x}\log\frac1x.
\]
Since $-\log\gamma\leq(1-\gamma)/\gamma$ and
$\delta\geq1-\gamma$, it follows that
\begin{equation}\label{eq:derivative_lower}
 \frac{y-x}{1-x}
 \geq\frac{\delta}{\log(1/p)}
 \geq\frac\gamma2,
 \qquad
 \log\frac1p\leq\frac{\delta+1-\gamma}{\gamma}
 \leq\frac{2\delta}{\gamma}.
\end{equation}
To bound the logarithm in~\eqref{eq:implicit_derivative}, rearrange
$\text{KL}(y\|x)=\delta$ to obtain
\begin{align*}
 \log\frac1x-\delta
 &=-y\log y-(1-y)\log(1-y)
   +(1-y)\log\frac{1-x}{x}\\
 &\leq\left(\log\frac1p+2\right)\sqrt{1-y}.
\end{align*}
Here $-y\log y\leq1-y$ and
$-(1-y)\log(1-y)\leq2(\sqrt{1-y}-(1-y))$, both following
from $\log t\leq t-1$, imply the stated bound on the entropy.
Together with~\eqref{eq:lb1-interior}, this yields
\[
 \sqrt{1-y}\geq
 \frac{15[\log(1/p)-\delta]}{16(\log(1/p)+2)}
 \geq\frac{15(1-\gamma)}{32}.
\]
For the last inequality, \eqref{eq:lb1-endpoint-slack} and
$\delta+\gamma\geq1$ give
\[
 \frac{\log(1/p)+2}{\log(1/p)-\delta}
 \leq\frac{\delta+1+\gamma}{(1-\gamma)(\delta+\gamma)}
 \leq\frac2{1-\gamma}.
\]
Consequently, throughout $x\in[p,q]$,
\begin{align}
 \log\frac{y(1-x)}{x(1-y)}
 &\leq\log\frac1p+\log(1-p)
       +2\log\frac{32}{15(1-\gamma)}\nonumber\\
 &\leq2\left(\delta+\log\frac2{1-\gamma}\right).
 \label{eq:lb1-log-odds}
\end{align}
For the last inequality, use
$\log(1/p)\leq(\delta+1-\gamma)/\gamma$ and
$1-p\leq\min\{1,1-\gamma+\delta\}$. Moreover,
\[
 \left(2-\frac1\gamma\right)\delta-\frac{1-\gamma}{\gamma}
 -\log\min\{1,1-\gamma+\delta\}
 \geq2\gamma-\frac1\gamma>\frac16>\log\frac{256}{225}.
\]
Indeed, the expression on the left decreases up to $\delta=\gamma$
and increases thereafter; its value there is $2\gamma-1/\gamma$.
Combining~\eqref{eq:implicit_derivative},
\eqref{eq:derivative_lower}, and~\eqref{eq:lb1-log-odds}, and
integrating over $[p,q]$, gives
\begin{align}
 \widetilde q-\widetilde p
 &=\int_p^q\frac{dy}{dx}\,dx\nonumber\\
 &\geq\frac{\gamma\log(q/p)}
 {4\left(\delta+\log\dfrac2{1-\gamma}\right)}
 =\frac{2(1-\gamma^2)^2\epsilon}{\gamma}.
 \label{eq:411}
\end{align}

The reference point $\gamma$ satisfies
\begin{equation}\label{eq:kl_gamma}
 \text{KL}(\gamma\|p)
 =\delta+(1-\gamma)\log\frac{1-\gamma}{1-p}<\delta.
\end{equation}
Hence $\widetilde q>\widetilde p>\gamma$. In particular,
\begin{equation}\label{eq:delta_v_lower}
 \Delta V
 >\frac{\gamma(\widetilde q-\widetilde p)}{(1-\gamma^2)^2}
 \geq2\epsilon.
\end{equation}
For a stationary randomized policy $\widehat\pi$, its worst-case
self-loop probability at $s_0$ is
$\widetilde p+(1-\widehat\pi(\phi\mid s_0))
 (\widetilde q-\widetilde p)$. Substituting this probability into
its robust Bellman equation gives
\begin{align}
 \|V_\phi^*-V_\phi^{\widehat\pi}\|_\infty
 &=\frac{\gamma(\widetilde q-\widetilde p)
                (1-\widehat\pi(\phi\mid s_0))}
 {(1-\gamma\widetilde p)
  \left[1-\gamma\big(\widetilde p+
    (1-\widehat\pi(\phi\mid s_0))(\widetilde q-\widetilde p)\big)\right]}
 \nonumber\\
 &\geq\frac{\gamma(\widetilde q-\widetilde p)}
 {(1-\gamma\widetilde p)^2}
 (1-\widehat\pi(\phi\mid s_0)).
 \label{eq:lb1-policy-gap}
\end{align}
The equality holds because the value at $s_1$ is independent of
the policy. Moreover, $\widetilde p>\gamma$ and~\eqref{eq:411} give
\[
 \frac{\gamma(\widetilde q-\widetilde p)}
 {(1-\gamma\widetilde p)^2}
 >\frac{\gamma(\widetilde q-\widetilde p)}{(1-\gamma^2)^2}
 \geq2\epsilon.
\]

To distinguish the two models, we bound the KL divergence of their
sample distributions. The two models interchange $p$ and $q$
between actions $0$ and $1$. Independence of the observations and
the exact sum of the two Bernoulli KL divergences give
\begin{align}
 \text{KL}(\mathbb P_0\|\mathbb P_1)
 &=\E_{\mathbb P_0}\!\left[
   \log\frac{\mathbb P_0}{\mathbb P_1}\right]
 =N\big(\text{KL}(p\|q)+\text{KL}(q\|p)\big)\nonumber\\
 &=N(q-p)\left(\varepsilon+\log\frac{1-p}{1-q}\right)\nonumber\\
 &\leq\frac{N(q-p)\varepsilon}{1-q}
 \leq\frac{Np\varepsilon^2}{1-p-\varepsilon}
 \leq\frac{32Np\varepsilon^2}{31(1-p)}.
 \label{eq:lb1-data-kl}
\end{align}
The ratio in the expectation is evaluated at the observed dataset;
independent algorithmic randomization leaves this divergence
unchanged. The first inequality uses
$\log((1-p)/(1-q))\leq(q-p)/(1-q)$ and
$q-p=q(1-e^{-\varepsilon})\leq q\varepsilon$.
For the second, $e^\varepsilon\leq(1-\varepsilon)^{-1}$ gives
\[
 \frac{q-p}{1-q}
 =\frac{p(e^\varepsilon-1)}{1-pe^\varepsilon}
 \leq\frac{p\varepsilon}{1-p-\varepsilon}.
\]
The last inequality follows from
$\varepsilon\leq(1-p)/32$ in~\eqref{eq:lb1-perturbation-range}.
If
\begin{equation}
 N\leq\frac{31}{2^{14}}\,
 \frac{\gamma^4(1-p)}
 {p(1+\gamma)^4(1-\gamma)^4\epsilon^2
  \left(\delta+\log\dfrac2{1-\gamma}\right)^2},
 \label{eq:lb1-testing-samples}
\end{equation}
then substituting~\eqref{eq:lb1-perturbation}
into~\eqref{eq:lb1-data-kl} gives
\[
 \text{KL}(\mathbb P_0\|\mathbb P_1)
 \leq\frac{2048}{31}\frac{31}{2^{14}}=\frac18.
\]
For fixed $\delta>0$, the following derivative is nonpositive
on $\gamma\in[3/4,1]$:
\begin{align*}
 &\frac{\partial}{\partial\gamma}
 \left[\frac{\gamma^4}{(1+\gamma)^4}
       \left(\frac{e^{\delta/\gamma}}\gamma-1\right)\right]\\
 &\quad=\frac{\gamma^2}{(1+\gamma)^5}
 \left[e^{\delta/\gamma}
       \left(3-\gamma-\frac{(1+\gamma)\delta}{\gamma}\right)
       -4\gamma\right]\leq0.
\end{align*}
Indeed, maximizing the exponential term in brackets over
$\delta\geq0$ gives
$(1+\gamma)e^{2(1-\gamma)/(1+\gamma)}
\leq2e^{2/7}<3\leq4\gamma$.
Evaluating at $\gamma=1$ and using $e^\delta\geq1+\delta$
therefore yields
\begin{align}
 \frac{\gamma^4}{(1+\gamma)^4}\frac{1-p}{p}
 &=\frac{\gamma^4}{(1+\gamma)^4}
   \left(\frac{e^{\delta/\gamma}}\gamma-1\right)\nonumber\\
 &\geq\frac{e^\delta-1}{16}
 \geq\frac{\delta e^\delta}{16(1+\delta)}.
 \label{eq:lb1-exponential-factor}
\end{align}
Since $1/8457<31/2^{18}$, this inequality shows that the
sample condition in Theorem~\ref{thm:lb1}
implies~\eqref{eq:lb1-testing-samples}.
By Pinsker's inequality, the total variation distance between
$\mathbb P_0$ and $\mathbb P_1$ is at most $1/4$.
Choose model $0$ when $\widehat\pi(0\mid s_0)\geq1/2$ and
model $1$ otherwise. An incorrect decision implies a robust value
loss strictly greater than $\epsilon$
by~\eqref{eq:lb1-policy-gap}. Moreover,
\begin{align*}
 &\mathbb P_0\big(\widehat\pi(0\mid s_0)<1/2\big)
 +\mathbb P_1\big(\widehat\pi(0\mid s_0)\geq1/2\big)\\
 &=1-\mathbb P_0\big(\widehat\pi(0\mid s_0)\geq1/2\big)
     +\mathbb P_1\big(\widehat\pi(0\mid s_0)\geq1/2\big)
 \geq\frac34.
\end{align*}
Consequently, the maximum failure probability is at least $3/8$.

The same conclusion holds for policies depending on the past.
Use the fixed worst-case rows with self-loop probabilities
$\widetilde p$ and $\widetilde q$, and evaluate the policy on
histories in which every transition returns to $s_0$.
Each actual action history that remains at $s_0$ for $t$ steps
has probability at least $\widetilde p^{t}$ times its probability
under these fixed-state histories. At $s_0$, each action other
than $\phi$ has Bellman value loss
$\gamma(\widetilde q-\widetilde p)/(1-\gamma\widetilde p)$
relative to $V_\phi^*$; action $\phi$ has zero loss.
Summing the discounted losses, and using that the fixed rows
are feasible in the uncertainty sets, yields
\begin{align*}
 V_\phi^*(s_0)-V_\phi^{\widehat\pi}(s_0)
 &\geq\frac{\gamma(\widetilde q-\widetilde p)}
 {(1-\gamma\widetilde p)^2}
 \left(1-(1-\gamma\widetilde p)\E_{\widehat\pi}\!\left[
 \sum_{t=0}^{\infty}(\gamma\widetilde p)^t
 \mathbf1\{a_t=\phi\}\right]\right).
\end{align*}
Here the expectation is over the policy's randomization along
these fixed-state histories. Choose model $0$ if
\[
 (1-\gamma\widetilde p)\E_{\widehat\pi}\!\left[
 \sum_{t=0}^{\infty}(\gamma\widetilde p)^t
 \mathbf1\{a_t=0\}\right]\geq\frac12,
\]
and model $1$ otherwise. The normalized frequencies of actions
$0$ and $1$ sum to one, so an incorrect decision again implies
loss strictly greater than $\epsilon$.
The same total variation bound gives
maximum failure probability at least $3/8$.

For $|\mathcal S|>2$, add absorbing states with reward zero,
unreachable from $s_0$. For $|\mathcal A|>2$, give every additional
action at $s_0$ the inferior nominal row $q$ in both models.
All added rows have identical data distributions in the two models.
The value-loss bounds continue to hold, and the frequencies of
actions $0$ and $1$ now sum to at most one, which suffices for
the same decision rule. The minimal non-zero nominal probability
is $p_\wedge=\min\{p,1-q\}$, so
$\delta<\log(1/p)\leq\log(1/p_\wedge)$.
For $\delta\geq1$, $q\leq e^{-1+1/64}<1/2$ and therefore
$p_\wedge=p=\gamma e^{-\delta/\gamma}$.
There are $N|\mathcal S||\mathcal A|$ samples in total.
Combining~\eqref{eq:lb1-testing-samples}
and~\eqref{eq:lb1-exponential-factor} with the theorem's sample
condition and taking the infimum over learning algorithms
proves~\eqref{eq:inf}. Finally,
$\delta<\log(1/p_\wedge)$ gives
\begin{align*}
 &(1+\delta)\left(\delta+\log\frac2{1-\gamma}\right)^2\\
 &\qquad\leq
 \left(1+\log\frac1{p_\wedge}\right)
 \left(\log\frac1{p_\wedge}+\log\frac2{1-\gamma}\right)^2.
\end{align*}
Thus the factors suppressed in the stated
$\widetilde{\Omega}$ bound are logarithmic in $1/p_\wedge$
and $1/(1-\gamma)$. This completes the proof of
Theorem~\ref{thm:lb1}.

\subsection{Proof of Theorem \ref{thm:lb2}}
We construct a family of hard instances to establish the minimax lower bound for KL-defined robust RL.
Consider a class of three-state MDPs $\{\mathcal{M}_\phi:\phi\in\{0,1\}\}$ parameterized by $p,q$ with an uncertainty level $\delta$.

The state space includes a state $s_0$ and two absorbing states $s_1,s_2$, with two available actions
$a \in \{0,1\}$:
\begin{align}
s_0 &: r(s_0)=0, \qquad
s_1: r(s_1)=1, \qquad
s_2: r(s_2)=0. \nonumber
\end{align}

The nominal transition probabilities from $s_0$ are defined as follows:
% probability $p$ and transitions to $s_1$ otherwise; for action $a=1-\phi$, the corresponding probabilities are
% $q$ and $1-q$:
\begin{align}
P(s_1 \mid s_0, a=\phi) &= p, &
P(s_2 \mid s_0, a=\phi) &= 1-p, \nonumber \\
P(s_1 \mid s_0, a=1-\phi) &= q, &
P(s_2 \mid s_0, a=1-\phi) &= 1-q. \nonumber
\end{align}

States $s_1$ and $s_2$ are absorbing for both actions:
\begin{align}
P(s_1 \mid s_1, a) = 1, \qquad \forall a \in \{0,1\};\nonumber
\\P(s_2 \mid s_2, a) = 1, \qquad \forall a \in \{0,1\}.\nonumber
\end{align}

\subsubsection*{Case 1: $0<\delta<\log(1/p_\wedge)$}
We set
\begin{align*}
0<\epsilon\le
\frac{\gamma\delta}
{8(1-\gamma)\log(1/p_\wedge)
 \log\!\left(
 \dfrac{4}{p_\wedge^3[\log(1/p_\wedge)-\delta]^2}
 \right)}
\min\!\left\{
1,\frac{\log(1/p_\wedge)-\delta}{2}
\right\}.
\end{align*}
Then, we have
\begin{equation}\label{eq:lb2-nominal-perturbation}
 0<\varepsilon\leq p_\wedge
 \min\left\{1,\frac{\log(1/p_\wedge)-\delta}{2}\right\}.
\end{equation}
Consequently,
\[
 p_\wedge=1-p<1-q\leq2p_\wedge<\frac14,
 \qquad \frac34<1-2p_\wedge\leq q<p<1.
\]
The minimal non-zero nominal transition probability is exactly
$p_\wedge$ in each model.

The worst-case transition from $s_0$ has support contained in
$\{s_1,s_2\}$. For $x\in[p_\wedge,p_\wedge+\varepsilon]$, let $y$
be the worst-case probability of reaching $s_2$ when its nominal
probability is $x$. The corresponding KL divergence is
\[
 \text{KL}(y\|x)
 =y\log\frac yx+(1-y)\log\frac{1-y}{1-x}.
\]
For every $x$ in this interval, \eqref{eq:lb2-nominal-perturbation}
and $\log(1+t)\leq t$ for $t\geq0$ imply
\begin{align}
 \log\frac1x-\delta
 &=\log\frac1{p_\wedge}-\delta-\log\frac{x}{p_\wedge}\nonumber\\
 &\geq\log\frac1{p_\wedge}-\delta
       -\frac{\varepsilon}{p_\wedge}
 \geq\frac{\log(1/p_\wedge)-\delta}{2}>0.
 \label{eq:lb2-distance-to-endpoint}
\end{align}
The divergence is zero at $y=x$, equals $\log(1/x)>\delta$
at $y=1$, and is strictly increasing for $y>x$, since
\[
 \frac{\partial}{\partial y}\text{KL}(y\|x)
 =\log\frac{y(1-x)}{x(1-y)}>0.
\]
Hence there is a unique $y\in(x,1)$ satisfying
\begin{equation}\label{eq:lb2-binary-kl-equation}
 \text{KL}(y\|x)=\delta.
\end{equation}
Since the divergence is strictly increasing for $y>x$, the
worst-case probabilities of reaching $s_2$ and $s_1$ are $y$
and $1-y$, respectively.

By convexity of $y\mapsto\text{KL}(y\|x)$ between
$y=x$ and $y=1$, we obtain
\begin{equation}\label{eq:lb2-y-minus-x}
 \delta=\text{KL}(y\|x)
 \leq\frac{y-x}{1-x}\log\frac1x,
 \qquad
 \frac{y-x}{1-x}\geq\frac{\delta}{\log(1/x)}.
\end{equation}
Rearranging~\eqref{eq:lb2-binary-kl-equation}, we obtain
\begin{align}
 \log\frac1x-\delta
 &=-y\log y-(1-y)\log(1-y)
      +(1-y)\log\frac{1-x}{x}\nonumber\\
 &\leq2\sqrt{1-y}+(1-y)\log\frac1{p_\wedge}\nonumber\\
 &\leq\big(\log(1/p_\wedge)+2\big)\sqrt{1-y}.
 \label{eq:lb2-endpoint-identity}
\end{align}
For the first inequality, $\log t\leq t-1$ gives
\[
 -y\log y\leq1-y,\qquad
 -(1-y)\log(1-y)
 \leq2\big(\sqrt{1-y}-(1-y)\big),
\]
and $x\geq p_\wedge$ gives
$\log((1-x)/x)\leq\log(1/p_\wedge)$.
The last inequality in~\eqref{eq:lb2-endpoint-identity} uses
$1-y\leq\sqrt{1-y}$.
Combining~\eqref{eq:lb2-distance-to-endpoint}
and~\eqref{eq:lb2-endpoint-identity}, and using
$\log(1/p_\wedge)+2\leq1/p_\wedge$ for $0<p_\wedge<1/8$, gives
\begin{align}
 1-y
 &\geq\frac{[\log(1/p_\wedge)-\delta]^2}
             {4[\log(1/p_\wedge)+2]^2}
 \geq\frac{p_\wedge^2[\log(1/p_\wedge)-\delta]^2}{4}.
 \label{eq:lb2-positive-good-probability}
\end{align}
It follows that
\begin{equation}\label{eq:lb2-log-odds-bound}
 0<\log\frac{y(1-x)}{x(1-y)}
 \leq\log\frac1{p_\wedge(1-y)}
 \leq\log\!\left(\frac{4}
 {p_\wedge^3[\log(1/p_\wedge)-\delta]^2}\right).
\end{equation}

The partial derivative of the KL divergence with respect to $x$ is
\[
 \frac{\partial}{\partial x}\text{KL}(y\|x)
 =\frac{x-y}{x(1-x)}.
\]
Since $0<x<y<1$ and the partial derivative with respect to $y$
is positive, the implicit function theorem applies throughout
$[p_\wedge,p_\wedge+\varepsilon]$. Differentiating
\eqref{eq:lb2-binary-kl-equation} and using
\eqref{eq:lb2-y-minus-x}, we obtain
\begin{align*}
 \frac{dy}{dx}
 &=\frac{y-x}
 {x(1-x)\log\!\left(\dfrac{y(1-x)}{x(1-y)}\right)}\\
 &\geq\frac{\delta}
 {x\log(1/x)\log\!\left(\dfrac{y(1-x)}{x(1-y)}\right)}.
\end{align*}
Now $x\leq2p_\wedge$, $\log(1/x)\leq\log(1/p_\wedge)$,
and~\eqref{eq:lb2-log-odds-bound} show that, throughout the interval,
\begin{equation}\label{eq:lb2-uniform-derivative}
 \frac{dy}{dx}\geq
 \frac{\delta}
 {2p_\wedge\log(1/p_\wedge)
  \log\!\left(\dfrac{4}
  {p_\wedge^3[\log(1/p_\wedge)-\delta]^2}\right)}.
\end{equation}
Let $\widetilde p$ and $\widetilde q$ be the worst-case probabilities
of reaching $s_1$ under the actions with nominal probabilities $p$
and $q$, respectively. They equal $1-y$ at $x=p_\wedge$ and
$x=p_\wedge+\varepsilon$, respectively.
Integrating~\eqref{eq:lb2-uniform-derivative} gives
\begin{align}
 \widetilde p-\widetilde q
 &=\int_{p_\wedge}^{p_\wedge+\varepsilon}\frac{dy}{dx}\,dx\nonumber\\
 &\geq\frac{\delta\varepsilon}
 {2p_\wedge\log(1/p_\wedge)
  \log\!\left(\dfrac{4}
  {p_\wedge^3[\log(1/p_\wedge)-\delta]^2}\right)}
 =\frac{4(1-\gamma)\epsilon}{\gamma}.
 \label{eq:lb2-robust-transition-gap}
\end{align}

Since finite KL divergence preserves the supports of deterministic
transitions, $s_1$ and $s_2$ remain absorbing, with robust values
$1/(1-\gamma)$ and zero, respectively. Thus action $\phi$ is optimal
in model $\mathcal M_\phi$, and
\[
 V_\phi^*(s_0)=\frac{\gamma\widetilde p}{1-\gamma}.
\]
For any returned randomized policy $\widehat\pi$, the
$(s,a)$-rectangular uncertainty set allows the transition for each
action to be minimized separately. The robust Bellman equation gives
\begin{align*}
 V_\phi^{\widehat\pi}(s_0)
 &=\gamma\E_{a\sim\widehat\pi(\cdot\mid s_0)}
 \!\left[\inf_{P_{s_0,a}\in\mathcal P_{s_0,a}(\delta)}
 \E_{s'\sim P_{s_0,a}}
 \!\left[V_\phi^{\widehat\pi}(s')\right]\right]\\
 &=\frac{\gamma}{1-\gamma}
 \left[\widehat\pi(\phi\mid s_0)\widetilde p
       +\big(1-\widehat\pi(\phi\mid s_0)\big)\widetilde q\right].
\end{align*}
Combining this with~\eqref{eq:lb2-robust-transition-gap}, we obtain
\begin{align}
 \|V_\phi^*-V_\phi^{\widehat\pi}\|_\infty
 &=\frac{\gamma}{1-\gamma}
       (\widetilde p-\widetilde q)
       \big(1-\widehat\pi(\phi\mid s_0)\big)\nonumber\\
 &\geq4\epsilon\big(1-\widehat\pi(\phi\mid s_0)\big).
 \label{eq:lb2-random-policy-gap}
\end{align}
The equality holds because the values at the absorbing states do
not depend on the policy. Since $s_0$ is visited only initially,
the same calculation applies to policies depending on the past.

To distinguish the two MDPs, we bound the KL divergence of their
data distributions. The models interchange $p$ and $q$ between
actions $0$ and $1$, and all other transition distributions and
rewards are identical. With $N$ independent samples per
state-action pair, independence gives
\begin{align}
 \text{KL}(\mathbb P_0\|\mathbb P_1)
 &=\E_{\mathbb P_0}\!\left[
       \log\frac{\mathbb P_0}{\mathbb P_1}\right]\nonumber\\
 &=N\big(\text{KL}(p\|q)
                   +\text{KL}(q\|p)\big)\nonumber\\
 &=N\big(\text{KL}(p_\wedge\|p_\wedge+\varepsilon)
             +\text{KL}(p_\wedge+\varepsilon\|p_\wedge)
       \big).
 \label{eq:lb2-data-kl}
\end{align}
The ratio in the expectation is evaluated at the observed dataset.
For any $x,y\in(0,1)$, the inequality $\log t\leq t-1$ gives
\begin{align*}
 \text{KL}(x\|y)
 &\leq x\left(\frac xy-1\right)
 +(1-x)\left(\frac{1-x}{1-y}-1\right)\\
 &=\frac{(x-y)^2}{y(1-y)}.
\end{align*}
When $x,y\in[p_\wedge,2p_\wedge]$, the denominator is at least
$p_\wedge(1-2p_\wedge)\geq3p_\wedge/4$. Therefore
\begin{align}
 \text{KL}(\mathbb P_0\|\mathbb P_1)
 &\leq\frac{8N\varepsilon^2}{3p_\wedge}
 \leq\frac{4N\varepsilon^2}{p_\wedge}.
 \label{eq:lb2-data-kl-upper}
\end{align}
In particular, if
\begin{equation}\label{eq:lb2-sufficient-sample-upper}
 N\leq\frac{p_\wedge}{32\varepsilon^2}
 =\frac{\gamma^2\delta^2}
 {2048p_\wedge(1-\gamma)^2\epsilon^2\log^2(1/p_\wedge)
  \log^2\!\left(\dfrac{4}
  {p_\wedge^3[\log(1/p_\wedge)-\delta]^2}\right)},
\end{equation}
then the KL divergence of the two data distributions is at most
$1/8$.

For any event $E$, grouping its probability and that of its
complement in the KL divergence gives
\[
 \text{KL}(\mathbb P_0\|\mathbb P_1)
 \geq\text{KL}\big(\mathbb P_0(E)\|\mathbb P_1(E)\big)
 \geq2\big(\mathbb P_0(E)-\mathbb P_1(E)\big)^2.
\]
The first inequality follows from convexity of $t\log t$, applied
separately on the event and its complement. The second follows
because, for fixed $y$, the function $x\mapsto\text{KL}(x\|y)$
is zero with zero derivative at $x=y$ and has second derivative
$1/[x(1-x)]\geq4$; boundary cases follow by continuity.
It follows that
\begin{equation}\label{eq:lb2-event-difference}
 \sup_E|\mathbb P_0(E)-\mathbb P_1(E)|\leq\frac14.
\end{equation}
Including independent algorithmic randomization with the observations
leaves the KL divergence and~\eqref{eq:lb2-event-difference}
unchanged, since its law is identical in both models.

Given the returned policy, choose model $0$ when
$\widehat\pi(0\mid s_0)\geq1/2$, and choose model $1$ otherwise.
Under $\mathcal M_0$, choosing model $1$ implies
$\widehat\pi(0\mid s_0)<1/2$ and hence a value loss greater than
$2\epsilon$ by~\eqref{eq:lb2-random-policy-gap}. Under
$\mathcal M_1$, choosing model $0$ implies
$\widehat\pi(1\mid s_0)\leq1/2$ and hence a value loss at least
$2\epsilon$. Both losses are strictly greater than $\epsilon$.
Equation~\ref{eq:lb2-event-difference} gives
\begin{align*}
 &\mathbb P_0\big(\widehat\pi(0\mid s_0)<1/2\big)
  +\mathbb P_1\big(\widehat\pi(0\mid s_0)\geq1/2\big)\\
 &=1-\mathbb P_0\big(\widehat\pi(0\mid s_0)\geq1/2\big)
       +\mathbb P_1\big(\widehat\pi(0\mid s_0)\geq1/2\big)
 \geq\frac34.
\end{align*}
Consequently, for every learning algorithm,
\begin{equation}\label{eq:lb2-completed-testing-bound}
 \max_{\phi\in\{0,1\}}
 \mathbb P_\phi\!\left(
 \|V_\phi^*-V_\phi^{\widehat\pi}\|_\infty>\epsilon\right)
 \geq\frac38>\frac18.
\end{equation}
Taking the infimum over learning algorithms preserves this lower
bound.

For $|\mathcal S|>3$, add $|\mathcal S|-3$ absorbing states with
reward zero, unreachable from $s_0$. For $|\mathcal A|>2$, give
each additional action at $s_0$ the probabilities $q$ and $1-q$
in both models. At absorbing states, every action remains absorbing
with the same reward. The minimal non-zero nominal probability
remains $p_\wedge$, and the added rows contribute zero to
\eqref{eq:lb2-data-kl}. Since every action other than $\phi$ at
$s_0$ has worst-case probability $\widetilde q$ of reaching $s_1$,
\eqref{eq:lb2-random-policy-gap} continues to hold. Moreover,
$\widehat\pi(0\mid s_0)\geq1/2$ implies
$\widehat\pi(1\mid s_0)\leq1/2$, so the same error bound applies.

There are $N$ samples per state-action pair, hence
$N|\mathcal S||\mathcal A|$ samples in total. Multiplying
\eqref{eq:lb2-sufficient-sample-upper} by $|\mathcal S||\mathcal A|$
gives
\[
 N|\mathcal S||\mathcal A|
 \leq\frac{\gamma^2\delta^2|\mathcal S||\mathcal A|}
 {2048p_\wedge(1-\gamma)^2\epsilon^2\log^2(1/p_\wedge)
  \log^2\!\left(\dfrac{4}
  {p_\wedge^3[\log(1/p_\wedge)-\delta]^2}\right)}.
\]
Combining this sample condition with
\eqref{eq:lb2-completed-testing-bound} and taking the infimum over
all learning algorithms gives \eqref{eq:inf}.

\subsubsection*{Case 2: $\delta\geq\log(1/p_\wedge)$}
With the same states and rewards, we set
\[
 p=1,\qquad q=1-p_\wedge.
\]
Thus action $\phi$ at $s_0$ reaches $s_1$ with probability one,
whereas action $1-\phi$ reaches $s_2$ with nominal probability
$p_\wedge$. A transition to $s_2$ with probability one has KL
divergence $\log(1/p_\wedge)\leq\delta$ from the latter row.
The deterministic row of action $\phi$ cannot change its support.
Consequently, the worst-case probabilities of reaching $s_1$ are
\[
 \widetilde p=1,\qquad\widetilde q=0.
\]
The same robust Bellman calculation as above gives
\begin{equation}\label{eq:lb2-support-policy-gap}
 \|V_\phi^*-V_\phi^{\widehat\pi}\|_\infty
 =\frac{\gamma}{1-\gamma}
      \big(1-\widehat\pi(\phi\mid s_0)\big).
\end{equation}

Consider the event that all samples from both actions $0$ and $1$
at $s_0$ are $s_1$. In either model, one of these two rows is
deterministic and the other reaches $s_1$ with probability
$1-p_\wedge$. The event therefore has the same probability
$(1-p_\wedge)^N$ under both models. Conditional on this event,
the entire observed dataset has the same distribution in both
models: the samples from these two rows are fixed, and all
remaining rows have identical distributions. The same is true
after including the independent randomization of the algorithm.

Use the same decision rule as before: choose model $0$ if
$\widehat\pi(0\mid s_0)\geq1/2$ and model $1$ otherwise.
The sum of its conditional error probabilities on the event is
one. Hence
\begin{align*}
 &\mathbb P_0\big(\widehat\pi(0\mid s_0)<1/2\big)
 +\mathbb P_1\big(\widehat\pi(0\mid s_0)\geq1/2\big)\\
 &\hspace{2em}\geq(1-p_\wedge)^N
 \geq1-Np_\wedge\geq\frac34,
\end{align*}
where the last inequality follows from
$N\leq1/(4p_\wedge)$.
An incorrect choice of model implies
$\widehat\pi(\phi\mid s_0)\leq1/2$ and therefore a value loss
at least $\gamma/[2(1-\gamma)]$, which is strictly greater than
$\epsilon$ under the stated accuracy condition.
It follows that the maximum failure probability is at least $3/8$.
Additional states and actions are added exactly as above, with
every additional action at $s_0$ having nominal probabilities
$q=1-p_\wedge$ and $1-q=p_\wedge$ in both models. They preserve
both the conditional equality of the data distributions and the
minimal non-zero nominal probability $p_\wedge$.

Taking the infimum over learning algorithms gives \eqref{eq:inf},
which completes the proof of Theorem~\ref{thm:lb2}.
% \begin{lemma}
% \label{lem:y-lower-bound}
% Let $x,y \in (0,1)$ and $\delta>0$ satisfy
% \begin{align}
% x = y \exp\!\left(-\frac{\delta}{y}\right).
% \end{align}
% Then it holds that
% \begin{align}
% y > \frac{\delta}{\log(1/x)} .
% \end{align}
% \end{lemma}

% \subsection{Lemmas and Proofs}

% \begin{proof}[Proof of Lemma \ref{lem:y-lower-bound}]
% Taking logarithms on both sides of the defining equation yields
% \begin{align}
% \log x = \log y - \frac{\delta}{y}.
% \end{align}
% Rearranging terms gives
% \begin{align}
% \frac{\delta}{y} = \log y - \log x = \log\!\left(\frac{y}{x}\right).
% \end{align}
% Since $y \in (0,1)$, we have $\log y < 0$, and therefore
% \begin{align}
% \log\!\left(\frac{y}{x}\right) < -\log x = \log(1/x).
% \end{align}
% Combining the above inequalities implies
% \begin{align}
% \frac{\delta}{y} < \log(1/x).
% \end{align}
% Because all quantities are positive, taking reciprocals reverses the inequality; multiplying by $\delta$ then gives
% \begin{align}
% y > \frac{\delta}{\log(1/x)}.
% \end{align}
% This completes the proof.
% \end{proof}
\section{Proof of Lemma \ref{lemma:worst}}
% \subsection{Lemmas and Proofs}
\begin{proof}[Proof of Lemma \ref{lemma:worst}]
From the duality Lagrange function, we can get that
    \begin{align}
\bar V
&=
\max_{\lambda \ge 0}
\left\{
-\lambda \log \Big(
\mathbb{E}_{P(\cdot \mid s,a)}
\big[ \exp(-V(s')/\lambda) \big]
\Big)
- \lambda \delta
\right\}.
\end{align}

Then, define 
$$\lambda^\star=
\arg\max_{\lambda \ge 0}
\left\{
-\lambda \log \Big(
\mathbb{E}_{P(\cdot \mid s,a)}
\big[ \exp(-V(s')/\lambda) \big]
\Big)
- \lambda \delta
\right\}. $$

We could get the worst-case mean value function as 
\begin{align*}
    \bar V
&=
-\lambda^\star
\log \Big(
\mathbb{E}_{P(\cdot \mid s,a)}
\big[ \exp(-V(s')/\lambda^\star) \big]
\Big)
- \lambda^\star \delta.
\end{align*}
Next, we set worst-case state, worst-case value function and corresponding probability as:
\begin{align}
s_{\min}
=
\arg\min_{s' : P(s' \mid s,a) > 0} V(s'),
V(s_{\min}) = V_{\min}, 
P_{\min}
= P(s_{\min} \mid s,a).
\end{align}

Then, from the fact $\mathbb{E}_{P(\cdot \mid s,a)}
\big[ \exp(-V(s')/\lambda^\star) \big]\ge
P_{\min} \exp(-V_{\min}/\lambda^\star)$, we could get the following result:
\begin{align}
\bar V
&\le\nonumber
-\lambda^\star
\log \Big(
P_{\min} \exp(-V_{\min}/\lambda^\star)
\Big)
- \lambda^\star \delta, \\
&\le\nonumber
V_{\min}
- \lambda^\star \log(P_{\min})
- \lambda^\star \delta
\\&\nonumber \le V_{\min}
+ \lambda^\star \delta
- \lambda^\star \delta
\\& \leq V_{\min}, 
\end{align}
where the third inequality follows from $ \log(P_{\min}) > -\delta$. 

% \begin{align}
% e^\delta > \frac{1}{P_{\min}}
% \quad \Longleftrightarrow \quad
% \log(P_{\min}) > -\delta,
% \end{align}

% \begin{align}
% -\lambda^\star \log(P_{\min}) - \lambda^\star \delta \le 0,
% \end{align}
Besides, combined with the fact $\bar V \geq V_{\min} $, we can conclude that
\begin{align}
\bar V = V_{\min}. \nonumber
\end{align}

\end{proof}

\newpage
\section*{NeurIPS Paper Checklist}

\begin{enumerate}

\item {\bf Claims}
    \item[] Question: Do the main claims made in the abstract and introduction accurately reflect the paper's contributions and scope?
    \item[] Answer: \answerYes{} % Replace by \answerYes{}, \answerNo{}, or \answerNA{}.
    \item[] Justification: Main claims made in the abstract and introduction accurately reflect the paper's contributions and scope.

\item {\bf Limitations}
    \item[] Question: Does the paper discuss the limitations of the work performed by the authors?
    \item[] Answer: \answerYes{} % Replace by \answerYes{}, \answerNo{}, or \answerNA{}.
    \item[] Justification: Please check Section \ref{sec:limit}. 

\item {\bf Theory assumptions and proofs}
    \item[] Question: For each theoretical result, does the paper provide the full set of assumptions and a complete (and correct) proof?
    \item[] Answer: \answerYes{} % Replace by \answerYes{}, \answerNo{}, or \answerNA{}.
    \item[] Justification: The paper provide the full set of assumptions and a complete and correct proof. 

    \item {\bf Experimental result reproducibility}
    \item[] Question: Does the paper fully disclose all the information needed to reproduce the main experimental results of the paper to the extent that it affects the main claims and/or conclusions of the paper (regardless of whether the code and data are provided or not)?
    \item[] Answer: \answerNA{} % Replace by \answerYes{}, \answerNo{}, or \answerNA{}.
    \item[] Justification: The paper does not include experiments. 

\item {\bf Open access to data and code}
    \item[] Question: Does the paper provide open access to the data and code, with sufficient instructions to faithfully reproduce the main experimental results, as described in supplemental material?
    \item[] Answer: \answerNA{} % Replace by \answerYes{}, \answerNo{}, or \answerNA{}.
    \item[] Justification: The paper does not include experiments. 

\item {\bf Experimental setting/details}
    \item[] Question: Does the paper specify all the training and test details (e.g., data splits, hyperparameters, how they were chosen, type of optimizer) necessary to understand the results?
    \item[] Answer: \answerNA{} % Replace by \answerYes{}, \answerNo{}, or \answerNA{}.
    \item[] Justification: The paper does not include experiments. 

\item {\bf Experiment statistical significance}
    \item[] Question: Does the paper report error bars suitably and correctly defined or other appropriate information about the statistical significance of the experiments?
    \item[] Answer: \answerNA{} % Replace by \answerYes{}, \answerNo{}, or \answerNA{}.
    \item[] Justification: The paper does not include experiments. 

\item {\bf Experiments compute resources}
    \item[] Question: For each experiment, does the paper provide sufficient information on the computer resources (type of compute workers, memory, time of execution) needed to reproduce the experiments?
    \item[] Answer: \answerNA{} % Replace by \answerYes{}, \answerNo{}, or \answerNA{}.
    \item[] Justification: The paper does not include experiments. 
    
\item {\bf Code of ethics}
    \item[] Question: Does the research conducted in the paper conform, in every respect, with the NeurIPS Code of Ethics \url{https://neurips.cc/public/EthicsGuidelines}?
    \item[] Answer: \answerNA{} % Replace by \answerYes{}, \answerNo{}, or \answerNA{}.
    \item[] Justification: The paper does not include experiments. 

\item {\bf Broader impacts}
    \item[] Question: Does the paper discuss both potential positive societal impacts and negative societal impacts of the work performed?
    \item[] Answer: \answerNA{} % Replace by \answerYes{}, \answerNo{}, or \answerNA{}.
    \item[] Justification: The paper is theory work, which does not include any directly social impact. 
    
\item {\bf Safeguards}
    \item[] Question: Does the paper describe safeguards that have been put in place for responsible release of data or models that have a high risk for misuse (e.g., pre-trained language models, image generators, or scraped datasets)?
    \item[] Answer: \answerNA{} % Replace by \answerYes{}, \answerNo{}, or \answerNA{}.
    \item[] Justification: The paper does not include experiments. 

\item {\bf Licenses for existing assets}
    \item[] Question: Are the creators or original owners of assets (e.g., code, data, models), used in the paper, properly credited and are the license and terms of use explicitly mentioned and properly respected?
    \item[] Answer: \answerNA{} % Replace by \answerYes{}, \answerNo{}, or \answerNA{}.
    \item[] Justification:  The paper does not use existing assets.

\item {\bf New assets}
    \item[] Question: Are new assets introduced in the paper well documented and is the documentation provided alongside the assets?
    \item[] Answer: \answerNA{} % Replace by \answerYes{}, \answerNo{}, or \answerNA{}.
    \item[] Justification: The paper does not release new assets.

\item {\bf Crowdsourcing and research with human subjects}
    \item[] Question: For crowdsourcing experiments and research with human subjects, does the paper include the full text of instructions given to participants and screenshots, if applicable, as well as details about compensation (if any)? 
    \item[] Answer: \answerNA{} % Replace by \answerYes{}, \answerNo{}, or \answerNA{}.
    \item[] Justification: The paper does not involve crowdsourcing nor research with human subjects.

\item {\bf Institutional review board (IRB) approvals or equivalent for research with human subjects}
    \item[] Question: Does the paper describe potential risks incurred by study participants, whether such risks were disclosed to the subjects, and whether Institutional Review Board (IRB) approvals (or an equivalent approval/review based on the requirements of your country or institution) were obtained?
    \item[] Answer: \answerNA{} % Replace by \answerYes{}, \answerNo{}, or \answerNA{}.
    \item[] Justification: The paper does not involve crowdsourcing nor research with human subjects.

\item {\bf Declaration of LLM usage}
    \item[] Question: Does the paper describe the usage of LLMs if it is an important, original, or non-standard component of the core methods in this research? Note that if the LLM is used only for writing, editing, or formatting purposes and does \emph{not} impact the core methodology, scientific rigor, or originality of the research, declaration is not required.
    %this research? 
    \item[] Answer: \answerYes{} % Replace by \answerYes{}, \answerNo{}, or \answerNA{}.
    \item[] Justification: The LLM was used solely for grammar polishing and \LaTeX{} formatting in this work; all technical results have been independently validated.

\end{enumerate}

\end{document}